\documentclass{ws-aa}

\usepackage[most]{tcolorbox}
\usepackage{hyperref,xcolor}

\ifpdf
\DeclareGraphicsExtensions{.eps,.pdf,.png,.jpg}
\else
\DeclareGraphicsExtensions{.eps}
\fi

\newcommand{\N}{\mathcal{N}}
\newcommand{\x}{\textbf{x}}
\newcommand{\y}{\textbf{y}}

\begin{document}

\markboth{Yeonjong Shin}{Layer-wise Training for Deep Linear Networks}

%
\catchline{}{}{}{}{}
%

\title{Effects of Depth, Width, and Initialization: A Convergence Analysis of Layer-wise Training for Deep Linear Neural Networks}

\author{YEONJONG SHIN\footnote{Typeset names in 8~pt roman,  
		uppercase. Use the footnote to indicate the
		present or permanent address of the author.}}

\address{Division of Applied Mathematics, Brown University,\\
	Providence, RI, 02912, USA
	\\
	yeonjong\_shin@brown.edu}

\maketitle

\begin{history}
	\received{(Day Month Year)}
	\revised{(Day Month Year)}
\end{history}

\begin{abstract}
  Deep neural networks have been used in various machine learning applications and achieved tremendous empirical successes. 
  However, training deep neural networks is a challenging task.
  Many alternatives have been proposed in place of end-to-end back-propagation.
  Layer-wise training is one of them, which trains a single layer at a time, rather than trains the whole layers simultaneously.
  In this paper, we study a layer-wise training using a block coordinate gradient descent (BCGD) for deep linear networks.
  We establish a general convergence analysis of BCGD and found the optimal learning rate, which results in the fastest decrease in the loss.
  We identify the effects of depth, width, and initialization.
  When the orthogonal-like initialization is employed, we show that 
  the width of intermediate layers plays no role in gradient-based training
  beyond a certain threshold.
  Besides, we found that the use of deep networks could drastically accelerate convergence when it is compared to those of a depth 1 network, even when the computational cost is considered. 
  Numerical examples are provided to justify our theoretical findings and demonstrate the performance of layer-wise training by BCGD.
\end{abstract}

\keywords{deep linear neural networks; layer-wise training; block coordinate gradient descent}

\ccode{Mathematics Subject Classification 2000: 65K05, 65F10, 68Q25, 90C26}

\section{Introduction}
Deep learning has drawn a lot of attention from both academia and industry
due to its tremendous empirical success in various applications \cite{krizhevsky2012imagenet,hinton2012deep,silver2016mastering,wu2016google}.
One of the key components in the success of deep learning is the intriguing ability of gradient-based optimization methods.
Despite of the non-convex and non-smooth nature of the loss function,
it somehow finds a local (or global) minimum, which performs well in practice. 
Mathematical analysis of this phenomenon has been undertaken.
There are several theoretical works, which show that 
under the assumption of over-parameterization,
more precisely, very wide networks,
the (stochastic) gradient descent algorithm finds a global minimum
\cite{allen2018convergence,du2018gradient-DNN,du2018gradient-shallow,zou2018stochastic,oymak2019towards}.
These theoretical progresses have its own importance, however, 
it does not directly help practitioners to have better training results. 
This is mainly because there are still many parameters to be determined a priori; learning rate, the depth of network, the width of intermediate layers, optimization algorithms with its own internal parameters, to name just a few. 
The learning rates from existing theoretical works are not applicable in practice. For example, when a fully-connected ReLU network of depth 10 is trained over 1,000 training data, 
theoretically guaranteed learning rate is
either 
$\eta \approx \frac{1}{1000^2\cdot 2^{10}}\approx 10^{-9}$ \cite{du2018gradient-DNN}
or
$\eta \approx \frac{1}{1000^4\cdot 10^{2}} \approx 10^{-14}$ \cite{allen2018convergence}.
Thus, 
practitioners typically choose these aforementioned parameters by either a grid search
or trial and error.

%
Despite its expressive power, training deep neural networks is not an easy task.
It has been widely known that the deeper the network is, the harder it is to be trained \cite{Srivastava2015training}.
Empirical success of deep learning heavily relies on
numerous engineering tricks used in the training process.
These includes but not limited to dropout \cite{Srivastava2014Dropout},
dropconnect \cite{Wan2013DropConnect}, batch-normalization \cite{Ioffe2015BN}, weight-normalization \cite{Salimans2016WN},
pre-training \cite{Dahl2011PreTrain}, and data augmentation \cite{Cirecsan2012DataAugmentation}.
Although these techniques are shown to be effective in many machine learning applications, it lacks rigorous justifications
and hinders a thorough mathematical understanding of the training process of deep learning.
The layer-wise training is an alternative to the standard end-to-end back-propagation, especially for training deep neural networks.
The underlying principle 
is to train only a few layers (or a single layer) at a time, 
rather than train the whole layers simultaneously. 
This approach is not new and has been proposed in several different contexts.
One stream of layer-wise training
is adaptive training. At each stage, only a few layers (or a single) are trained.
Once training is done, new layers are added.
By fixing all the previously trained layers for the rest of the training,
only newly added layers are trained. 
This procedure is repeated.
The works of this direction include \cite{Fahlman1990cascade,Lengelle1996training,Kulkarni2017LayerWiseTrain,Belilovsky2018LayerWiseTrain,Marquez2018DeepCascade,Malach2018PCADL,Mosca2017Boosting,Huang2017Boosting}.
Another stream of layer-wise training is the block coordinate descent (BCD) method \cite{Zhang2017BCDtwo,Zeng2018BCDanalysis,Carreira2014BCDtwo,Askari2018BCDtwo,Gu2018BCDtwo,Lau2018BCDthree,Taylor2016BCTthree}. 
The BCD is a Gauss-Seidel type of gradient-free methods, 
which trains each layer at a time by freezing all other layers,
in a sequential order.
Thus, all layers are updated once in every sweep of training.
This paper concerns with the layer-wise training in this line of approach.
In \cite{Hinton2006ReduceDimNN,Bengio2007LayerWiseTrain}, layer-wise training is employed as a pretraining strategy.

Deep linear network (DLN) is a neural network that uses linear activation functions. 
Although DLN is not a popular choice in practice, it is an active research subject 
as it is a class of decent simplified models for understanding the deep neural network with non-linear activation functions \cite{Saxe2013orthInit,Hardt2016identityMatters,Arora2018optAccelerationDLN,Arora2018convergenceDLN,Bartlett2019gdIdentityDLN}.
DLN has a trivial representation power (product of weight matrices),
however,
its training process is not trivial at all. 
It has been studied the loss surface of DLNs \cite{Lu2017depth,Kawaguchi2016deep,Laurent2018deep}
and it is shown that although the loss surface is not convex, there are no spurious local minima. 
The works of \cite{Arora2018convergenceDLN, Du2019width} studied a convergence analysis of gradient descent for DLNs, under various settings. 
\cite{Arora2018convergenceDLN} showed that under some assumptions, the gradient descent finds a global optimum. 
The learning rate from the analysis, however, is not applicable in practice as it requires prior knowledge of the global minimizer.
The theoretically guaranteed learning rate of \cite{Arora2018convergenceDLN} should meet
$\eta \le \frac{c^{(4L-2)/L}}{6144L^3 \|\bm{W}^*\|_F^{(6L-4)/L}}$,
where $\bm{W}^*$ is the global minimizer, $c$ is a constant related to the initial error, and $L$ is the depth.
\cite{Du2019width} showed that under the assumptions of Gaussian random initialization, and severely wide networks, the gradient descent finds a global optimum. 
The learning rate from the analysis of \cite{Du2019width} does not require any prior knowledge of $\bm{W}^*$ and can be applied in practice.
However, we found that it leads divergence of GD in all of our tests.
The theoretically guaranteed width of \cite{Du2019width} is too large to be used.
For example, if the condition number of the input data matrix (full rank) is $100$, 
the width should be at least $(100^2)^3=10^{12}$.

\begin{figure}[htbp]
	\centerline{
		\includegraphics[height=6cm]{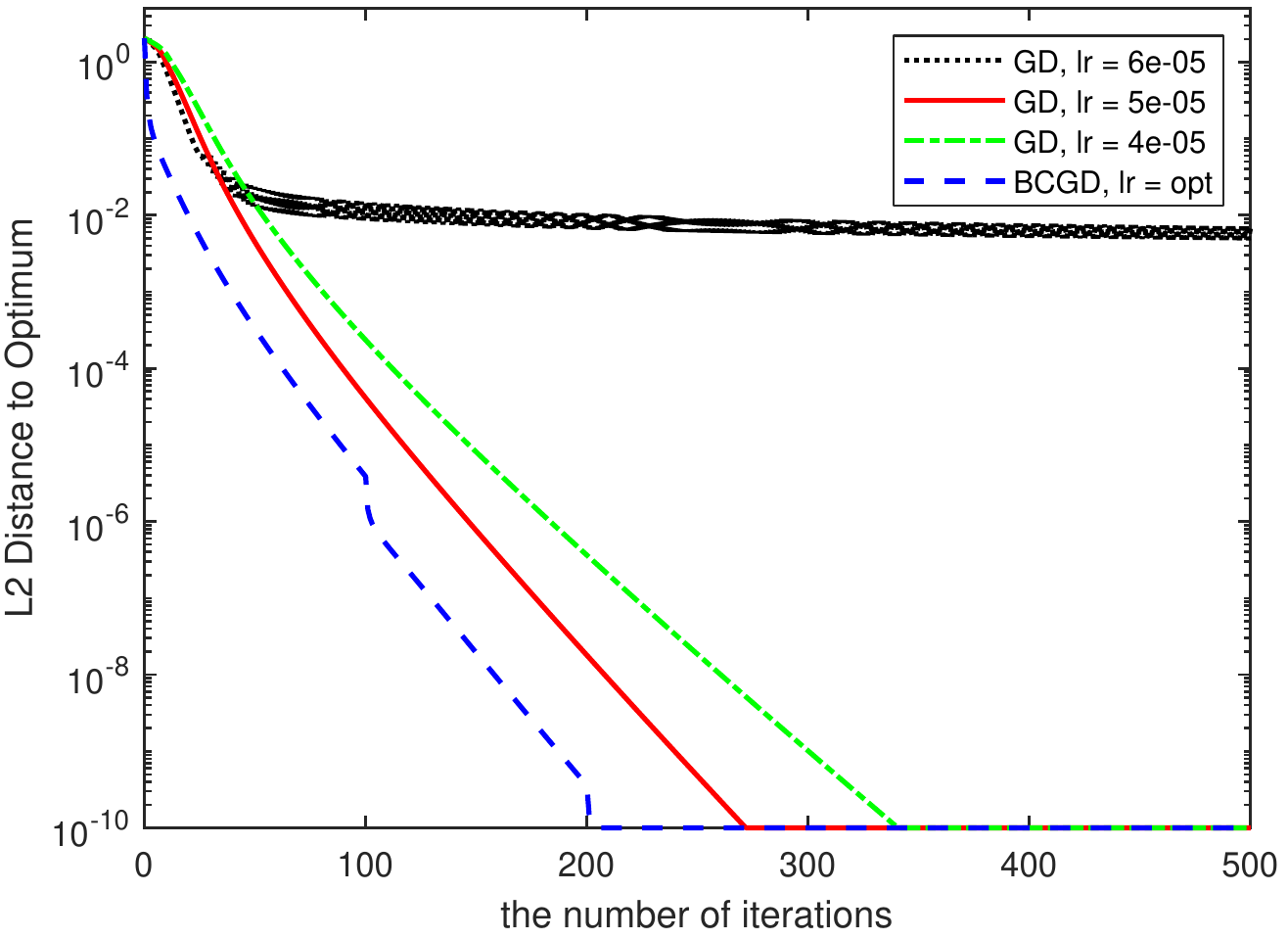}}
	\caption{The $L_2$-distance to the optimum with respect to the number of iterations.
	The input dimension is 128 and 
	the output dimension is 10.
	A 100-layer linear network of width 128 is employed
	and the orthogonal initialization is employed. 
	BCGD uses the optimal learning rate \eqref{LR-l2-Optimal} 
	and GD uses learning rates from trial-and-error.
	}
	\label{fig:motivation}
\end{figure}

In this paper, we study a layer-wise training for DLNs
using a block coordinate gradient descent (BCGD) \cite{Tseng2009BCGD-sepa,Tseng2009BCGD-linC}.
Similar to BCD, the BCGD trains each layer at a time 
in a sequential order 
by freezing all other layers at their last updated values.
However, a key difference is the use of gradient descent 
in every update.
\textit{We aim to identify the effects of depth, width, and initialization in the training process}
through the lens of DLNs.
We first establish a general convergence analysis and found the optimal learning rate, which leads to the fastest decrease in the loss for the next iterate.
More importantly, the optimal learning rate can directly be applied in practice.
Neither trial and error nor a grid search for tuning parameters are required.
To illustrate the performance of BCGD with the optimal learning rate, 
we consider a learning task of fitting 600 data (see Section~\ref{subsec:Random} for details)
and plot the training loss trajectories by BCGD and GD
in Figure~\ref{fig:motivation}.
Despite the fact that 
BCGD updates only a single matrix per iteration,
while GD updates all the weight matrices per iteration,
we clearly see that BCGD converges drastically faster than GD. 
The learning rates for GD are found by trial-and-error.

Next, we show that when the orthogonal-like initialization is employed,
as long as the width of intermediate layers is greater than or equal to both the input and output dimensions, \textit{the width plays no role in any gradient-based training}.
Also, we rigorously prove that when (i) the orthogonal-like initialization is used, (ii) the initial loss is sufficiently small,  
whenever the depth is sufficiently large,
the convergence to the global optimum (within machine accuracy) is guaranteed by updating 
each weight matrix only once. 
Furthermore, we found that 
a well-chosen depth could result in a significant acceleration in convergence
when it is compared to those of a single layer, 
even when the computational cost is considered.
This clearly demonstrates the benefit of using deep networks (over-parameterization via depth).
Similar behavior was empirically reported in \cite{Arora2018optAccelerationDLN} as implicit acceleration.

Lastly, we establish a convergence analysis of the block coordinate stochastic gradient descent (BCSGD).
Our analysis indicates that the BCSGD cannot reach the global optimum, however, the converged loss will be staying close to the global optimum. This can be understood as an implicit regularization, which avoids over-fitting.

The rest of paper is organized as follows.
In Section~\ref{sec:setup}, we present 
the mathematical setup and introduce the block coordinate (stochastic) gradient descent.
We then present a general convergence analysis and the optimal learning rate in Section~\ref{sec:analysis}.
In Section~\ref{sec:example}, several numerical examples using both synthetic and real data sets are presented to demonstrate the effectiveness of the layer-wise training by BCGD and justify our theoretical findings.

\section{Setup and Preliminary} \label{sec:setup}
Let $\mathcal{N}^L:\mathbb{R}^{d_{\text{in}}} \to \mathbb{R}^{d_{\text{out}}}$ be a feed-forward linear neural network with $L$ layers and having $n_\ell$ neurons in the $\ell$-th layer. 
We denote the weight matrix in the $\ell$-th layer 
by $\bm{W}_\ell \in \mathbb{R}^{n_\ell \times n_{\ell-1}}$. 
Here $n_0 = d_{\text{in}}$ and $n_L = d_{\text{out}}$.
Let $\bm{\theta} = \{\bm{W}_\ell\}_{\ell=1}^L$
be the set of all weight matrices. 
Then the $L$-layer linear neural network can be written as 
\begin{align*} 
    \mathcal{N}^L(\textbf{x};\bm{\theta}) = \bm{W}_{L}\bm{W}_{L-1}\cdots \bm{W}_{1}\x.
\end{align*}
Given a set of training data 
$\mathcal{T}=\{(\x^i, \y^i)\}_{i=1}^m$,
the goal is to learn the parameters $\{\bm{W}_j\}_{j=1}^L$
which minimize the loss function $\mathcal{L}(\bm{\theta})$ defined by
\begin{equation} \label{def:loss}
    \mathcal{L}(\bm{\theta}) = 
    \sum_{i=1}^m 
    \mathcal{L}_i(\bm{\theta}), \qquad
    \mathcal{L}_i(\bm{\theta}) =
    \sum_{j=1}^{d_{\text{out}}}
    \ell(\N^L_j(\x^i; \bm{\theta}); \y^i_j).
\end{equation}
Here $\ell(a;b)$ is a metric which measures the discrepancy between the prediction and the output data.
For example, the choice of $\ell(a;b) = (a-b)^p/p$ results in 
the standard $L_p$-loss function.

For a matrix $\bm{A} \in \mathbb{R}^{m\times n}$, the spectral norm, the condition number
and 
the scaled condition number
are defined to be
$$
\|\bm{A}\| = \max_{\|x\|_2=1} \|\bm{A}\x\|_2, \qquad 
\kappa_r(\bm{A})=\frac{\sigma_{\max}(\bm{A})}{\sigma_{r}(\bm{A})},
\qquad
\tilde{\kappa}(\bm{A}) = \frac{\|\bm{A}\|_F}{\sigma_{\min}(\bm{A})},
$$
respectively.
Here $\|\cdot\|_2$ is the Euclidean norm, 
$\|\cdot\|_F$ is the Frobenius norm,
$\sigma_{\max}(\cdot)$ is the largest singular value,
and 
$\sigma_{r}(\cdot)$ is
the $r$-th largest singular value.
Also, 
we denote 
the $\min\{m,n\}$-th largest singular value
by $\sigma_{\min}(\cdot)$.
When $r = \min\{m,n\}$, we simply write the condition number 
as $\kappa(\cdot)$.

\subsection{Global minimum of $L_2$ loss}
Since this paper mainly concerns with the standard $L_2$-loss, here we discuss its global minimum, which depends on the network architecture being used. 
Let $\bm{X} =[\x^1,\cdots,\x^m] \in \mathbb{R}^{n_0 \times m}$ be the input data matrix 
and $\bm{Y} = [\y^1, \cdots, \y^m] \in \mathbb{R}^{n_L \times m}$ be the output data matrix.
Then, the problem of minimizing the $L_2$-loss function is
\begin{equation} \label{def:depthL-prob}
    \min_{\bm{W}_j\in \mathbb{R}^{n_j\times n_{j-1}}, 1\le j\le L} \|\bm{W}_{L:1}\bm{X} - \bm{Y}\|_F^2, \quad \text{where} \quad 
    \bm{W}_{L:1}:=\bm{W}_L\cdots \bm{W}_1.
\end{equation}
This problem is closely related to 
\begin{equation} \label{def:depth1-prob}
    \min_{\bm{W}\in \mathbb{R}^{n_L\times n_0}} \|\bm{W}\bm{X} - \bm{Y}\|_F^2, \quad \text{subject to} \quad \text{rank}(\bm{W}) \le \min \{n_0,\cdots, n_L\}.
\end{equation}
Since the rank of $\bm{W}_{L:1}$ is at most $n^*:=\min \{n_0,\cdots, n_L\}$, 
the minimized losses from \eqref{def:depthL-prob} and \eqref{def:depth1-prob} should be the same. 
Thus, if $\{\bm{W}_\ell^{*}\}_{\ell=1}^L$ is a solution of \eqref{def:depthL-prob},
$\bm{W}^*_{L:1}$ should be a global minimizer of \eqref{def:depth1-prob}. 
Therefore, a global minimizer of \eqref{def:depthL-prob} and its corresponding minimized loss can be understood through \eqref{def:depth1-prob}.
In \ref{app:lsq-sol}, we briefly discuss the solutions of \eqref{def:depth1-prob}.

%

\subsection{Block Coordinate Gradient Descent} \label{subsec:bcgd}
In this paper, we consider 
the block coordinate gradient descent (BCGD).
The method commences with an initialization 
$\bm{\theta}^{\textbf{k}_{0}} = \{\bm{W}_{\ell}^{(0)}\}_{\ell=1}^L$.
Let $\textbf{k} = (\text{k}_1,\cdots,\text{k}_L)$ be a multi-index,
where each $\text{k}_\ell$ indicates the number of updates of the $\ell$-th layer weight matrix $\bm{W}_\ell$.
After the $k$-th iteration, 
we obtain a multi-index $\textbf{k}^{(k)} = (\text{k}_1^{(k)},\cdots, \text{k}_L^{(k)})$
and its corresponding parameters are $\bm{\theta}^{\textbf{k}^{(k)}} = \{\bm{W}_\ell^{(\text{k}_\ell^{(k)})}\}_{\ell=1}^L$.
Given $\textbf{k}^{(k)} = (\text{k}_1^{(k)},\cdots,\text{k}_L^{(k)})$,
let 
$$
\bm{W}_{i:j}^{\textbf{k}^{(k)}} := \bm{W}_i^{(\text{k}_i^{(k)})}\bm{W}_{i-1}^{(\text{k}_{i-1}^{(k)})}\cdots \bm{W}_j^{(\text{k}_j^{(k)})}, \qquad 1\le j < i \le L.
$$
If $\text{k}_j^{(k)}=k$ for all $j$, 
we write $\bm{W}_{i:j}^{(k)} := \bm{W}_i^{(k)}\bm{W}_{i-1}^{(k)}\cdots \bm{W}_j^{(k)}$
for $1\le j < i \le L$.
For notational completeness, we set $\bm{W}_{i:j} = \bm{I}$ whenever $i<j$.
Also, we simply write $\bm{W}_{L:1}^{\textbf{k}}$
as $\bm{W}^{\textbf{k}}$.

At the $(Lk+\ell)$-th iteration of BCGD, 
the $\mathfrak{i}(\ell)$-th layer weight matrix is updated 
according to 
\begin{equation} \label{def:bcgd}
\begin{split}
\bm{W}_{\mathfrak{i}(\ell)}^{(k+1)} = 
\bm{W}_{\mathfrak{i}(\ell)}^{(k)}
- \eta_{\mathfrak{i}(\ell)}^{\textbf{k}_{(k,\ell-1)}} \frac{\partial \mathcal{L}(\bm{\theta}) }{\partial \bm{W}_{\mathfrak{i}(\ell)}}\bigg|_{\bm{\theta} = \bm{\theta}^{\textbf{k}_{(k,\ell-1)}}},
\end{split}
\end{equation}
where $\textbf{k}_{(k,\ell)} = \textbf{k}_{(k,\ell-1)} + \bm{e}_{\mathfrak{i}(\ell)}$, $\bm{e}_{j} = (0,\cdots, 0, \overset{j \text{-th}}{1}, 0, \cdots, 0)$, and
$$
\textbf{k}_{(k,0)} = \textbf{k}_{k} = (k,\cdots,k), \quad 
\textbf{k}_{(k,L)} = \textbf{k}_{k+1} =(k+1,\cdots,k+1).
$$
Here 
$\mathfrak{i}(\ell) = \ell$ if the ascending (bottom to top) ordering is employed
and $\mathfrak{i}(\ell) = L-\ell+1$ if the descending (top to bottom) ordering is employed.
%
%
We refer the BCGD with the bottom to top (top to bottom) ordering 
as the ascending (descending) BCGD.
Given a linear neural network of depth $L$, 
a single sweep of the ascending (descending) BCGD
consists of $L$-iterations starting from the first layer (the last layer) to the last layer (the first layer).
That is, after a single sweep, all weight matrices are updated only once, in the order of from $\bm{W}_1$ to $\bm{W}_L$ ($\bm{W}_L$ to $\bm{W}_1$).
When $L=1$, the BCGD is identical to GD.

\subsection{Initialization}
\label{subsec:initialization}

Any gradient-based optimization starts with an initialization 
$\bm{\theta}^{\textbf{k}_0} = \{\bm{W}_{\ell}^{(0)}\}_{\ell=1}^L$,
where $\textbf{k}_{0} = (0,\cdots,0)$.
Here we present three initialization schemes for training DLNs.

Let $\bm{A}$ be a matrix of size $m\times n$
and $\bm{B}$ be of size $k \times s$
where $m \ge k, n \ge s$.
We say $\bm{A}$ is equivalent to $\bm{B}$ upto zero-valued padding
if 
$$
\bm{A} = \begin{bmatrix}
\bm{B} & \bm{0} \\
\bm{0} & \bm{0}
\end{bmatrix},
$$
and write $\bm{A} \approxeq \bm{B}$.
Suppose $\min\{m,n\} > k=s$. 
We then write $\bm{A} \approxeq_1 \bm{B}$ if 
$\bm{A} \approxeq \tilde{\bm{B}}$ where
$\tilde{\bm{B}}$ is a square matrix of size $\min\{m,n\}$ such that
$$
\tilde{\bm{B}} = \begin{bmatrix}
\bm{B} & \bm{0} \\
\bm{0} & \bm{I}_{\min\{m,n\} - k}
\end{bmatrix}.
$$
Here $\bm{I}_n$ is the identity matrix of size $n$.
We consider the following weight initialization schemes.
\begin{itemize}
	\item Orthogonal Initialization \cite{Saxe2013orthInit}: $\bm{W}_j^{(0)} \approxeq \bm{Q}^j_{\min\{n_j,n_{j-1}\}}$ for all $1 \le j \le L$,
	where $\bm{Q}_{n}$ is an orthogonal matrix of size 
	$n$.
	\begin{itemize}
		\item Orth-Identity Initialization:
		$\bm{W}_j^{(0)} \approxeq_1 \bm{Q}^j_{\min\{n_j,n_{j-1},\max\{n_0,n_L\}\}}$ for $1 \le j \le L$.
		This is a special case of orthogonal initialization
		that is proposed in the present work.
		\item Identity Initialization \cite{Hardt2016identityMatters,Bartlett2019gdIdentityDLN}: $\bm{W}_j^{(0)} \approxeq \bm{I}_{\min\{n_j,n_{j-1}\}}$ for $1 \le j \le L$.
	\end{itemize}
	\item Balanced Initialization \cite{Arora2018convergenceDLN}:
	Given a randomly drawn matrix $\bm{W}^{(0)} \in \mathbb{R}^{n_L\times n_0}$,
	let us take a singular value decomposition $\bm{W}^{(0)} = \bm{U\Sigma V}^T$,
	where $\bm{U} \in \mathbb{R}^{n_L \times \min\{n_0,n_L\}}$,
	$\Sigma \in \mathbb{R}^{\min\{n_0,n_L\} \times \min\{n_0,n_L\}}$
	is diagonal, and
	$\bm{V} \in \mathbb{R}^{n_0 \times \min\{n_0,n_L\}}$
	have orthogonal columns.
	Set 
	$\bm{W}^{(0)}_L \approxeq \bm{U\Sigma}^{1/L}$,
	$\bm{W}^{(0)}_j \approxeq \bm{\Sigma}^{1/L}$ for $1< j < L$,
	$\bm{W}^{(0)}_1 \approxeq \bm{\Sigma}^{1/L}\bm{V}^T$.
	\item Random Initialization: $(\bm{W}_j^{(0)})_{ik} \sim N(0,\sigma_j^2)$ for all $1\le j \le L$.
	Often $\sigma_j^2$ is chosen to $1/n_{j-1}$ so that 
	the expected value of the square norm of each row is 1.
\end{itemize}
The orth-indentity initialization 
can be viewed as a hybrid initialization between the orthogonal and the identity initialization schemes. 
This paper primarily concerns with the orth-indentity initialization.

\section{Convergence Analysis} \label{sec:analysis}
In this section, we present a convergence analysis of BCGD
and establish the optimal learning rate.
The optimality is defined to be the learning rate
which results in the fastest decrease in the loss
at the current parameters.
The standard $L_2$-loss will be mainly discussed.
However, we also present a convergence result for 
general differentiable convex loss functions whose gradient are Lipshitz continuous in a bounded domain, such as $L_p$-loss where $p$ is even. 
We measure the approximation error in terms of the
distance to the global optimum.
For example, when the $L_2$-loss is employed, 
the error is $\mathcal{L}(\bm{W}^{\textbf{k}_k}) - \mathcal{L}(\bm{W}^*)=\|\bm{W}^{\textbf{k}_k}\bm{X} - \bm{W}^*\bm{X}\|_F^2$.

We first identify the effects of width in DLNs in gradient-based training under either the orth-identity or the balanced \cite{Arora2018convergenceDLN} initialization (Section~\ref{subsec:initialization}).

\begin{theorem} \label{thm:role of width}
    Suppose the weight matrices are initialized 
    according to either the orth-identity or the balanced initialization, described in
    Section~\ref{subsec:initialization}.
    Let $n_\ell$ be the width of the $\ell$-th layer.
    Then, the training process of
    any gradient-based optimization methods (including GD, SGD, BCGD, BCSGD) is
    independent of 
    the choice of $n_\ell$'s as long as it satisfies 
    \begin{equation} \label{role-width}
    \min_{1 \le \ell < L} n_\ell \ge \max\{n_0, n_L \}.
    \end{equation}
\end{theorem}
\begin{proof}
    The proof can be found in Appendix~\ref{app:thm:role of width}.
\end{proof}

Theorem~\ref{thm:role of width} implies that 
the width does not play any role in gradient-based training if the condition of \eqref{role-width} is met and 
the weight matrices are initialized in a certain manner.

However, the same conclusion does not follow if the random initialization is employed.
This indicates that the role of width highly depends on
how the weight matrices are initialized.
With a proper initialization, 
over-parameterization by the width 
can be avoided.

\subsection{Convergence of BCGD}

We first focus on the standard $L_2$ loss function
and present a general convergence analysis of BCGD.
We do not make any assumptions other than 
$\text{range}(\bm{Y}\bm{X}^\dagger) \subset \text{range}(\bm{W}_L^{(0)})$.
We follow the convention of 
$0\times \infty = \frac{1}{\infty} = 0 \times \frac{1}{0} = 0$. 
\begin{theorem} \label{thm:convg-l2}
    Let $\ell(z;b) = (z-b)^2/2$.
    Suppose 
    all columns of $\bm{W}_{L}^{(0)}$
    are initialized to be in a subspace $K$ in $\mathbb{R}^{n_L}$ such that
    $\text{range}(\bm{Y}\bm{X}^\dagger) \subset K$.
    Then, the $k$-th sweep (the $kL$-th iteration) of 
    BCGD \eqref{def:bcgd}
    with the learning rates of
    \begin{equation}  \label{LR-l2-loss-exact}
        \eta^{\textbf{k}_{(s,\ell-1)}}_\ell =
        \frac{\eta}{\|\bm{W}_{L:(\mathfrak{i}(\ell)+1)}^{\textbf{k}_{(s,\ell-1)}}\|^2
        	\|\bm{W}_{(\mathfrak{i}(\ell)-1):1}^{\textbf{k}_{(s,\ell-1)}}\bm{X}
        	\|^2}, \qquad 0 < \eta < 2,
    \end{equation}
    where 
    $\mathfrak{i}(\ell) = \ell$ if the ascending BCGD is employed
    and
    $\mathfrak{i}(\ell) = L-\ell+1$ if the descending BCGD is employed,
    satisfies
    \begin{equation} \label{Rate-l2-loss-exact}
        \mathcal{L}(\bm{W}^{\textbf{k}_{k}}) - \mathcal{L}(\bm{W}^*)
        \le 
        \left(\mathcal{L}(\bm{W}^{\textbf{k}_0}) - \mathcal{L}(\bm{W}^*)\right)
        \prod_{s=0}^{k-1}
        \prod_{\ell=1}^L
        \left(\gamma^{\textbf{k}_{(s,\ell-1)}}\right)^2,
    \end{equation}
    where $\bm{W}^* = \bm{Y}\bm{X}^\dagger$,
    $r_x= \text{rank}(\bm{X})$, 
    $r= \dim(K)$, and
    \begin{align*}
    \gamma^{\textbf{k}_{(s,\ell-1)}}
    =
    \max\left\{1-\frac{\eta}{\kappa_{r}^2(\bm{W}_{L:(\mathfrak{i}(\ell)+1)}^{\textbf{k}_{(s,\ell-1)}})\kappa^2_{r_x}(\bm{W}_{(\mathfrak{i}(\ell)-1):1}^{\textbf{k}_{(s,\ell-1)}}\bm{X})}, \eta -1 \right\}.
    \end{align*}
	Furthermore, the optimal learning rate is 
	\begin{equation} \label{LR-l2-Optimal}
	\eta^{\textbf{k}_{(s,\ell-1)}}_\text{opt}
	= \frac{
		\left\|\frac{\partial \mathcal{L}}{\partial \bm{W}_{\mathfrak{i}(\ell)}}\big|_{\bm{\theta}=\bm{\theta}^{\textbf{k}_{(s,\ell-1)}}}\right\|_F^2}{\left\|\bm{W}_{L:(\mathfrak{i}(\ell)+1)}^{\textbf{k}_{(s,\ell-1)}}\frac{\partial \mathcal{L}}{\partial \bm{W}_{\mathfrak{i}(\ell)}}\big|_{\bm{\theta}=\bm{\theta}^{\textbf{k}_{(s,\ell-1)}}}\bm{W}_{(\mathfrak{i}(\ell)-1):1}^{\textbf{k}_{(s,\ell-1)}}\bm{X}\right\|_F^2},
	\end{equation}
	and  with the optimal learning rate of \eqref{LR-l2-Optimal}, we obtain
	\begin{align*}
	\mathcal{L}(\bm{W}^{\textbf{k}_{k}})
	&=\mathcal{L}(\bm{W}^{\textbf{k}_{0}}) 
	-
	\sum_{s=0}^{k-1}\sum_{\ell=1}^{L}
	\frac{
		\left\|\frac{\partial \mathcal{L}}{\partial \bm{W}_{\mathfrak{i}(\ell)}}\big|_{\bm{\theta}=\bm{\theta}^{\textbf{k}_{(s,\ell-1)}}}\right\|_F^4}{\left\|\bm{W}_{L:(\mathfrak{i}(\ell)+1)}^{\textbf{k}_{(s,\ell-1)}}\frac{\partial \mathcal{L}}{\partial \bm{W}_\ell}\big|_{\bm{\theta}=\bm{\theta}^{\textbf{k}_{(s,\ell-1)}}}\bm{W}_{(\mathfrak{i}(\ell)-1):1}^{\textbf{k}_{(s,\ell-1)}}\bm{X}\right\|_F^2}.
	\end{align*}
\end{theorem}
\begin{proof}
    The proof can be found in \ref{app:thm:convg-l2}.
\end{proof}

The optimality of \eqref{LR-l2-Optimal}
should be understood 
in the sense that 
it gives the smallest loss
for the next iterate.

The assumption of all columns of $\bm{W}_L^{(0)}$ being in $ \text{range}(\bm{Y}\bm{X}^\dagger) \subset K$
is automatically satisfied if $n_{L-1} \ge n_L$ and
$\bm{W}_L^{(0)}$ is a full rank matrix.
Also, since $\text{range}(\bm{W}^{(0)}_L)$ affects the rate of convergence 
through $\kappa_{r}(\bm{W}_{L:(\mathfrak{i}(\ell)+1)}^{\textbf{k}_{(k,\ell-1)}})$,
a faster convergence is expected 
if $\text{range}(\bm{W}^{(0)}_L) = \text{range}(\bm{Y}\bm{X}^\dagger)$.
If $n_L > n_{L-1} \ge n_0$, 
the choice of $\bm{W}^{(0)}_L \approxeq \bm{Q}$
satisfies this, where 
$\bm{Q}$ is orthogonal and $\text{range}(\bm{Q}) = \text{range}(\bm{Y}\bm{X}^T)$.
We remark that in many practical applications, the number of training data is typically larger than both the input and the output dimensions, i.e., $m > \max\{n_0, n_L\}$. Also, the input dimension is greater than the output dimension, i.e., $n_0 > n_L$.
For example, the MNIST handwritten digit dataset
contains $60,000$ training data whose input and output dimensions are $784$
and $10$, respectively.

Theorem~\ref{thm:convg-l2} indicates 
that 
as long as $n_\ell \ge \min\{r_x, r\}$, the approximation error is strictly decreasing after a single sweep of BCGD
if  
either
$\kappa_{r}^2(\bm{W}_{L:(\mathfrak{i}(1)+1)}^{\textbf{k}_{(k,0)}})$
or
$\kappa_{r_x}^2(\bm{W}_{(\mathfrak{i}(\ell)-1):1}^{\textbf{k}_{(s,\ell-1)}}\bm{X})$
is positive.
Also, our analysis shows 
the ineffectiveness of training a network which has a layer whose width is less than 
$\max\{r_x, r\}$.
This is because 
if $n_\ell < \max\{r_x, r\}$, 
either $\sigma_{r}(\bm{W}_{L:(\mathfrak{i}(\ell)+1)}^{\textbf{k}_{(s,\ell-1)}})$
or
$\sigma_{r_x}(\bm{W}_{(\mathfrak{i}(\ell)-1):1}^{\textbf{k}_{(s,\ell-1)}}\bm{X})$
is zero and thus, $\gamma^{\textbf{k}_{(k,\ell-1)}} = 1$.
This indicates that
in order for the faster convergence,
one should employ a network whose architecture satisfying $n_\ell \ge \max\{r_x, r\}$ for all $1 \le \ell < L$.
Also, if $\bm{W}_1^{(0)}$ 
is initialize in a way that all rows are in $\text{range}(\bm{X})$,
one can expect to find the least norm solution.


In order for an iteration of BCGD to strictly 
decrease the approximation error,
it is important to guarantee the condition of 
\begin{equation} \label{condition-non-singular}
	\sigma_{r}^2(\bm{W}_{L:(\mathfrak{i}(\ell)+1)}^{\textbf{k}_{(s,\ell-1)}})
	\sigma_{r_x}^2(\bm{W}_{(\mathfrak{i}(\ell)-1):1}^{\textbf{k}_{(s,\ell-1)}}\bm{X}) > 0.
\end{equation}
In what follows, we show that 
if the initial approximation error is sufficiently close to
the global optimum under the orth-identity initialization (Section~\ref{subsec:initialization}), 
the convergence to the global optimum is guaranteed at a linear rate by the layer-wise training (BCGD).
\begin{theorem} \label{thm-l2-identity}
    Under the same conditions of Theorem~\ref{thm:convg-l2},
    let $\bm{X}$ be a full-row rank matrix
    and $n_\ell \ge \max\{n_0, n_L\}$ for all $1 \le \ell < L$.
    Suppose the weight matrices are initialized from the orth-identity initialization (Section~\ref{subsec:initialization}) and
    the initial loss $\|\bm{W}^{\textbf{k}_{0}} - \bm{W}^*\|_F$
    is less than or equal to $\tilde{\sigma}_{\min}/c$, where  
    $\tilde{\sigma}_{\min} = \sigma_{\min}(\bm{W}^*\bm{X})/\|\bm{X}\|$, 
    \begin{equation} \label{def:c-min}
    c=1 +  \kappa^2(\bm{X})\left(\frac{1+\sqrt{1+4h(L)\tilde{\sigma}_{\min}/\kappa^2(\bm{X})}}{2h(L)\tilde{\sigma}_{\min}}\right), \qquad
    h(L) = \frac{LR_L(1-R_L)^{2L-2}}{(1+R_L)^{3L-1}},
    \end{equation}
    and $R_L=\frac{2}{(5L-3)+\sqrt{(5L-3)^2-4L}}$.
    Then, with the learning rates of \eqref{LR-l2-loss-exact},
    the $k$-th sweep of BCGD satisfies 
    \begin{align*}
        \mathcal{L}(\bm{W}^{\textbf{k}_k}) - \mathcal{L}(\bm{W}^*)
        \le 
        \left(\mathcal{L}(\bm{W}^{\textbf{k}_0}) - \mathcal{L}(\bm{W}^*)\right)
        (\gamma^{2L})^{k},
    \end{align*}
    where 
    $\gamma = 1 - \frac{\eta}{5\kappa^2(\bm{X})}$
    and $0 < \eta \le 1$.
\end{theorem}
\begin{proof}
By Lemma~\ref{lemma-min-sing-value}, the proof readily follows from Theorem~\ref{thm:convg-l2}.
\end{proof}

\begin{lemma} \label{lemma-min-sing-value}
    Under the same conditions of Theorem~\ref{thm-l2-identity},
    we have 
    \begin{align*}
        \gamma^{\textbf{k}_{(k,\ell-1)}} < 1 - \frac{\eta}{\kappa^2(\bm{X})}
        \left(\frac{1-R_L}{1+R_L}\right)^{2(L-1)} \le 
        \gamma = 1 - \frac{\eta}{5\kappa^2(\bm{X})}.
    \end{align*}
\end{lemma}
\begin{proof}
	The proof can be found in \ref{app:lemma-min-sing-value}.
\end{proof}

We remark that the rate of convergence for a single sweep is $\gamma^{2L}$.
When the speed of convergence is measured against the number of sweeps, this implies that the deeper the network is, the faster convergence is obtained.
Thus, if
the depth of a linear network is sufficiently large,
the global optimum can be reached (within machine accuracy) by the layer-wise training (BCGD) after updating each weight matrix only once.
Also, we note that the work of \cite{Arora2018convergenceDLN} also 
has a similar initialization condition.

Theorem~\ref{thm-l2-identity} relies on the assumption that 
the initial approximation is sufficiently close to the global optimum $\bm{W}^*\bm{X}$
in terms of $\bm{X}$, $\sigma_{\min}(\bm{W}^*\bm{X})$ and the depth $L$.
As a special case of $d_{\text{out}} = 1$,
a similar result can be obtained without this restriction.

\begin{theorem} \label{thm:l2-dout1}
    Under the same conditions of Theorem~\ref{thm:convg-l2},
    let $n_L = 1$, 
    $n_\ell \ge n_0$ for all $1 \le \ell < L$
    and $\bm{X}$ is a full-row rank matrix.
    Suppose 
    the weight matrices are initialized 
    from the orth-identity initialization (Section~\ref{subsec:initialization}),
    and 
    the global minimizer is not 
    $\bm{W}^* \ne \bm{W}^{\textbf{k}_{(0,\ell-1)}}\left(\bm{I}_{n_0} - \|\bm{X}\|^2(\bm{XX}^T)^{-1}/\eta\right)$ for all $1 \le \ell \le L$,
    and the depth $L$ is chosen to satisfy
    $$
    L \ge \frac{\log \left(\frac{\sigma_{\min}(\bm{W}^*\bm{X})}{c\|(\bm{W}^{\textbf{k}_0} - \bm{W}^*)\bm{X}\|_F}\right)}{\log(1-\eta/\kappa^2(\bm{X}))},
    $$
    where $c$ is defined in \eqref{def:c-min} and $0<\eta \le 1$.
    Then, the $k$-th sweep of descending BCGD 
    with the learning rate of \eqref{LR-l2-loss-exact}
    satisfies
    \begin{equation}
        \begin{split}
        \mathcal{L}(\bm{W}^{\textbf{k}_k}) - \mathcal{L}(\bm{W}^*)
        &<
        \left(\mathcal{L}(\bm{W}^{\textbf{k}_0}) - \mathcal{L}(\bm{W}^*)\right)
        \left(1 -\frac{\eta}{\kappa^2(\bm{X})} \right)^{2(L+k-1)}(\gamma^{2(L-1)})^{k-1},
        \end{split}
    \end{equation} 
    where $\gamma = 1 - \frac{\eta}{5\kappa^2(\bm{X})}$.
\end{theorem}
\begin{proof}
    The proof can be found in \ref{app:thm:l2-dout1}.
\end{proof}



\subsection{Convergence of BCGD for general convex loss function} \label{convg-gen-loss}
We present a general convergence analysis of the layer-wise training (BCGD) for convex differentiable loss functions.
For general loss functions, let $\bm{W}^*$ be the solution to 
$\min_{\bm{W}}\mathcal{L}(\bm{W})$.
For a matrix $A$, the matrix $L_{p,q}$ norm is defined by
$$
\|A\|_{p,q} = \left(\sum_{j=1}^n \left(\sum_{i=1}^m |a_{ij}|^p\right)^{q/p}\right)^{1/q}, \quad p, q \ge 1,
$$
and the max norm is $\|A\|_{\max} = \max_{i,j} |a_{ij}|$.

\begin{theorem} \label{thm:convg-convex}
	Suppose $\ell(z;b)$ is convex and twice differentiable (as a function of $z$), 
	and that its second derivative satisfies 
	$
	|\ell''(z;b)| \le \mathcal{C}(z).
	$
	If the learning rates satisfy
	\begin{equation} \label{LR-convex}
		0 < \eta_\ell^{\textbf{k}_{(k,\ell-1)}} \le \frac{1}{\|\mathcal{C}(\Delta^{\textbf{k}_{(k,\ell-1)}})\|_{\max} \|\bm{W}_{L:(\mathfrak{i}(\ell)+1)}^{\textbf{k}_{(k,\ell-1)}}\|^2\|\bm{W}_{(\mathfrak{i}(\ell)-1):1}^{\textbf{k}_{(k,\ell-1)}}\bm{X}\|^2},
	\end{equation}
	where $\mathcal{C}$ is applied element-wise
	and $\Delta^{\textbf{k}_{(k,\ell-1)}} = \bm{W}^{\textbf{k}_{(k,\ell-1)}}\bm{X} - \bm{Y}$,
	the $(Lk+\ell)$-th iteration of BCGD satisfies 
	\begin{align}
		\mathcal{L}(\bm{\theta}^{\textbf{k}_{(k,\ell)}})
		&\le 
		\mathcal{L}(\bm{\theta}^{\textbf{k}_{(k,\ell-1)}})
		-\frac{\eta_\ell^{\textbf{k}_{(k,\ell-1)}}}{2}
		\| \mathcal{J}^{\textbf{k}_{(k,\ell-1)}}\|_F^2,
	\end{align}
	where
	$
	\mathcal{J}^{\textbf{k}_{(k,\ell-1)}}=\frac{\partial \mathcal{L}(\bm{\theta}) }{\partial \bm{W}_{\mathfrak{i}(\ell)}}\big|_{\bm{\theta} = \bm{\theta}^{\textbf{k}_{(k,\ell-1)}}}=
	(\bm{W}_{(\mathfrak{i}(\ell)-1):1}^{\textbf{k}_{(k,\ell-1)}}\bm{X}\bm{J}^{\textbf{k}_{(k,\ell-1)}}\bm{W}_{L:(\mathfrak{i}(\ell)+1)}^{\textbf{k}_{(k,\ell-1)}})^T.
	$
	Furthermore,
	\begin{itemize}
		\item The (near) optimal learning rate is 
		\begin{equation} \label{LR-gen-Opt}
			\eta_\text{opt}^{\textbf{k}_{(k,\ell-1)}} = 
			\frac{\| \mathcal{J}^{\textbf{k}_{(k,\ell-1)}}\|_F^2}
			{\|\mathcal{C}(\Delta^{\textbf{k}_{(k,\ell-1)}})\|_{\max}
				\|\bm{W}_{L:(\ell-1)}^{(k+1)}
				\mathcal{J}^{\textbf{k}_{(k,\ell-1)}}
				\bm{W}_{(\ell-1):1}^{(k)}\bm{X}\|_F^2}.
		\end{equation}
		\item For each $\ell$, 
		$\lim_{k \to \infty} \eta_\ell^{\textbf{k}_{(k,\ell-1)}}\| \mathcal{J}^{\textbf{k}_{(k,\ell-1)}}\|_F^2 = 0$.
		\item  
		$\frac{1}{kL}\sum_{s=0}^{k-1}\sum_{\ell=1}^L
		\eta_\ell^{\textbf{k}_{(k,\ell-1)}}\| \mathcal{J}^{\textbf{k}_{(k,\ell-1)}}\|_F^2	= \mathcal{O}(\frac{1}{kL})$.
		\item If $0 < \inf_k \eta_\ell^{\textbf{k}_{(k,\ell-1)}}
		\le \sup_k \eta_\ell^{\textbf{k}_{(k,\ell-1)}} \le 1$,
		we have 
		$$\lim_{k \to \infty} \|\eta_\ell^{\textbf{k}_{(k,\ell-1)}} \mathcal{J}^{\textbf{k}_{(k,\ell-1)}}\|_F^2 = 0,
		\qquad \lim_{k \to \infty} \| \mathcal{J}^{\textbf{k}_{(k,\ell-1)}}\|_F^2 = 0.$$
		Therefore, 
		$\{\bm{W}_\ell^{(k)}\}_{\ell=1}^L \overset{k\to\infty}{\to} \{\hat{\bm{W}}_\ell\}_{\ell=1}^L$
		and 
		$\{\hat{\bm{W}}_\ell\}_{\ell=1}^L$ is a stationary point.
		If $\hat{\bm{W}}_{L:1}$ is a local minimum,
		then it is the global minimum.
	\end{itemize}
\end{theorem}
\begin{proof}
	The proof can be found in \ref{app:thm:convg-convex}.
\end{proof}

Theorem~\ref{thm:convg-convex} shows that as long as the learning rates satisfying \eqref{LR-convex} are bounded below away from 0 and above by 1 for all $k$ but finitely many,
the BCGD finds a stationary point at the rate of $\mathcal{O}(1/kL)$ where $k$ is the number of sweeps and $L$ is the depth of DLN.
Also,
since the loss $\ell$ is known a prior,
the (near) optimal learning rate can directly be applied in practice.
For example, when the $p$-norm is used for the loss, i.e., $\ell(z;b) = |z-b|^p/p$ where $1 < p < \infty$ and $p$ is even,
the (near) optimal learning rate is
\begin{equation} \label{LR-lp-Optimal}
	\eta_\text{opt}^{\textbf{k}_{(k,\ell-1)}} = \frac{\| \mathcal{J}^{\textbf{k}_{(k,\ell-1)}}\|_F^2}
	{(p-1)\|\Delta^{\textbf{k}_{(k,\ell-1)}}\|_{\max}^{p-2}
		\|\bm{W}_{L:(\ell-1)}^{(k+1)}
		\mathcal{J}^{\textbf{k}_{(k,\ell-1)}}
		\bm{W}_{(\ell-1):1}^{(k)}\bm{X}\|_F^2}.
\end{equation}
Note that when $p=2$, the above is identical to the optimal learning rate of \eqref{LR-l2-Optimal}.

\subsection{Convergence of BCSGD}
In this subsection, a convergence analysis of BCSGD \eqref{def:bcsgd} is presented
with the standard $L_2$-loss.

We first describe the block coordinate stochastic gradient descent (BCSGD) as follow.
At the $(Lk+\ell)$-th iteration,
an index $i_{Lk+\ell}$ is randomly chosen from $\{1,\cdots,m\}$
and the $\mathfrak{i}(\ell)$-th layer weight matrix is updated 
according to 
\begin{equation} \label{def:bcsgd} 
	\begin{split}
	\bm{W}_{\mathfrak{i}(\ell)}^{(k+1)} = 
	\bm{W}_{\mathfrak{i}(\ell)}^{(k)}
	- \eta_{\mathfrak{i}(\ell)}^{\textbf{k}_{(k,\ell-1)}} \frac{\partial \mathcal{L}_{i_{Lk+\ell}}(\bm{\theta}) }{\partial \bm{W}_{\mathfrak{i}(\ell)}}\bigg|_{\bm{\theta} = \bm{\theta}^{\textbf{k}_{(k,\ell-1)}}},
	\end{split}
\end{equation}
where
$\textbf{k}_{(k,\ell)} = \textbf{k}_{(k,\ell-1)} + \bm{e}_{\mathfrak{i}(\ell)}$.
Again,  
$\mathfrak{i}(\ell) = \ell$ if the ascending (bottom to top) ordering is employed
and $\mathfrak{i}(\ell) = L-\ell+1$ if the descending (top to bottom) ordering is employed.

Given a discrete random variable $i \sim \bm{\pi}$ on $[m]$,
we denote the expectation with respect to $i$ 
conditioned on all other previous random variables
by $\mathbf{E}_{i}$.
%

\begin{theorem} \label{thm:convg-l2-loss-BCSGD}
    Let $\{\bm{W}_\ell^{(0)}\}_{\ell=1}^L$ be the initial weight matrices.
    At the $(Lk+\ell)$-th iteration,
    a data point $x_{i_{Lk+\ell}}$ is randomly independently chosen
    where $i_{Lk+\ell}$ is a random variable whose probability distribution  $\bm{\pi}^{\textbf{k}_{(k,\ell-1)}}$ is defined by
    \begin{equation}
    \bm{\pi}^{\textbf{k}_{(k,\ell-1)}}(i) = \frac{\|(\bm{W}_{(\mathfrak{i}(\ell)-1):1}^{\textbf{k}_{(k,\ell-1)}}x_{i})^T\bm{W}_{(\mathfrak{i}(\ell)-1):1}^{\textbf{k}_{(k,\ell-1)}}\bm{X}\|^2}{\|\bm{W}_{(\mathfrak{i}(\ell)-1):1}^{\textbf{k}_{(k,\ell-1)}}\bm{X}\|_F^4}, \qquad
    1 \le i \le m.
    \end{equation}
    Then, the approximation by BCSGD \eqref{def:bcsgd} with the learning rates of
    \begin{equation}
        \eta_{{i_{Lk+\ell}}}^{\textbf{k}_{(k,\ell-1)}} =  \frac{\sigma_{\min}^2(\bm{W}_{(\mathfrak{i}(\ell)-1):1}^{\textbf{k}_{(k,\ell-1)}}\bm{X})}{\sigma^2_{\max}(\bm{W}_{L:(\mathfrak{i}(\ell)+1)}^{\textbf{k}_{(k,\ell-1)}})}
        \frac{\eta}{\|(\bm{W}_{(\mathfrak{i}(\ell)-1):1}^{\textbf{k}_{(k,\ell-1)}}x_{{i_{Lk+\ell}}})^T\bm{W}_{(\mathfrak{i}(\ell)-1):1}^{\textbf{k}_{(k,\ell-1)}}\bm{X}\|^2},
    \end{equation}
    for $0 < \eta < 2$, 
    satisfies 
    \begin{align*}
    \mathbf{E}_{{i_{Lk+\ell}}}[\|{\Delta}^{\textbf{k}_{(k,\ell)}}\|_F^2]
    &\le
    \gamma_{\text{upp}}^{\textbf{k}_{(k,\ell-1)}}\|{\Delta}^{\textbf{k}_{(k,\ell-1)}}\|_F^2
    +
    \frac{\eta^2\mathcal{L}(\bm{W}^*)}{\tilde{\kappa}^4(\bm{W}_{(\mathfrak{i}(\ell)-1):1}^{\textbf{k}_{(k,\ell-1)}}\bm{X})},
	\\
    \mathbf{E}_{{i_{Lk+\ell}}}[\|{\Delta}^{\textbf{k}_{(k,\ell)}}\|_F^2]
    &\ge
    \gamma_{\text{low}}^{\textbf{k}_{(k,\ell-1)}} \|\Delta^{\textbf{k}_{(k,\ell-1)}}\|^2_F
    +
    \frac{\eta^2\mathcal{L}(\bm{W}^*)}{\kappa^4(\bm{W}_{L:(\mathfrak{i}(\ell)+1)}^{\textbf{k}_{(k,\ell-1)}})\tilde{\kappa}^4(\bm{W}_{(\mathfrak{i}(\ell)-1):1}^{\textbf{k}_{(k,\ell-1)}}\bm{X})},
    \end{align*}
    where 
    $\bm{W}^* = \bm{Y}\bm{X}^\dagger$, 
    ${\Delta}^{\textbf{k}_{(k,\ell)}} = \bm{W}_{L:1}^{\textbf{k}_{(k,\ell)}}\bm{X} - \bm{W}^*\bm{X}$, 
    \begin{align*}
    \gamma_{\text{upp}}^{\textbf{k}_{(k,\ell-1)}} &=1 - \frac{1 - \left(1-\frac{\eta}{\kappa^2(\bm{W}_{L:(\mathfrak{i}(\ell)+1)}^{\textbf{k}_{k}})}\right)^2 }{\tilde{\kappa}^4(\bm{W}_{(\mathfrak{i}(\ell)-1):1}^{\textbf{k}_{k}}\bm{X})}, 
    \\
    \gamma_{\text{low}}^{\textbf{k}_{(k,\ell-1)}} 
    &=
    1 - \frac{1 - \left(1 - \frac{\eta}{\kappa^2(\bm{W}_{(\mathfrak{i}(\ell)-1):1}^{\textbf{k}_{k}}\bm{X})}\right)^2 }{\tilde{\kappa}^4(\bm{W}_{(\mathfrak{i}(\ell)-1):1}^{\textbf{k}_{k}}\bm{X})/\kappa^4(\bm{W}_{(\mathfrak{i}(\ell)-1):1}^{\textbf{k}_{k}}\bm{X})}.
    \end{align*}
\end{theorem}
\begin{proof}
    The proof can be found in \ref{app:thm:convg-l2-loss-BCSGD}.
\end{proof}

Under the assumption that $\kappa^4(\bm{W}_{L:(\mathfrak{i}(\ell)+1)}^{\textbf{k}_{(k,\ell-1)}})\tilde{\kappa}^4(\bm{W}_{(\mathfrak{i}(\ell)-1):1}^{\textbf{k}_{(k,\ell-1)}}\bm{X})$ uniformly bounded above by $M_{\text{upp}}$
and $\gamma_{\text{low}}^{\textbf{k}_{(k,\ell-1)}}$ is uniformly bounded below away from zero by $\gamma_{\text{low}} > 0$, 
one can conclude that 
$$
\mathbf{E}[\|{\Delta}^{\textbf{k}_{k}}\|_F^2]
\ge \gamma_{\text{low}}^{Lk} \|{\Delta}^{\textbf{k}_{0}}\|_F^2 + \frac{\eta^2 \mathcal{L}(\bm{W}^*)(1 - \gamma_{\text{low}}^{Lk})}{M_{\text{upp}}(1-\gamma_{\text{low}}^L)} \to \frac{\eta^2 \mathcal{L}(\bm{W}^*)}{M_{\text{upp}}(1-\gamma_{\text{low}}^L)} \quad \text{as } \quad k \to \infty.
$$
Similarly, under the assumption that $\tilde{\kappa}^4(\bm{W}_{(\mathfrak{i}(\ell)-1):1}^{\textbf{k}_{(k,\ell-1)}}\bm{X})$ uniformly bounded below by $M_{\text{low}}$,
and $\gamma_{\text{upp}}^{\textbf{k}_{(k,\ell-1)}}$ is uniformly bounded above by $\gamma_{\text{upp}} < 1$, 
we have
$$
\mathbf{E}[\|{\Delta}^{\textbf{k}_{k}}\|_F^2]
\le \gamma_{\text{upp}}^{Lk} \|{\Delta}^{\textbf{k}_{0}}\|_F^2 + \frac{\eta^2 \mathcal{L}(\bm{W}^*)(1 - \gamma_{\text{upp}}^{Lk})}{M_{\text{low}}(1-\gamma_{\text{upp}}^L)} \to \frac{\eta^2 \mathcal{L}(\bm{W}^*)}{M_{\text{low}}(1-\gamma_{\text{upp}}^L)} \quad \text{as } \quad k \to \infty.
$$
This indicates that unlike the BCGD,
if a randomly chosen datum is used to update a weight matrix,
an extra term, which is proportional to $\mathcal{L}(\bm{W}^*)$, is introduced
in both upper and lower bounds of the expected error.
Therefore, the BCSGD would not achieve the global optimum, unless $\mathcal{L}(\bm{W}^*) = 0$.
However, the expected loss by BCSGD will be within the distance proportional to $\mathcal{L}(\bm{W}^*)$ from $\mathcal{L}(\bm{W}^*)$.
In practice, $\mathcal{L}(\bm{W}^*)$ will almost never be zero. 
This indicates that the stochasticity introduced by the random selection of mini-batch (of size 1) results in an implicit regularization effect, which avoids over-fitting.
We defer further characterization of BCSGD to future work. 

Remark: The proposed stochastic gradient-descent in Theorem~\ref{thm:convg-l2-loss-BCSGD} can be viewed as a generalized version of the sampling used in \cite{strohmer2009randomized,needell2010randomized,leventhal2010randomized,zouzias2013randomized}.

\section{Numerical Examples}
\label{sec:example}

We provide numerical examples to demonstrate the performance of layer-wise training by BCGD and justify our theoretical findings.
We employ three different initialization schemes, described in Section~\ref{subsec:initialization}.
In all examples, the network architectures are met the condition of $n_\ell \ge \max\{d_\text{in},d_\text{out}\}$ unless otherwise stated. According to Theorem~\ref{thm:role of width}, when either the orth-indentity or the balanced initialization is employed, we simply set $n_\ell = \max\{n_0,n_L\}$ for all $1\le \ell < L$. 
The approximation error is measured by the normalized distance to the global optimum, i.e.,
$\frac{1}{m}\mathcal{L}(\bm{W}^{\textbf{k}_k}) - \frac{1}{m}\mathcal{L}(\bm{W}^*)$.
When the $L_2$-loss is employed, the error after the $k$-th sweep is $\frac{1}{m}\left[\|\bm{W}^{(k)}\bm{X} - \bm{Y}\|_F^2 - \|\bm{W}^*\bm{X}-\bm{Y}\|_F^2\right]$.
For the convenience of visualization, 
if the error is less than $10^{-10}$, 
we simply set $10^{-10}$.
We note that the speed of convergence can be measured by either the number of sweeps
or the number of iterations.
Note also that updating each weight matrix once in a deep network
will require more time than doing so in a shallow network. 

In what follows, we employ the layer-wise training by BCGD 
for deep linear neural networks. 
The learning rate is chosen to be (near) optimal according to \eqref{LR-gen-Opt}.
We emphasize that the (near) optimal learning rate of \eqref{LR-gen-Opt}
does not require any prior knowledge, and can completely be determined by the loss function, the current weight matrices and the input data matrix. 
This allows us to avoid a cumbersome grid-search over learning rate. 
When the $L_2$-loss is employed, the optimal learning rate of \eqref{LR-l2-Optimal}
is identical to the one of \eqref{LR-gen-Opt}. 

\subsection{Random Data Experiments} \label{subsec:Random}
Unless otherwise stated, we generate the input data matrix $\bm{X} \in \mathbb{R}^{d_\text{in}\times m}$ whose entries are i.i.d. samples from 
a Gaussian distribution $N(0,1/n_0)$
and the output data matrix $\bm{Y} \in \mathbb{R}^{d_\text{out}\times m}$ whose entries are i.i.d. samples from a uniform distribution on $(-1,2)$.
The number of training data is set to $m = 600$.

\subsubsection{Small Condition number} \label{subsec:smallC}
\begin{figure}[htbp]
	\centerline{
		\includegraphics[height=4.8cm]{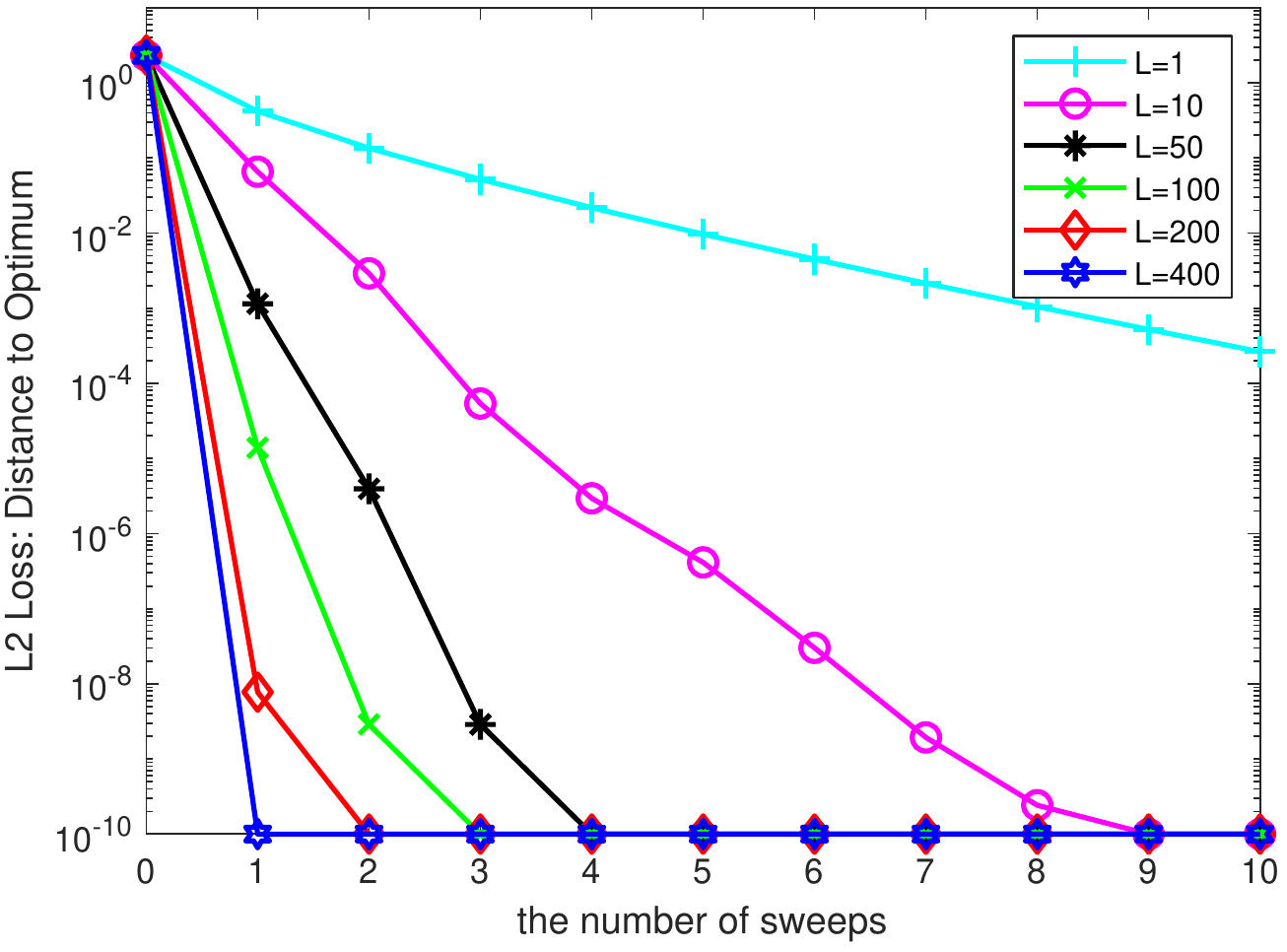}
		\includegraphics[height=4.8cm]{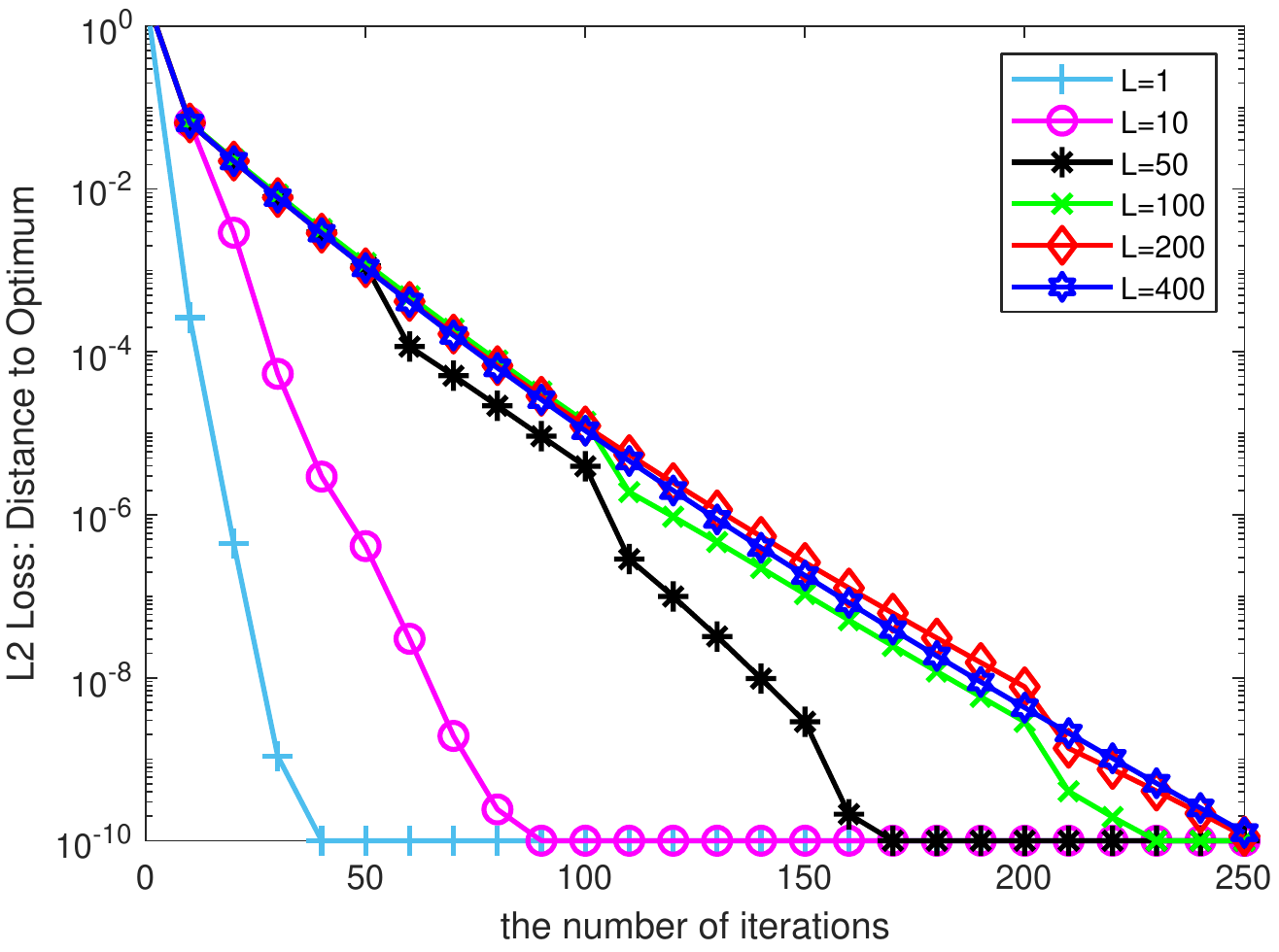}}
	\caption{The approximation errors with respect to the number of (left) sweeps and (right) iterations of the descending BCGD with the optimal learning rate \eqref{LR-l2-Optimal} at different depths
		$L=1, 10, 50, 100, 200, 400$.
		The width is set to $\max\{n_0,n_L\} = 128$
		and the orth-identity initialization is employed. 
		When the depth is 400, the global optimum is achieved by
		after updating each weight matrix only once.
		However, when the errors are compared against the number of iterations, 
		updating a single layer $L$ times results in
		the faster loss decay 
		than updating a $L$ layer network once.
	}
	\label{fig:OrhtInit-Cond2}
\end{figure}

On the left of Figure~\ref{fig:OrhtInit-Cond2}, the approximation errors are plotted with respect to the number of sweeps of the descending BCGD at different depths $L$.
The input and output dimensions are 
$d_{\text{in}} = n_0 = 128$ and $d_\text{out} = n_L = 10$, respectively.
The width of the $\ell$-th layer is $n_\ell = 128 = \max\{n_0,n_L\}$
and the orth-identity initialization (Section~\ref{subsec:initialization}) is employed. 
We see that the faster convergence is obtained as the depth grows.
In an extreme case of the depth $L=400$, the global optimum is achieved by only after updating each weight matrix once. 
These results are expected from Theorem~\ref{thm:convg-l2}.
To fairly compare the effects of depth in the acceleration of convergence, the approximation errors need to be plotted with respect to the number of iterations.
On the right of Figure~\ref{fig:OrhtInit-Cond2}, 
the errors are shown with respect to the number of iterations.
We now see that training a depth 1 network multiple times results in the fastest decrease in the loss.  This implies that in order for the faster convergence, it is better to train a depth 1 network $L$ times
than to train a depth $L$ network once in this case.
We remark that the condition number of the input data matrix was $2.6614$.
In this case, we do not have any advantages of using deep networks over a depth 1 network. 

\subsubsection{Big Condition number} \label{subsec:bigC}
\begin{figure}[htbp]
	\centerline{
		\includegraphics[height=4.8cm]{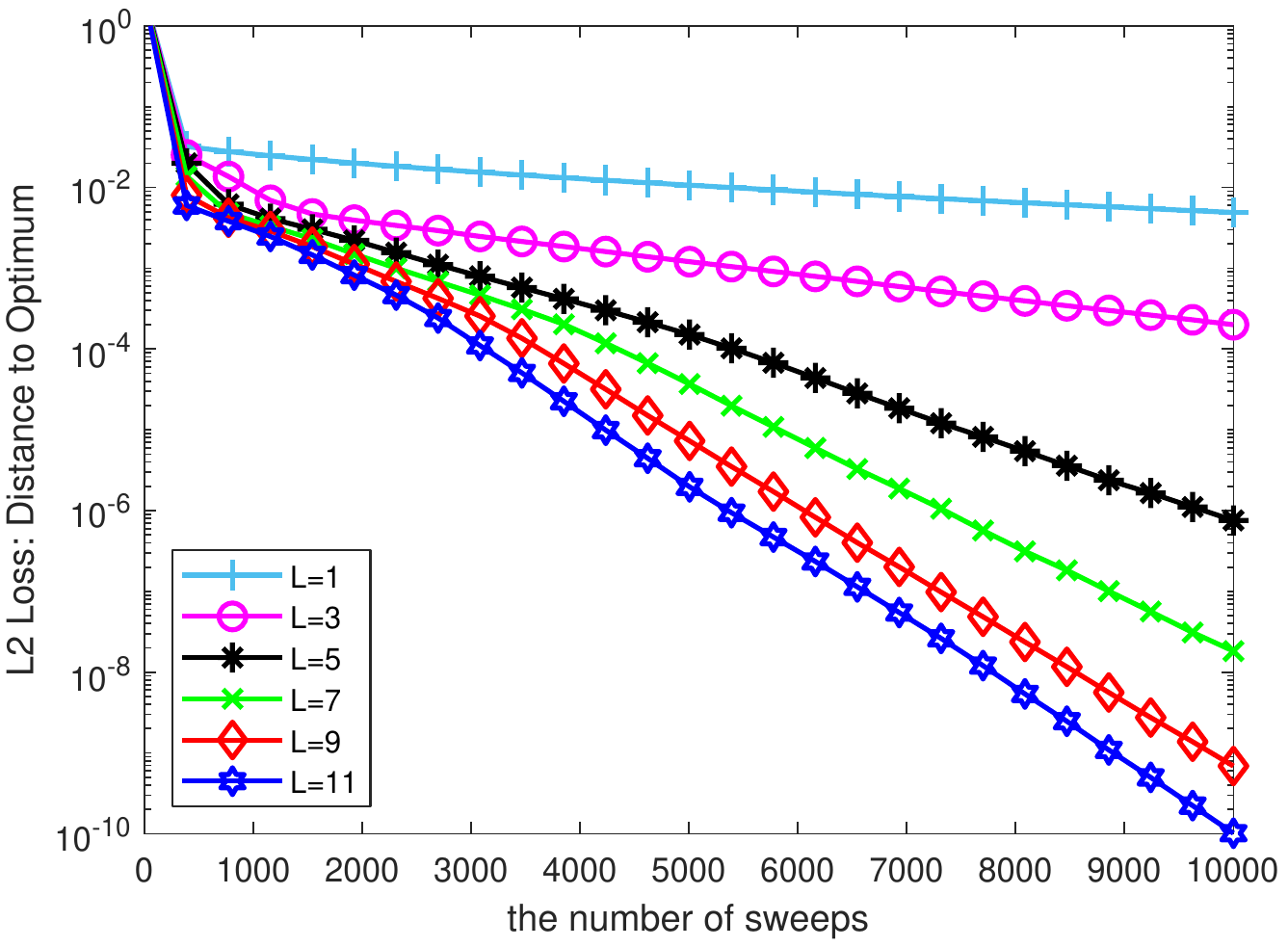}
		\includegraphics[height=4.8cm]{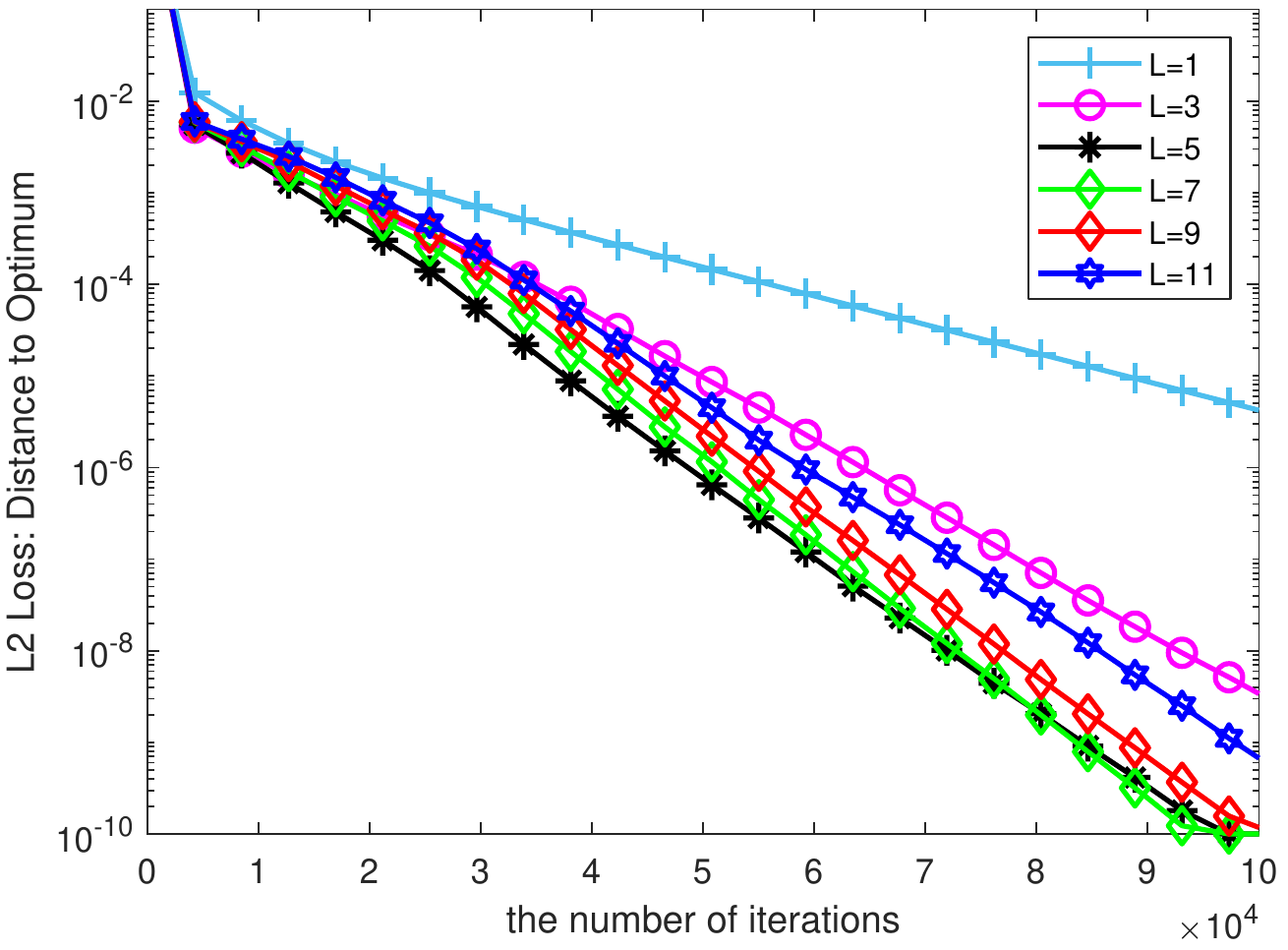}}
	\caption{The approximation errors with respect to the number of (left) sweeps and (right) iterations of the descending BCGD with the optimal learning rate \eqref{LR-l2-Optimal} at different depths.
		The width is set to $\max\{n_0,n_L\} = 128$
		and the orth-identity initialization is employed. 
		The condition number of the input data matrix is $236$.
		In terms of the number of sweeps,
		the deeper the network is, the faster convergence is obtained.
		In terms of the number of iterations (i.e., the computational cost is considered),
		unlike Figure~\ref{fig:OrhtInit-Cond2} where $\text{cond}(\bm{X}) \approx 2$, 
		the use of deep networks
		drastically accelerates convergence of the loss
		when it is compared to those by a depth 1 network.
	}
	\label{fig:OrhtInit-Cond200}
\end{figure}

We now consider the input data matrix $\bm{X}$ whose condition number is rather big.
To do this, we first generate $\bm{X}$ as in the above
and conduct the singular value decomposition.
We then assign randomly generated numbers from $10^{-5} + \mathcal{U}(0,1)$
to the singular values.
In our experiment, the condition number of $\bm{X}$ was 236.
The output data matrix $\bm{Y}$ is generated in the same way as before.
In Figure~\ref{fig:OrhtInit-Cond200}, 
the approximation errors are plotted with respect to the number of (left) sweeps and (right) iterations of the descending BCGD at different depths $L=1,3,5,7,9,11$.
When the speed of convergence is measured against the number of sweeps, we see that the deeper the network is, the faster the convergence
is obtained. 
When the amount of computation is considered, unlike the case where $\bm{X}$ has a good condition number,
we now see that the errors by deep linear networks 
decay drastically faster than those by a shallow network of depth 1.
This demonstrates that over-parameterization by the depth 
can indeed accelerate convergence, even when the computational cost
is considered.
We note that from Theorem~\ref{thm:role of width},  the width plays no role in gradient-based training, as the width of intermediate layers is $\max\{d_\text{in},d_\text{out}\}$. 
Furthermore, the optimal learning rate is employed and adding more layers does not increase any representational power.
Therefore, this acceleration is solely contributed by the depth
and this clearly demonstrates the benefit of using deep networks.
We also observe that 
the error decrease per iteration does not grow proportionally to the depth. In this case, either depth 5 or 7 performs the best among others.

\subsubsection{Comparison with GD}
\begin{figure}[htbp]
	\centerline{
		\includegraphics[height=4.8cm]{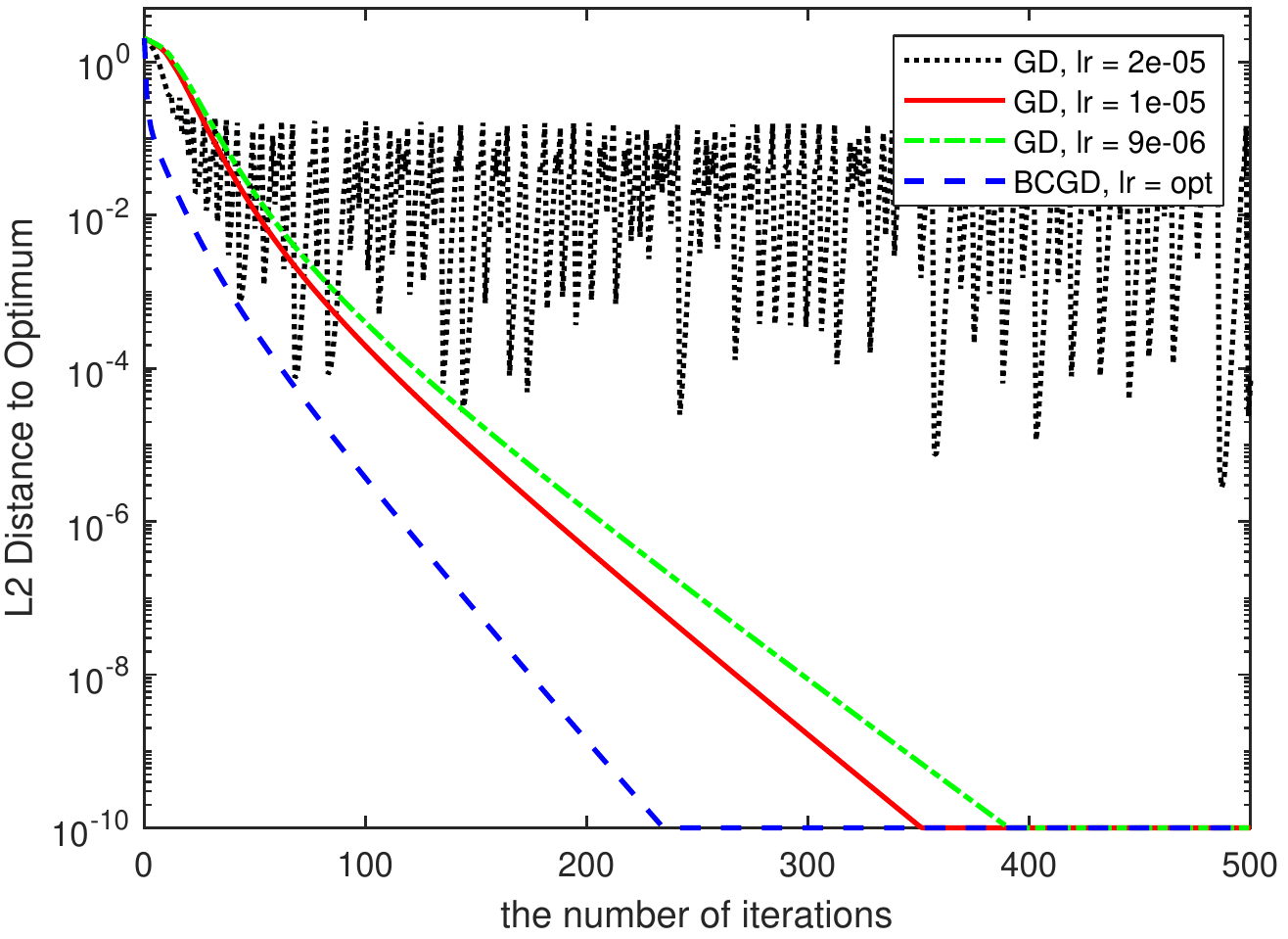}
		\includegraphics[height=4.8cm]{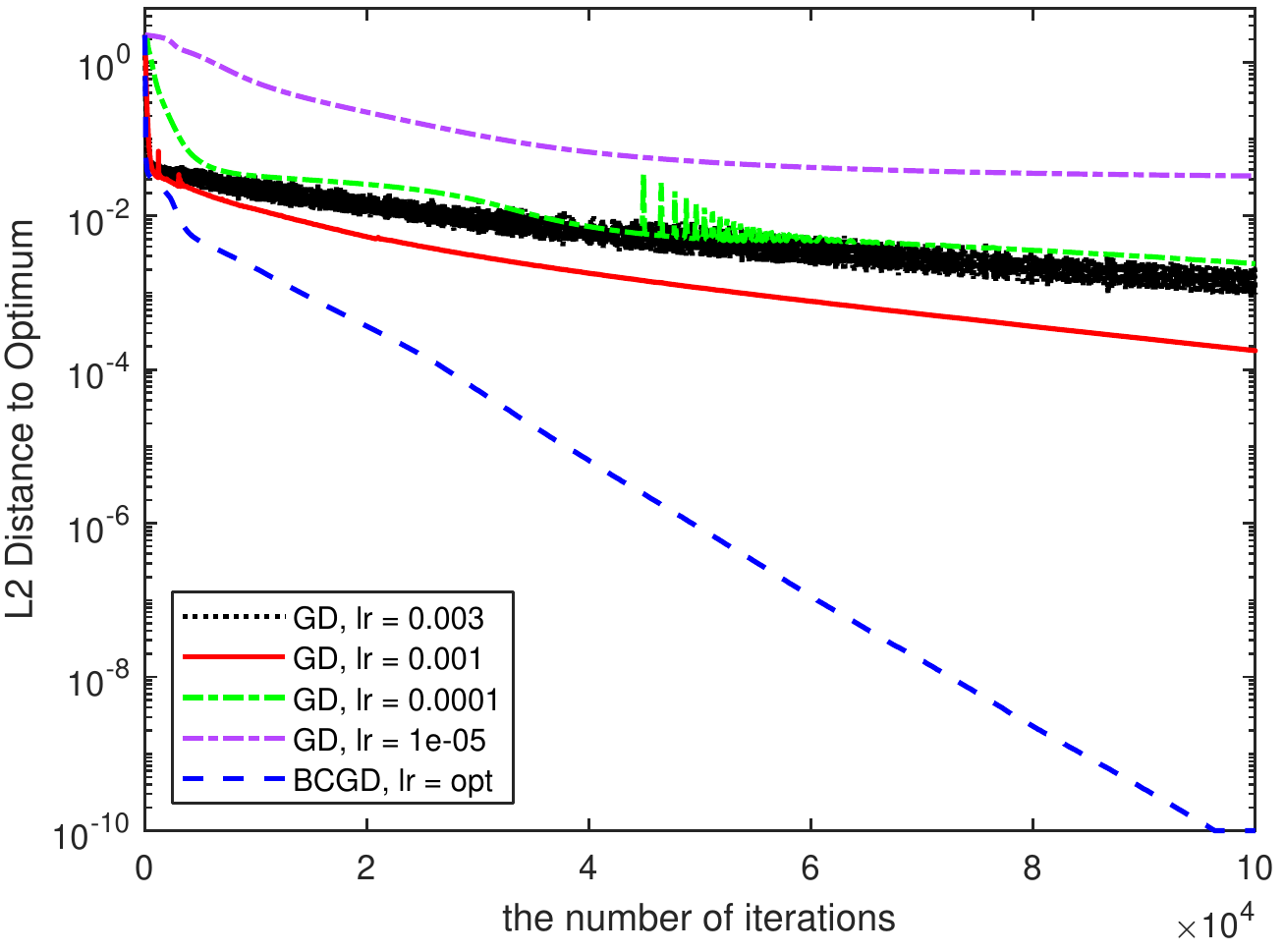}
		}
	\caption{The approximation errors by BCGD and GD with respect to the number of iterations.
	Learning rates for GD are found by trial-and-error.
	(Left) 
	The model matrix has a condition number of $2.6$ and
	the depth and width of DLN is 400 and 128, respectively.
	(Right) The model matrix has a condition number of $236$
	and the depth and width of DLN is 5 and 128, respectively.
	In all cases, the orth-identity initialization is employed. 
	We note that 
	BCGD updates only a single matrix per iteration,
	while GD updates all the weight matrices per iteration.
	}
	\label{fig:OrhtInit-Cond2-GD}
\end{figure}

Next, we compare the performance between
BCGD and the standard gradient descent (GD)
on the two same tasks of Section~\ref{subsec:smallC}
and ~\ref{subsec:bigC}.
By trial-and-error, we choose constant learning rates for GD that leads the fastest convergence.
We tried the learning rate of 
$\eta = \frac{n_L}{3L \|X\|^2}$ from \cite{Du2019width},
however, we observed that it makes GD diverge within few iterations.
Despite the fact that GD updates all the weight matrices
in a single iteration, while BCGD updates only a singe matrix, we compare the performance 
with respect to the number of iterations
to emphasize the performance of BCGD.
Figure~\ref{fig:OrhtInit-Cond2-GD}
shows the approximation errors by GD and BCGD.
The results for the task with a small (big) condition number are presented on the left (right).
In both cases, we observe that GD converges linearly 
when learning rate is chosen properly.
However, choosing an appropriate learning rate 
requires a time consuming fine tuning.
It is also clear that GD is highly sensitive 
with respect to learning rate.
For example, on the left of Figure~\ref{fig:OrhtInit-Cond2-GD},
we see that GD with the learning rate of $10^{-5}$
produces a linear convergence,
however,
GD with the learning rate of $2\times 10^{-5}$
leads a highly oscillatory behavior in the error.
When it is compared to the results by BCGD
and also considering the fact that BCGD updates only a single matrix per iteration, 
it is clear that BCGD converges significantly faster than GD, especially when the model matrix has a big condition number.
Furthermore, BCGD does not require one to put 
any efforts on finding a proper learning rate.
This clearly demonstrate superior performances of BCGD
over GD in these cases.

\subsubsection{Effect of Width}
\begin{figure}[htbp]
	\centerline{
	\includegraphics[height=4.8cm]{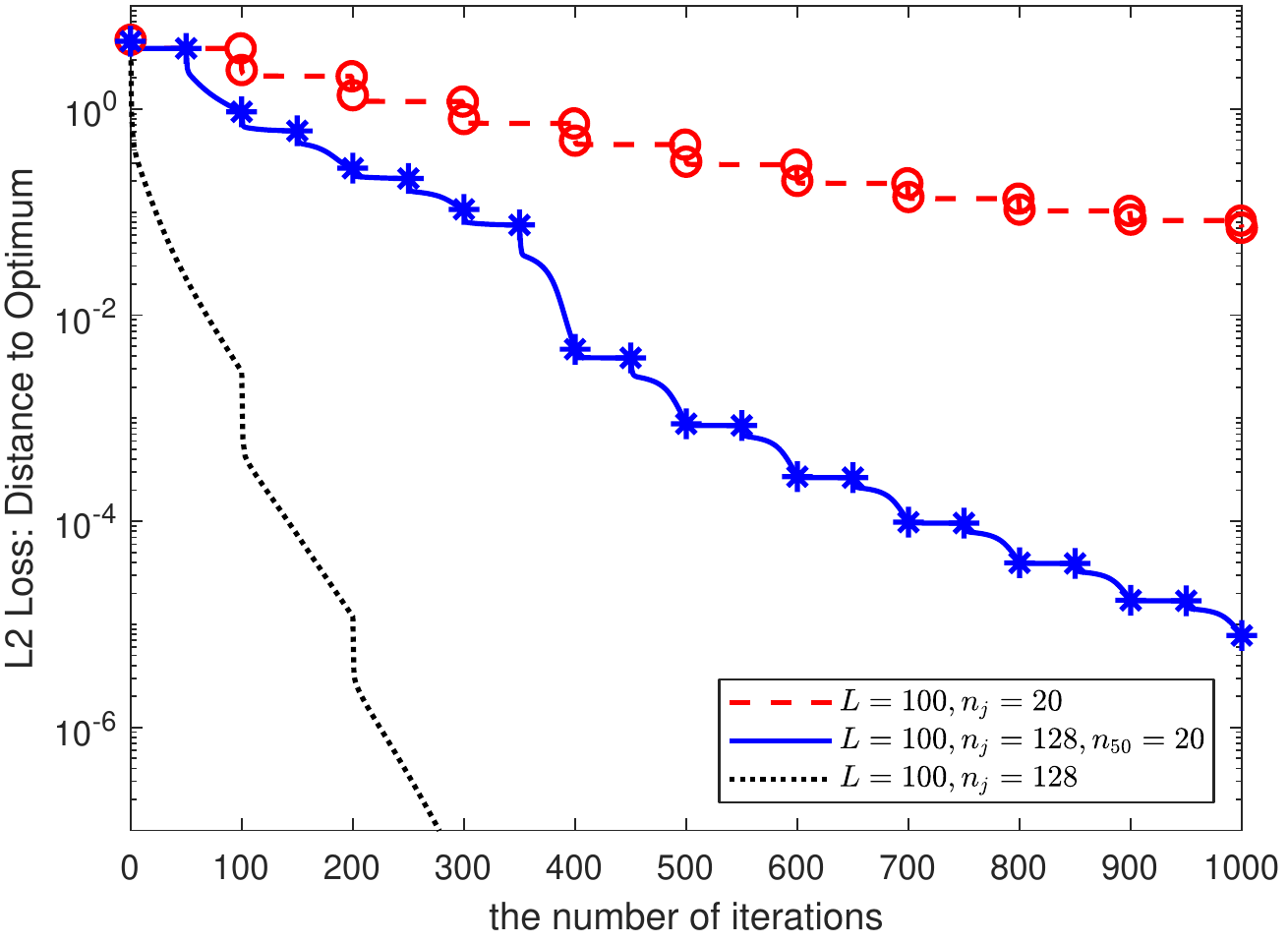}}
	\caption{
	The approximation errors with respect to the number of iterations of the descending BCGD
	by three different network architectures. 
	The results by the network of Arch 1 $(n_0=128,n_j=20)$
	are shown as the dash line,
	those by the network of Arch 2
	$(n_j=128, n_{50}=n_L=20)$
	are shown as the solid line,
	and those by the network of Arch 3 
	$(n_j=128, n_L=20)$
	are shown as the dotted line.
	This demonstrates the ineffectiveness of training a network which has a layer whose width is less than $\max\{n_0,n_L\}$.
	}
	\label{fig:ineffectiveness}
\end{figure}

From now on, the convergence speed is only measured against the number of iterations.
Next, we show the ineffectiveness of training a network which has a layer whose width is less than $\max\{d_\text{in},d_\text{out}\}$.
Figure~\ref{fig:ineffectiveness} shows the approximation errors
with respect to the number of iterations of the descending BCGD.
The input and output dimensions are 
$d_{\text{in}} = 128$ and $d_\text{out} = 20$, respectively.
Two deep linear networks of depth $L=100$ are compared. 
One has the architecture (Arch 1) of 
$n_\ell = 20$ for all $1 \le \ell < L$.
The other has the architecture (Arch 2) of $n_\ell = 128$ for all $1 \le \ell < L$, but $n_{50}=20$.
Note that at the $k$-th iteration where $k =L-\ell+1 \bmod L$,
the $(L-\ell+1)$-th layer weight matrix is the only matrix updated. 
For the network of Arch 1, we see that the errors decrease mostly only
after updating the first layer weight matrix.
The errors before and after updating the first layer 
are marked as the circle symbols ($\circ$).
For the network of Arch 2, we see that the errors decrease mostly
after updating from the 50th to the 1st layer weight matrices.
The errors before and after updating the 50th and the 1st layer matrices 
are marked as the asterisk symbols ($\ast$).
These are expected from Theorem~\ref{thm:convg-l2},
as 
either $\sigma_{\min}(\bm{W}_{L:(\mathfrak{i}(\ell)+1)}^{\textbf{k}_{(s,\ell-1)}})$
or
$\sigma_{\min}(\bm{W}_{(\mathfrak{i}(\ell)-1):1}^{\textbf{k}_{(s,\ell-1)}}\bm{X})$
is zero,
due to the network architecture.
Precisely, the Arch 1 results in $\sigma_{\min}(\bm{W}_{(L-\ell):1}^{\textbf{k}_{(s,\ell-1)}}\bm{X}) = 0$,
for all $s$ and $1 \le \ell < L$,
and the Arch 2 results in
$\sigma_{\min}(\bm{W}_{(L-\ell):1}^{\textbf{k}_{(s,\ell-1)}}\bm{X}) = 0$
for all $s$ and $1 \le \ell \le 50$.
For reference, the results by the network architecture (Arch 3) of $n_\ell = 128$ for all $\ell$ are shown as the dotted line. 
We see the fastest convergence by the network of Arch 3 among others.
This demonstrates the ineffectiveness of training a deep linear network which has a layer whose width is less than $\max\{n_0,n_L\}$.

\subsubsection{Ascending versus Descending}
\begin{figure}[htbp]
	\centerline{
	\includegraphics[height=4.8cm]{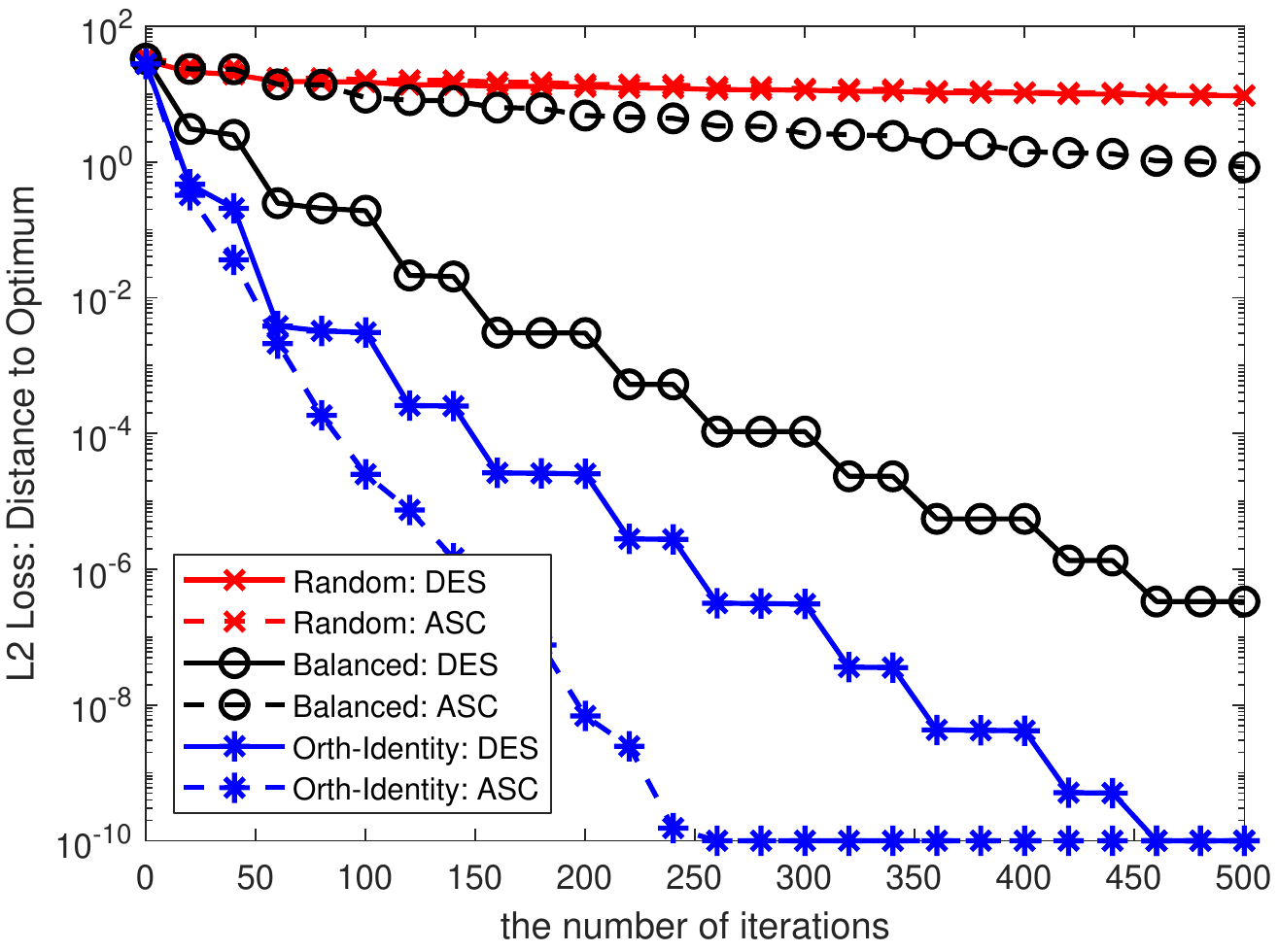}
	\includegraphics[height=4.8cm]{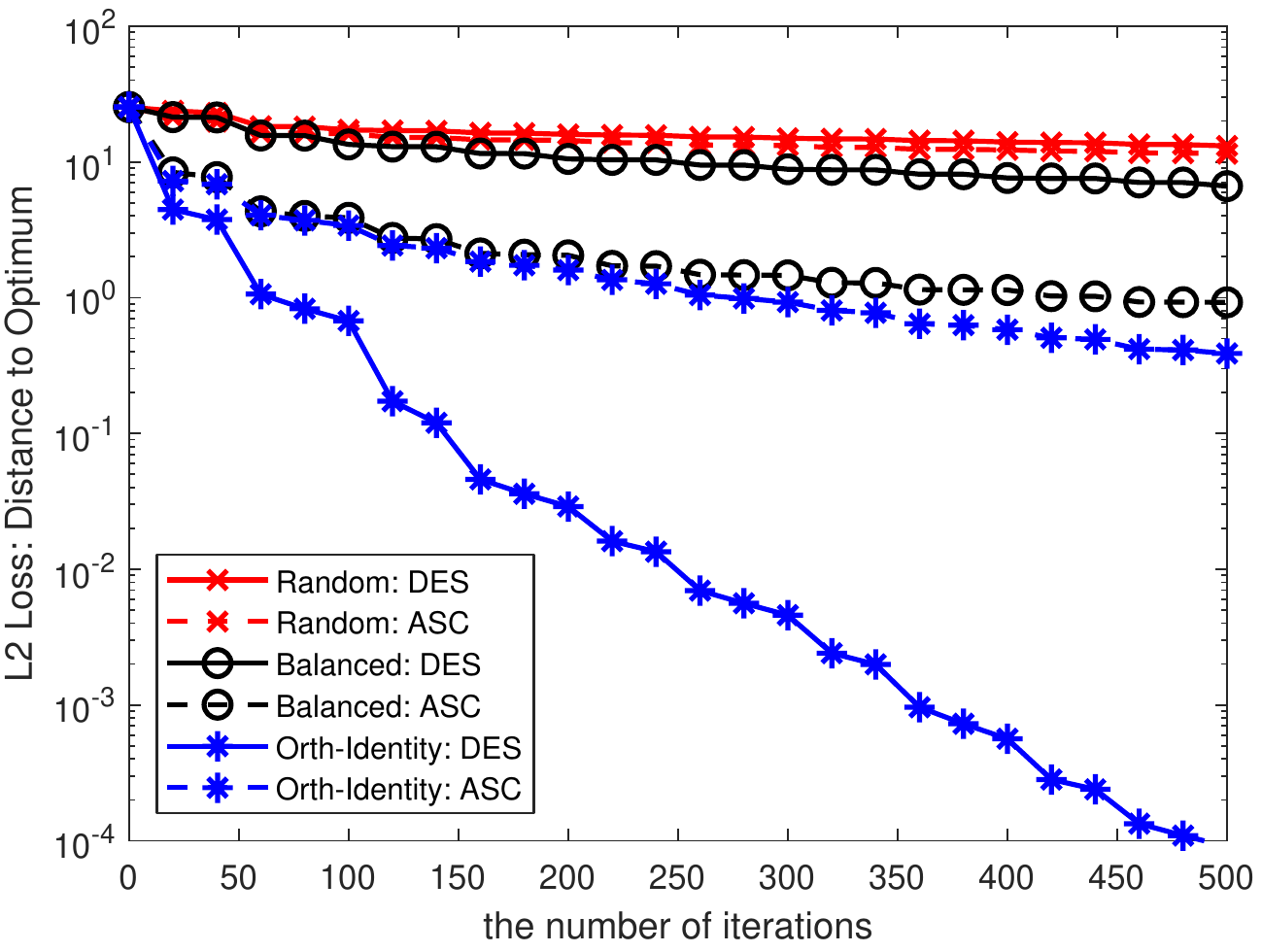}}
	\caption{The approximation errors with respect to the number of iterations of both the ascending and descending BCGD 
	by three different initialization schemes. The depth is $L=50$
	and the training is done over $600$ data points. 
	(Left) $n_0 = 50, n_j = 300$ for $0 < j \le L$.
	(Right) $n_j = 300, n_L = 50$ for $0\le j < L$.
	When $n_0=50, n_j=300$, the ascending BCGD with the orth-identity initialization results in the fastest convergence among others.
	When $n_j=300, n_L=50$, the descending BCGD with the orth-identity initialization results in the fastest convergence among others.
	}
	\label{fig:asc_vs_des}
\end{figure}

We now compare the performance of layer-wise training by BCGD
with two update orderings (top to bottom and bottom to top).
Figure~\ref{fig:asc_vs_des}
shows the approximation errors with respect to the number of iterations of both the ascending and descending BCGD
at three different initialization schemes (Section~\ref{subsec:initialization}).
We employ the DLNs of depth $L=50$  
and set the width of the $\ell$-th layer to $n_\ell = \max\{n_0,n_L\}$.
On the left, the input and output dimensions are  
$d_{\text{in}} = 50$ and $d_\text{out} = 300$, respectively.
It can be seen that 
for the orth-identity initialization,
the errors by the ascending BCGD
decay faster than those by the descending BCGD.
For the balanced initialization, 
the opposite is observed. 
For the random initialization, the errors by both the ascending and descending orderings behave similarly.
We see that the ascending BCGD with the orth-identity initialization
results in the fastest convergence among others.  
On the right, the input and output dimensions are  
$d_{\text{in}} =  300$ and $d_\text{out} = 50$, respectively.
It can be seen that 
for the balanced and the random initialization,
the errors by the ascending BCGD
decay faster than those by the descending BCGD.
For the orth-identity initialization, 
the opposite is observed. 
In this case, the descending BCGD with the orth-identity initialization
results in the fastest convergence among others.
In all cases, we observe that the orth-identity initialization outperforms than other initialization schemes, regardless of the update ordering.
Also, we found that when the orth-identity initialization is employed, 
the ascending BCGD performs better than the descending BCGD if the output dimension is larger than the input dimension,
and vice versa.

\subsection{Real Data Experiments}
\begin{figure}[htbp]
	\centerline{
		\includegraphics[height=4.8cm]{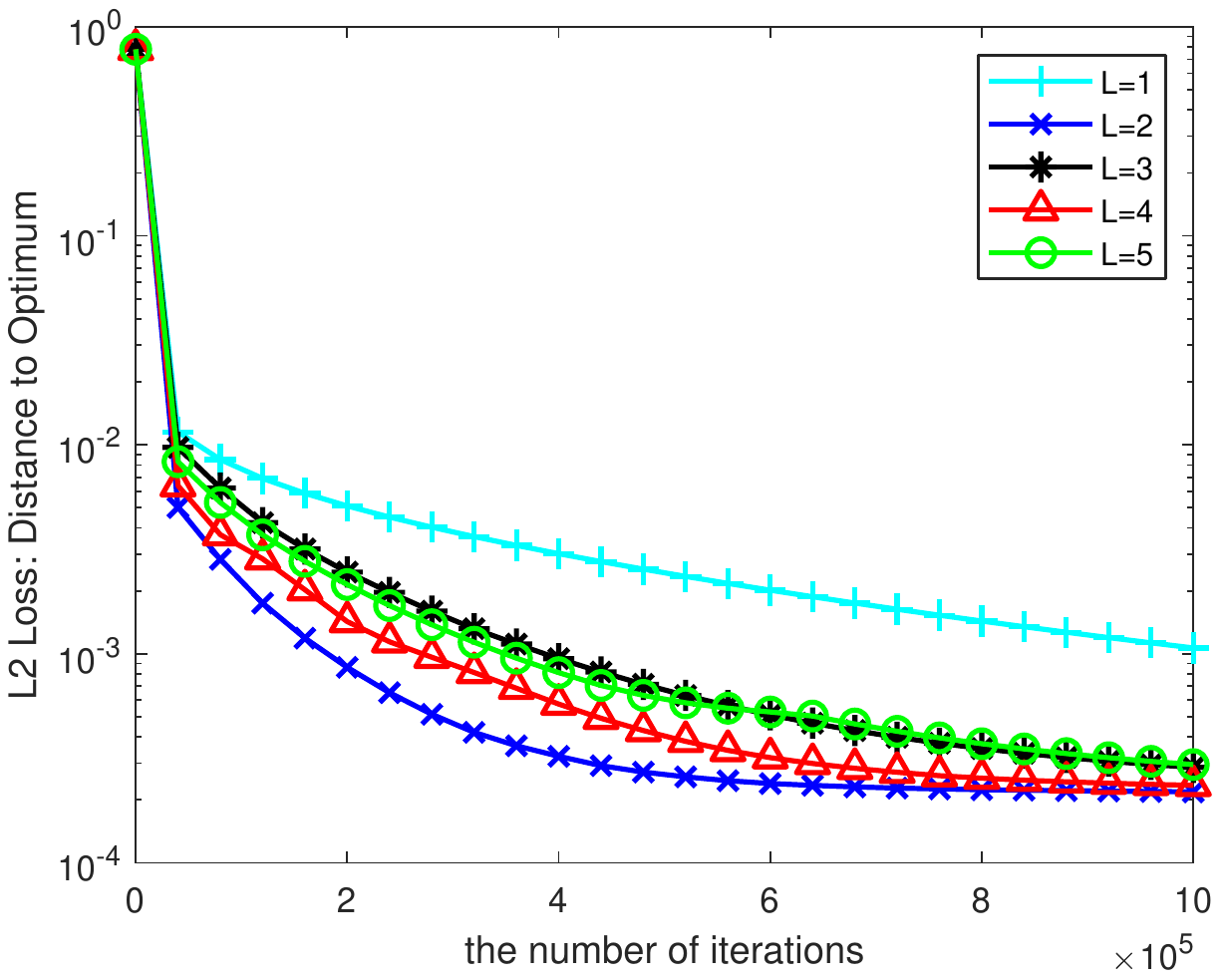}
		\includegraphics[height=4.8cm]{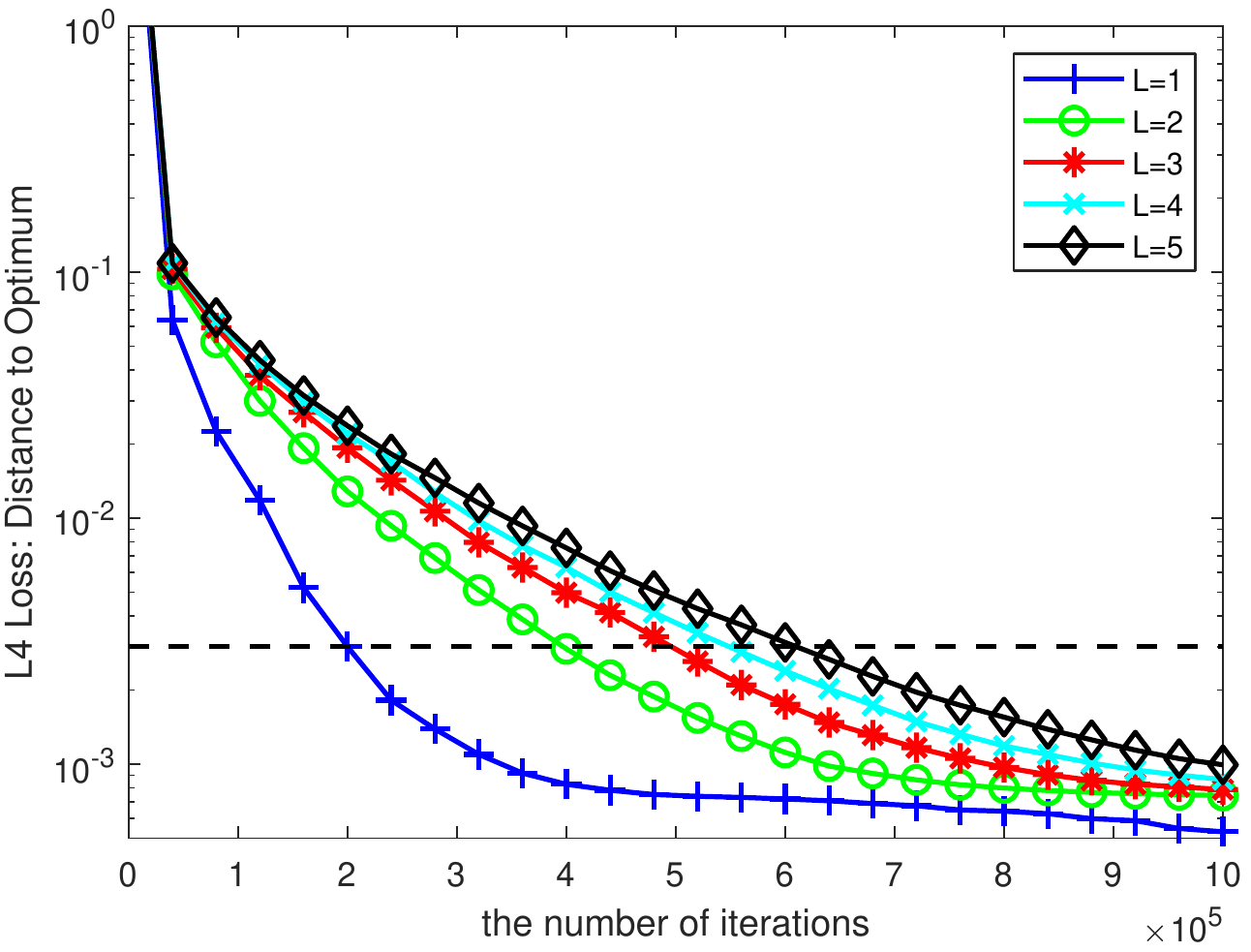}
	}
	\caption{The distances to the global optimum by (left) $L_2$-loss and (right) $L_4$-loss with respect to the number of iterations. 
		The network is trained over the UCI Machine Learning Repository's dataset of 2565 examples.
		The condition number of $\bm{X}$ is 70,980.
		The identity initialization is employed.
		The width is set to $n_\ell = 128$.
		In all depths, 
		the errors by deep linear networks
		decay faster than those by a single layer one.
	}
	\label{fig:UCI-data}
\end{figure}

We employ
the dataset from UCI Machine Learning Repository’s “Gas Sensor Array Drift at Different Concentrations”
\cite{Vergara2012Chemical,Rodriguez2014Calibration}.
Specifically, we used the dataset’s
“Ethanol” problem — a scalar regression task with 2565 examples, each comprising 128 features
(one of the largest numeric regression tasks in the repository). 
The input and output data sets are normalized to have zero mean and unit variance.
After the normalization, the condition number of the input data matrix is 70,980.
We note that this is the same data set used in \cite{Arora2018optAccelerationDLN}.
The width of intermediate layers is set to $\max\{d_\text{in},d_\text{out}\}$
and the identity initialization (Section~\ref{subsec:initialization})
is employed.
On the left of Figure~\ref{fig:UCI-data}, we show the errors by the descending BCGD with respect to the number of iterations at five different depths $L=1, 2, 3, 4, 5$.
We use the optimal learning rate \eqref{LR-l2-Optimal},
which does not require any prior knowledge.
We clearly see that
the over-parameterization by depth significantly accelerates convergence.
We remark that in the work of \cite{Arora2018optAccelerationDLN}, although a different optimization method is used, the same problem is considered and the learning rate is chosen by a grid search. 
Similar implicit acceleration was demonstrated only for
$L_4$-loss, not $L_2$-loss.
In our experiment, by exploiting the layer-wise training and the optimal learning rate,
we demonstrate implicit acceleration for $L_2$-loss.
On the right of Figure~\ref{fig:UCI-data}, we show the results by $L_4$-loss, i.e, 
\begin{align*}
\frac{1}{m}\left[\|\bm{W}^{(k)}\bm{X} - \bm{Y}\|_{4,4}^4 - \|\bm{W}^*\bm{X}-\bm{Y}\|_{4,4}^4\right].
\end{align*}
The near optimal learning rate of \eqref{LR-lp-Optimal} is employed.
We observe that updating a single layer multiple-times results in the fastest error convergence than updating multiple layers once.
In this case, there is no advantages of using deep networks.
For reference, we also plot the best error shown at \cite{Arora2018optAccelerationDLN} after 1,000,000 iterations
as the dashed line.
Unlike the conclusion of \cite{Arora2018optAccelerationDLN},
we found that the depth leads to acceleration for the $L_2$-loss,
but not for the $L_4$-loss.
%
%

\begin{figure}[htbp]
	\centerline{
		\includegraphics[height=4.8cm]{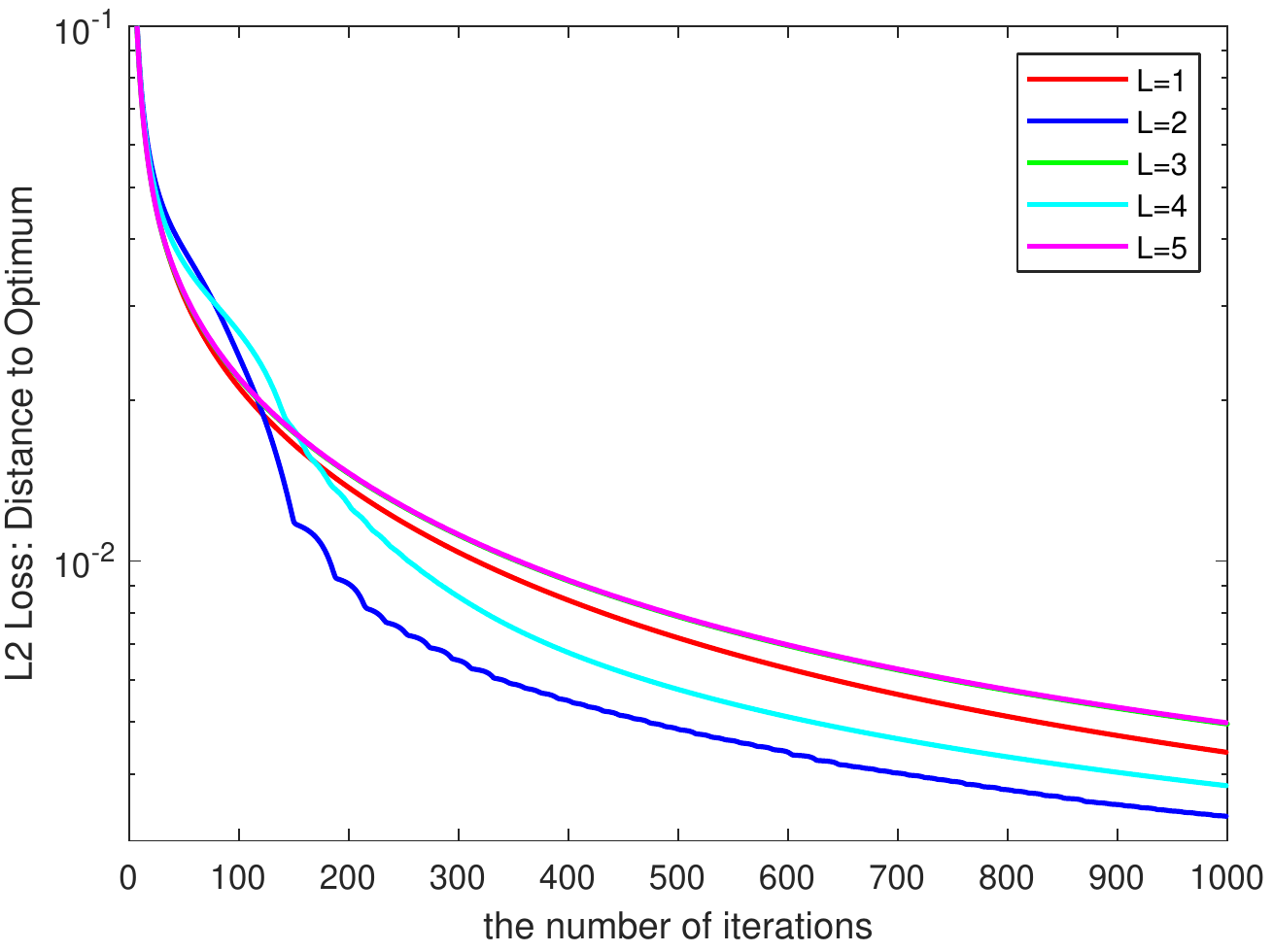}
		\includegraphics[height=4.8cm]{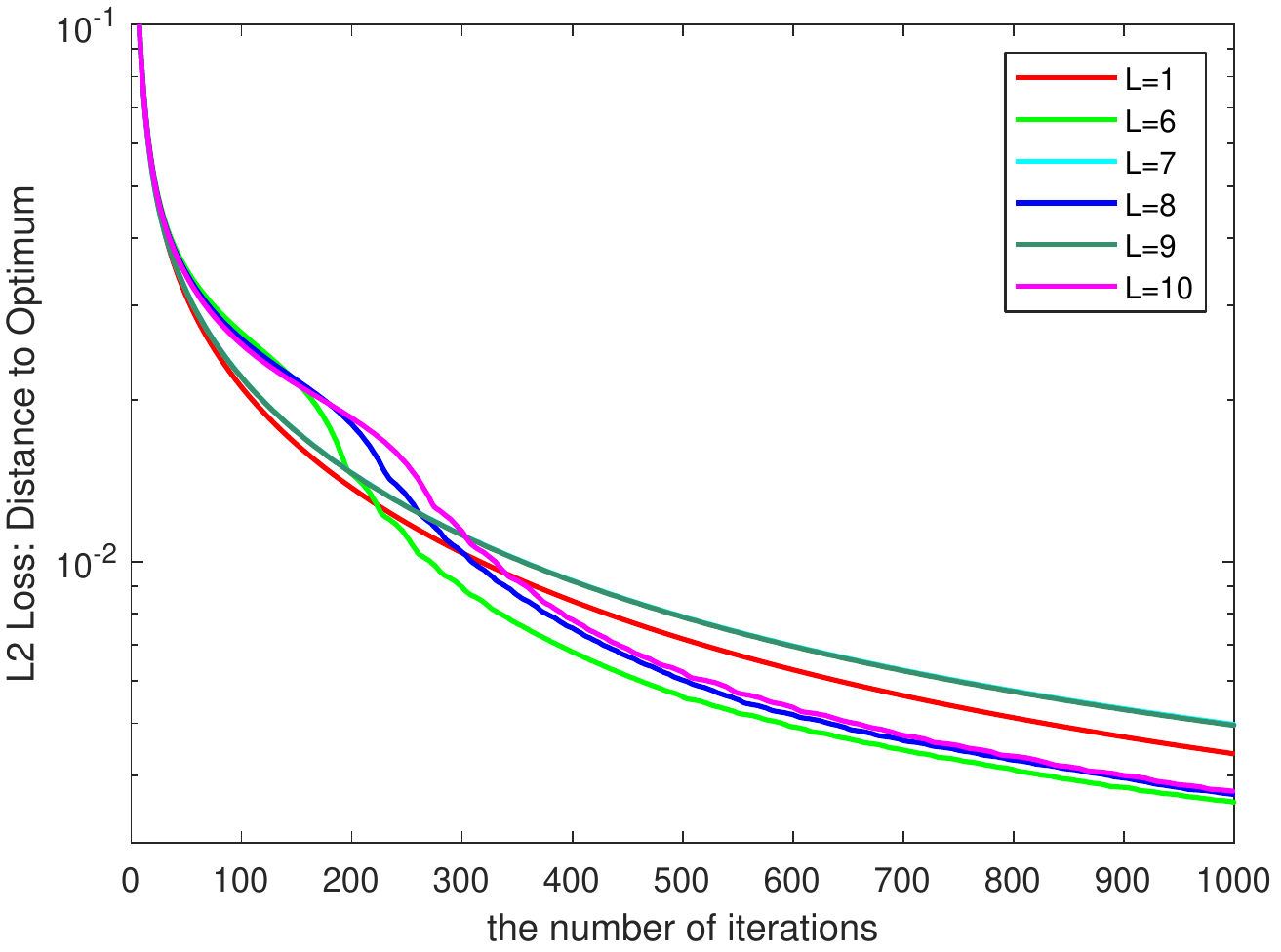}
	}
	\caption{(to be viewed in color) The distances to the global optimum by $L_2$-loss with respect to the number of iterations of the descending BCGD. 
	The identity initialization is employed.
	The network is trained over the MNIST training dataset of 60,000 samples. 
	The width of intermediate layers is $n_\ell = 784$.
	The results by DLNs of odd-depth are very similar so that the lines are overlapping each other.
	The acceleration in convergence is observed by DLNs of even-depth.
	}
	\label{fig:mnist}
\end{figure}

We now train DLNs on the MNIST handwritten digit classification dataset.
For an input image, its corresponding output vector contains a 1 in the index for the correct class and zeros elsewhere.  
The input and output dimensions are $d_\text{in}=784$ and $d_\text{out}=10$, respectively.
In order to strictly compare the effect of depth, we employ the identity initialization to completely remove the randomness from the initialization.
Also, we set the width to $784 = \max\{d_\text{in},d_\text{out}\}$ according to Theorem~\ref{thm:role of width}.
The networks are trained over the entire MNIST training dataset of 60,000 samples.
The input data matrix $\bm{X}$ is not full rank.
Figure~\ref{fig:mnist} shows the distances to the global optimum by $L_2$-loss with respect to the number of iterations of the descending BCGD at ten different depths $L=1,\cdots,10$.
Thus, the speed of convergence is measured against 
the amount of computation. 
We observe the accelerated convergence
by the network whose depth is even but not odd.
We also see that the results by DLNs of odd-depth are very similar so that the lines are overlapping each other.
In this case, the depth 2 network performs the best among others.
We suspect that there is a connection between the parity of depth 
and the acceleration in convergence.
We defer such further investigation to future work.

\begin{figure}[htbp]
	\centerline{
		\includegraphics[height=4.8cm]{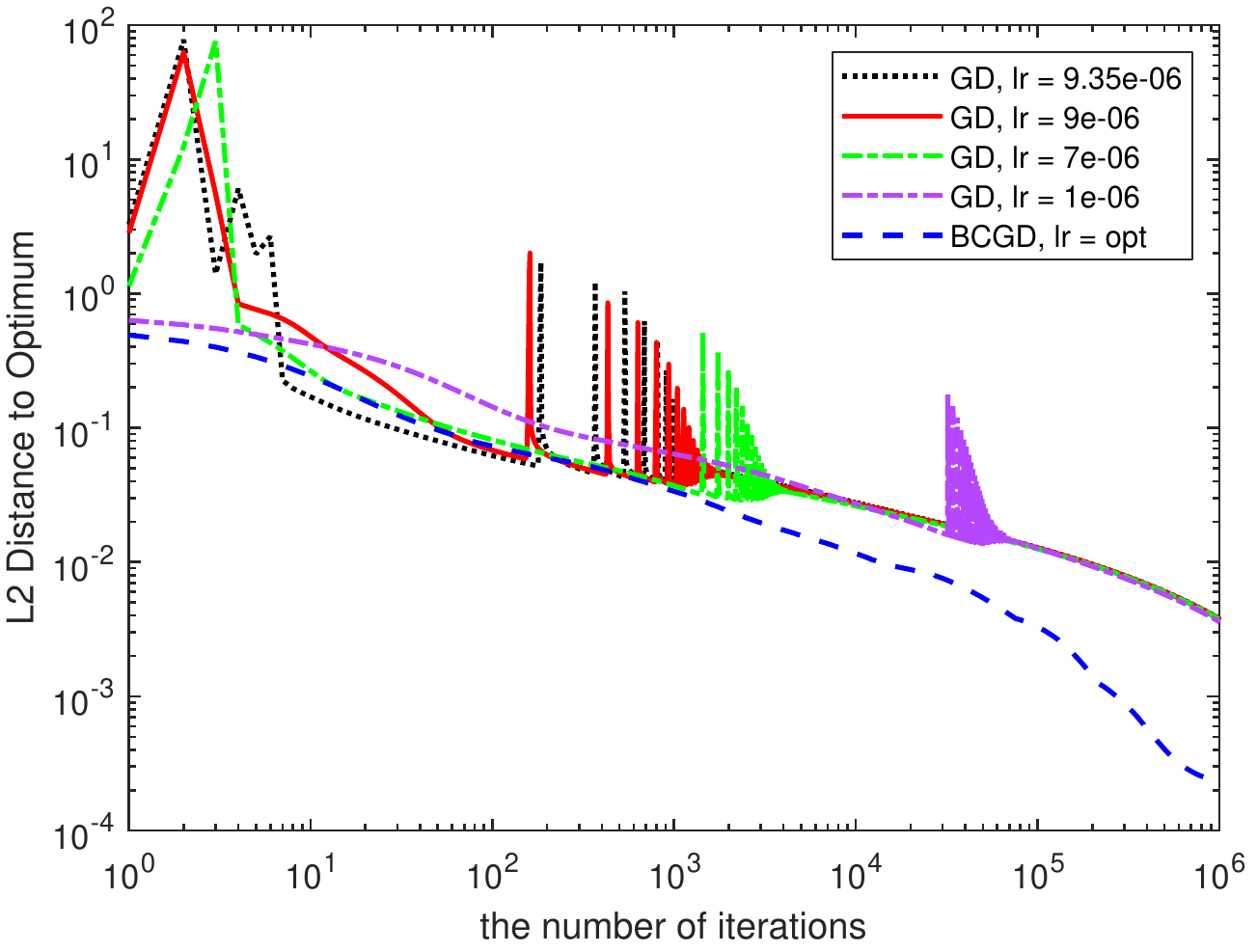}
		\includegraphics[height=4.8cm]{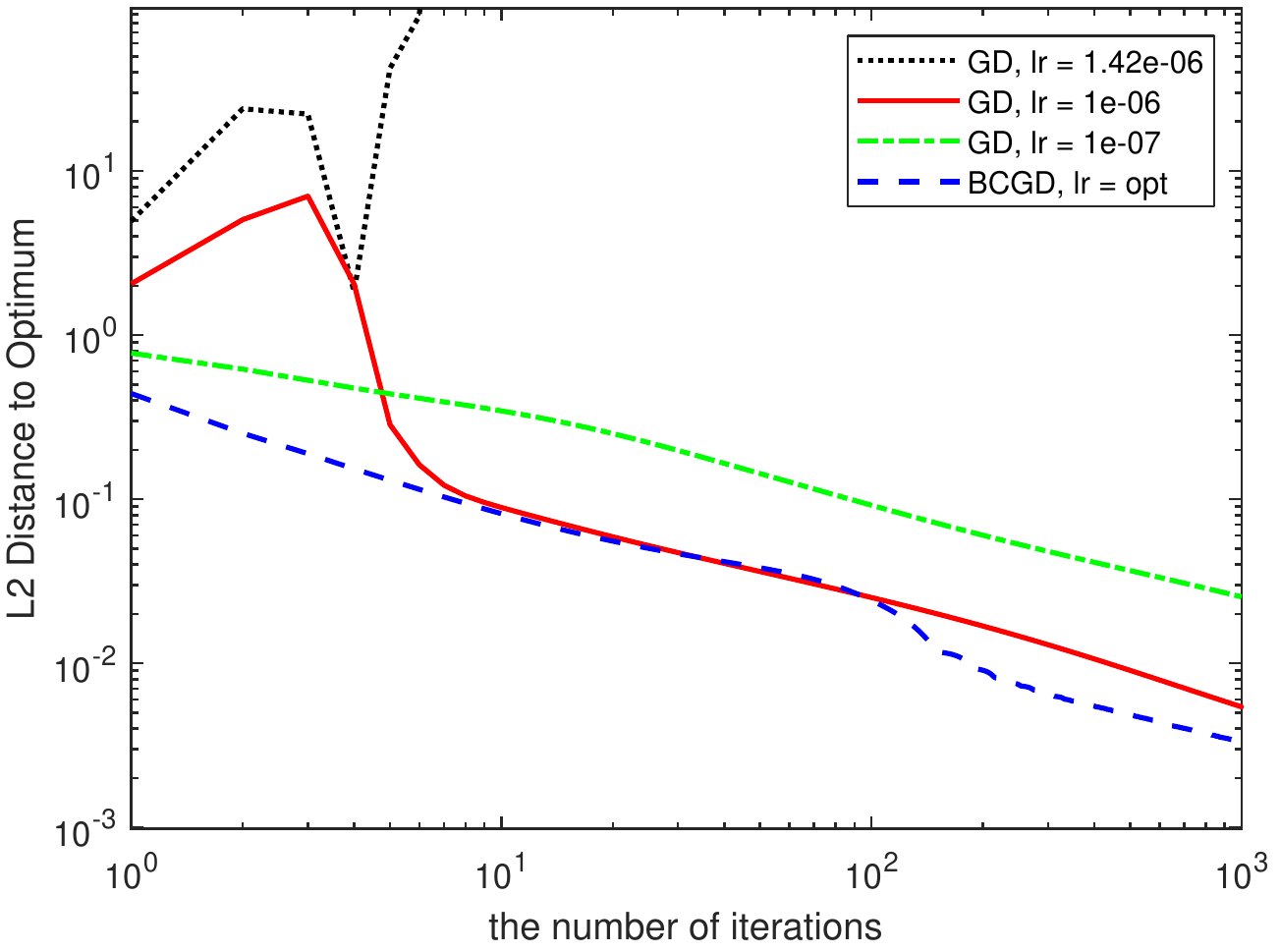}
	}
	\caption{(to be viewed in color) The distances to the global optimum by $L_2$-loss with respect to the number of iterations. 
	The identity initialization is employed.
	(Left) The UCI dataset with $L=4$ and $n_\ell = 128$. 
	(Right) The MNIST dataset with $L=2$ and $n_\ell = 784$.
	}
	\label{fig:realdata-GD}
\end{figure}

Lastly, we compare the performance of BCGD
to GD on the same real data sets.
Again, the learning rates for GD are chosen 
by trial-and-error.
Figure~\ref{fig:realdata-GD}
shows the error trajectories 
for the same learning tasks.
On the left and right, the results for the UCI and the MNIST datasets are presented, respectively.
In all cases, we see that 
BCGD converges faster than GD
while GD with a well-chosen learning rate converges linearly.
We emphasize that GD updates all the weights matrices per iteration,
while BCGD updates only a single matrix per iteration.

\section{Conclusion} \label{sec:conclusion}
In this paper, we studied a layer-wise training for deep linear networks using the block coordinate gradient descent (BCGD).
We established a convergence analysis
and found the optimal learning rate which results in
the fastest decrease in the loss for the next iterate. 
More importantly, the optimal learning rate can directly be applied in practice as no prior knowledge is required.
Also, we identified the effects of depth, width, and initialization
in the training process.
Firstly, we showed that when the orthogonal-like initialization is employed
and the width of the intermediate layers is great than or equal to both the input and output dimensions,
the width plays no roles in gradient-based training. 
Secondly, under some assumptions, 
we proved that the deeper the network is, 
the faster the convergence is guaranteed (when the speed is measured against the number of sweeps).
In an extreme case, the global optimum (within machine accuracy) is achieved after updating each weight matrix only once.
Thirdly, we empirically demonstrated that 
adding more layers
could drastically accelerate convergence,
when it is compared to those of a single layer,
even when the computational cost is considered.
Lastly, we establish a convergence analysis of the block coordinate stochastic gradient descent (BCSGD).
Our analysis indicates that the BCSGD cannot reach the global optimum, however, the converged loss will be staying close to the global optimum. This can be understood as an implicit regularization, which avoids over-fitting.
Numerical examples were provided to justify our theoretical findings and demonstrate the performance of the layer-wise training by BCGD.

\section*{Acknowledgments}
The author would like to thank Dr. Pual Dupuis for his helpful discussion in the early stages of this work,
Dr. Mark Ainsworth for his helpful comments and suggestions on both analysis and examples,
and
Dr. Nadav Cohen for sharing code for numerical experiments.

\appendix
\section{Least square solution} \label{app:lsq-sol}
Without the rank constraint, 
the solution of \eqref{def:depth1-prob} is 
\begin{equation} \label{def:glob-minimizer}
	\bm{W}^*_\text{gen} = \bm{Y}\bm{X}^\dagger + \bm{M}(\bm{X}\bm{X}^\dagger - \bm{I}_{n_0}), \quad \forall \bm{M} \in \mathbb{R}^{n_L \times n_0},
\end{equation}
where $\bm{I}_{n}$ is the identity matrix of size $n\times n$ and $\bm{X}^\dagger$ is the Moore-Pensore pseudo-inverse of $\bm{X}$.
Assuming $\bm{X}$ is a full row rank matrix,
we have $\bm{W}^* = \bm{Y}\bm{X}^\dagger$,
which allows an explicit formula
$\bm{W}^*_{LSQ}= \bm{Y}\bm{X}^T(\bm{X}\bm{X}^T)^{-1}$.
If $\bm{X}$ is not a full row rank matrix,
\eqref{def:depth1-prob} allows infinitely many solutions.
In this case, the least norm solution is often sought and 
it is $\bm{W}^* = \bm{Y}\bm{X}^\dagger$.
Also, for any $\bm{W}$, the following holds:
\begin{align*}
	\mathcal{L}(\bm{W}) = \|\bm{W}\bm{X} - \bm{Y}\|_F^2 
	=  \|\bm{W}\bm{X} - \bm{W}^*\bm{X}\|_F^2 + \mathcal{L}(\bm{W}^*).
\end{align*}
Thus, the minimizing $L_2$-loss is equivalent to minimizing $\|\bm{W}\bm{X} - \bm{W}^*\bm{X}\|_F^2$. 
Furthermore, for whitened data,
the least norm solution is simply 
$\bm{W}^* = \bm{Y}\bm{X}^T$.

With the rank constraint, we consider two cases. 
If $\text{rank}(\bm{Y}\bm{X}^\dagger) \le n^*$,
the rank constrain plays no role in the minimization. Thus, the global minimizer 
is \eqref{def:glob-minimizer}.
Let us consider the case of  $\text{rank}(\bm{Y}\bm{X}^\dagger) > n^*$.
Let $r_x = \text{rank}(\bm{X})$, 
and $\bm{X} = U_x\Sigma_xV_x^T$ be a compact singular value decomposition (SVD) of $\bm{X}$
where only $r_x$ left-singular vectors and 
$r_x$ right-singular vectors corresponding to the non-zero singular values are calculated.
Then, $\bm{X}^\dagger = V_x\Sigma_x^{-1}U_x^T$.
and it can be checked that $\text{rank}(\bm{Y}V_x) = r^*=  \text{rank}(\bm{Y}\bm{X}^\dagger)$.
Let $\bm{Y}V_x = \hat{U}_y\hat{\Sigma}_y\hat{V}_y^T$
be a compact SVD of $\bm{Y}V_x$.
It then can be shown that 
the problem \eqref{def:depth1-prob} is equivalent to 
$$
\min_{\bm{Z}} \|\bm{Z} - \bm{Y}V_x\|_F, \quad \text{subject to} \quad 
\text{rank}(\bm{Z}) \le n^*.
$$ 
To be more precise, if $\bm{Z}^*$ is a solution (the best $n^*$-rank approximation to $\bm{Y}V_x$) to the above,
$\bm{W}^* = \bm{Z}^*\Sigma_x^{-1} U_x^T$ is a solution of \eqref{def:depth1-prob}, which can be explicitly written as 
\begin{equation} \label{lsq-sol-rank-defi}
	\bm{W}^* = \hat{U}_y \Sigma^* \hat{V}_y^T \Sigma_x^{-1} U_x^T,
	\quad \Sigma^*= \begin{bmatrix} \bm{D}_{s} & \bm{0} \\ \bm{0} & \bm{0} \end{bmatrix},
\end{equation}
where $s = \min \{ n^*, r^*\}$ and $\bm{D}_{s}$ is the principal submatrix consisting of the first $s$ rows and columns of $\hat{\Sigma}_y$.
We remark that in general, \eqref{lsq-sol-rank-defi} and the best $n^*$-rank approximation to $\bm{Y}\bm{X}^\dagger$ are not the same.

\section{Gradient of the loss}
For reader's convenience, here we present the calculation of the gradient.
First, let us define the matrix $\bm{J} \in \mathbb{R}^{m \times d_{\text{out}}}$,
\begin{align*}
	\bm{J}^{(k)} = [J_{ij}^{(k)}], \qquad
	J_{ij}^{(k)} = \ell'( \N^L_j(\x^i;\bm{\theta}^{(k)}); \y^i_j), \qquad
	1 \le i \le m, 1 \le j \le d_{\text{out}}.
\end{align*}
Note that if $\ell(a,b) = (a-b)^2/2$, 
$
\bm{J}^{(k)} =(\bm{W}_{L:1}^{(k)}\bm{X} - \bm{Y} )^T.
$
\begin{lemma}
	Let $\bm{\theta} = \{\bm{W}_\ell\}_{\ell=1}^L$
	and $\N^L(\x;\bm{\theta}) = \bm{W}_L\bm{W}_{L-1}\cdots \bm{W}_1\x$,
	where $\bm{W}_\ell \in \mathbb{R}^{n_\ell \times n_{\ell-1}}$ for $1\le \ell \le L$.
	Then
	\begin{align*}
		\frac{\partial \mathcal{L}(\bm{\theta}) }{\partial \bm{W}_\ell}
		= 
		(\bm{W}_{L}\cdots \bm{W}_{\ell+1})^T\bm{J}^T (\bm{W}_{\ell-1}\cdots \bm{W}_1 \bm{X})^T.
	\end{align*}
\end{lemma}
\begin{proof}
	Let us consider the case of $L=2$.
	Let $\bm{\theta} = \{\bm{W}_2, \bm{W}_1\}$, i.e.,
	$\N^2(\x) = \bm{W}_2\bm{W}_1\x$, where
	$\bm{W}_1 \in \mathbb{R}^{n \times d_{\text{in}}}$,
	and $\bm{W}_2 \in \mathbb{R}^{ d_{\text{out}} \times n}$.
	For a matrix $\bm{M}$, let us denote the $j$-th row of $\bm{M}$ by $\bm{M}_{(j,:)}$ 
	and the $i$-th column of $\bm{M}$ by
	$\bm{M}_{(:,i)}$.
	Since $L=2$, the loss function is 
	$\mathcal{L}(\bm{\theta}) = 
	\sum_{j=1}^{d_{\text{out}}}
	\sum_{i=1}^m 
	\ell((\bm{W}_2)_{(j,:)}\bm{W}_1\x^i; \y^i_j)$.
	The direct calculation shows that
	$
	\frac{\partial \mathcal{L}(\bm{\theta}) }{\partial (\bm{W}_1)_{(t,:)}^T}
	= \bm{X} \bm{J}(\bm{W}_2)_{(:,t)}$, 
	$\frac{\partial \mathcal{L}(\bm{\theta}) }{\partial (\bm{W}_2)_{(t,:)}^T} 
	= \bm{W}_1\bm{X}\bm{J}_{(:,t)}$,
	which gives
	$$
	\frac{\partial \mathcal{L}(\bm{\theta}) }{\partial (\bm{W}_1)^T}
	= \bm{X} \bm{J}\bm{W}_2, \qquad
	\frac{\partial \mathcal{L}(\bm{\theta}) }{\partial (\bm{W}_2)^T}
	= \bm{W}_1\bm{X}\bm{J}.
	$$
	For general $L$, it readily follows from the case of $L=2$
	by letting 
	$\bm{X} \to \bm{W}_{\ell-1}\cdots \bm{W}_1 \bm{X}$,
	$\bm{W}_1 \to \bm{W}_\ell$,
	and 
	$\bm{W}_2 \to \bm{W}_{L}\cdots \bm{W}_{\ell+1}$.
\end{proof}

\section{Proof of Theorem \ref{thm:role of width}} 
\label{app:thm:role of width}

For a matrix $\bm{A}$ of size $m\times n$
and a matrix $\bm{B}$ of size $k \times s$
where $m \ge k, n \ge s$, 
we say $\bm{A}$ is equivalent to $\bm{B}$ upto zero-valued padding
if 
$$
\bm{A} = \begin{bmatrix}
\bm{B} & \bm{0} \\
\bm{0} & \bm{0}
\end{bmatrix},
$$
and write $\bm{A} \approxeq \bm{B}$.
\begin{lemma} \label{lemma:width-zero-padding}
	Suppose $\bm{W}_1^{\textbf{k}_{(k,\ell-1)}} \approxeq \tilde{\bm{W}}_1^{\textbf{k}_{(k,\ell-1)}} \in \mathbb{R}^{\max\{n_0, n_L\} \times n_0}$,
	$\bm{W}_L^{\textbf{k}_{(k,\ell-1)}} \approxeq \tilde{\bm{W}}_L^{\textbf{k}_{(k,\ell-1)}} \in \mathbb{R}^{n_L\times \max\{n_0, n_L\}}$
	and 
	$\bm{W}_j^{\textbf{k}_{(k,\ell-1)}} \approxeq \tilde{\bm{W}}_j^{\textbf{k}_{(k,\ell-1)}} \in \mathbb{R}^{\max\{n_0, n_L\} \times \max\{n_0, n_L\}}$ for all $1 < j < L$.
	Then,
	\begin{align*}
	\bm{W}_\ell^{\textbf{k}_{(k,\ell)}} \approxeq
	\tilde{\bm{W}}_\ell^{\textbf{k}_{(k,\ell)}} \in
	\begin{cases}
	\mathbb{R}^{\max\{n_0, n_L\} \times n_0}, & \text{if $\ell = 1$}, \\
	\mathbb{R}^{\max\{n_0, n_L\} \times \max\{n_0, n_L\}}, & \text{if $1 < \ell < L$}, \\
	\mathbb{R}^{n_L \times \max\{n_0, n_L\}}, & \text{if $\ell = L$},
	\end{cases}.
	\end{align*}
\end{lemma}
\begin{proof}
	Let $d_{\max} = \max\{n_0,n_L\}$.
	Note that if 
	$\bm{W}_1 \approxeq \tilde{\bm{W}}_1 \in \mathbb{R}^{d_{\max} \times n_0}$,
	$\bm{W}_L \approxeq \tilde{\bm{W}}_L \in \mathbb{R}^{n_L \times d_{\max}}$,
	and
	$\bm{W}_j \approxeq \tilde{\bm{W}}_j \in \mathbb{R}^{d_{\max} \times d_{\max}}$ for $1 < j < L$,
	since $n_j \ge d_{\max}$ for $1 < j < L$,
	we have 
	$\bm{W}_{L:(j+1)} \approxeq \tilde{\bm{W}}_{L:(j+1)}$ and 
	$\bm{W}_{(j-1):1} \approxeq \tilde{\bm{W}}_{(j-1):1}$ 
	for any $1 < j < L$.
	Specifically,  
	\begin{align*}
	\bm{W}_{L:(j+1)} = 
	\begin{bmatrix}
	\tilde{\bm{W}}_{L:(j+1)} & \bm{0}
	\end{bmatrix}, 
	\qquad
	\bm{W}_{(j-1):1} = 
	\begin{bmatrix}
	\tilde{\bm{W}}_{(j-1):1} \\
	\bm{0}
	\end{bmatrix}.
	\end{align*}
	It then follows from the gradient descent update
	\begin{align*}
	\bm{W}_{\mathfrak{i}(\ell)}^{\textbf{k}_{(s,\ell)}} &= \bm{W}_{\mathfrak{i}(\ell)}^{\textbf{k}_{(s,\ell-1)}} - \eta (\bm{W}_{L:({\mathfrak{i}(\ell)}+1)}^{\textbf{k}_{(s,\ell-1)}})^T\Delta^{\textbf{k}_{(s,\ell-1)}}\bm{XX}^T(\bm{W}_{({\mathfrak{i}(\ell)}-1):1}^{\textbf{k}_{(s,\ell-1)}})^T,
	\end{align*}
	where ${\mathfrak{i}(\ell)} = \ell$ if the ascending BCGD is employed
	and ${\mathfrak{i}(\ell)} = L-\ell+1$ if the descending BCGD is employed,
	that 
	we obtain
	\begin{align*}
	&(\bm{W}_{L:({\mathfrak{i}(\ell)}+1)}^{\textbf{k}_{(s,\ell-1)}})^T\Delta^{\textbf{k}_{(s,\ell-1)}}\bm{XX}^T(\bm{W}_{({\mathfrak{i}(\ell)}-1):1}^{\textbf{k}_{(s,\ell-1)}})^T \\
	&\approxeq 
	(\tilde{\bm{W}}_{L:({\mathfrak{i}(\ell)}+1)}^{\textbf{k}_{(s,\ell-1)}})^T
	\Delta^{\textbf{k}_{(k,\ell-1)}}\bm{XX}^T
	(\tilde{\bm{W}}_{({\mathfrak{i}(\ell)}-1):1}^{\textbf{k}_{(s,\ell-1)}})^T
	\in \mathbb{R}^{d_{\max} \times d_{\max}},
	\end{align*}
	and
	\begin{align*}
	(\bm{W}_{L:{\mathfrak{i}(1)}}^{\textbf{k}_{(s,0)}})^T\Delta^{\textbf{k}_{(s,0)}}\bm{XX}^T
	&\approxeq
	(\tilde{\bm{W}}_{L:{\mathfrak{i}(1)}}^{\textbf{k}_{(s,0)}})^T
	\Delta^{\textbf{k}_{(k,0)}}\bm{XX}^T
	\in \mathbb{R}^{d_{\max} \times n_0},
	\\
	\Delta^{\textbf{k}_{(s,L-1)}}\bm{XX}^T(\bm{W}_{({\mathfrak{i}(L)}-1):1}^{\textbf{k}_{(s,L-1)}})^T
	&\approxeq
	\Delta^{\textbf{k}_{(s,L-1)}}\bm{XX}^T(\tilde{\bm{W}}_{({\mathfrak{i}(L)}-1):1}^{\textbf{k}_{(s,L-1)}})^T
	\in \mathbb{R}^{n_L \times d_{\max}}.
	\end{align*}
	By the assumption on $\bm{W}_j^{\textbf{k}_{(k,\ell-1)}} \approxeq \tilde{\bm{W}}_j^{\textbf{k}_{(k,\ell-1)}}$,
	the proof is completed.
\end{proof}
\begin{proof}[Proof of Theorem \ref{thm:role of width}]
	If the initial weight matrices satisfy 
	$$
	\bm{W}^{(0)}_j \approxeq \tilde{\bm{W}}_j^{(0)} 
	\in
	\begin{cases}
	\mathbb{R}^{\max\{n_0, n_L\} \times n_0}, & \text{if $\ell = 1$}, \\
	\mathbb{R}^{\max\{n_0, n_L\} \times \max\{n_0, n_L\}}, & \text{if $1 < \ell < L$}, \\
	\mathbb{R}^{n_L \times \max\{n_0, n_L\}}, & \text{if $\ell = L$},
	\end{cases},
	$$
	it follows from Lemma~\ref{lemma:width-zero-padding} that 
	for any $s$ and $j$, there exists 
	$\tilde{\bm{W}}_j^{(s)}$ such that 
	$$
	\bm{W}^{(s)}_j \approxeq \tilde{\bm{W}}_j^{(s)} 
	\in
	\begin{cases}
	\mathbb{R}^{\max\{n_0, n_L\} \times n_0}, & \text{if $\ell = 1$}, \\
	\mathbb{R}^{\max\{n_0, n_L\} \times \max\{n_0, n_L\}}, & \text{if $1 < \ell < L$}, \\
	\mathbb{R}^{n_L \times \max\{n_0, n_L\}}, & \text{if $\ell = L$},
	\end{cases},
	$$
	which completes the proof for the balanced initialization.
	
	Suppose $\min\{m,n\} > k=s$. 
	We then write $\bm{A} \approxeq_1 \bm{B}$ if 
	$\bm{A} \approxeq \tilde{\bm{B}}$ where
	$\tilde{\bm{B}}$ is a square matrix of size $\min\{m,n\}$ such that
	$$
	\tilde{\bm{B}} = \begin{bmatrix}
	\bm{B} & \bm{0} \\
	\bm{0} & \bm{I}_{\min\{m,n\} - k}
	\end{bmatrix}.
	$$
	Let $\bm{W}_j$ be a matrix of size $n_j \times n_{j-1}$
	and $n_j \ge \max\{n_0, n_L\}$ for all $1\le j \le L$.
	Suppose 
	\begin{equation} \label{cond-identity-init}
	\begin{split}
	\bm{W}_j &\approxeq \tilde{\bm{W}}_j
	\in
	\begin{cases}
	\mathbb{R}^{\max\{n_0, n_L\} \times n_0}, & \text{if $j = 1$}, \\
	\mathbb{R}^{n_L \times \max\{n_0, n_L\}}, & \text{if $j = L$},
	\end{cases}, \\
	\bm{W}_j &\approxeq_1 \tilde{\bm{W}}_j
	\in
	\mathbb{R}^{\max\{n_0, n_L\} \times \max\{n_0, n_L\}},
	\text{if $1 < j < L$}.
	\end{split}
	\end{equation}
	Let $
	\bm{W}_L = \begin{bmatrix}
	\tilde{\bm{W}_L} & \bm{0}
	\end{bmatrix},
	$
	where 
	$\tilde{\bm{W}_L} \in \mathbb{R}^{n_L \times \max\{n_0, n_L\}}$.
	Then, 
	\begin{align*}
	\bm{W}_{L:(j+1)} &= 
	\begin{bmatrix}
	\tilde{\bm{W}_L} & \bm{0}
	\end{bmatrix}
	\begin{bmatrix}
	\hat{\bm{B}}_{(L-1):(j+1)} & \bm{0} \\
	\bm{0} & \bm{0}
	\end{bmatrix},
	\\
	\hat{\bm{B}}_{(L-1):(j+1)} &= \begin{bmatrix}
	\tilde{\bm{W}}_{(L-1):(j+1)} & \bm{0} \\
	\bm{0} & \bm{I}_{n_{\min}^{L-1}(j+1) - r}
	\end{bmatrix},
	\end{align*}
	where $n_{\min}^{i}(j) = \min_{j-1 \le \ell \le i} n_\ell$
	for $1\le j \le i+1$.
	Thus, $\bm{W}_{L:(j+1)} \approxeq \tilde{\bm{W}}_{L:(j+1)}$.
	Similarly, $\bm{W}_{(j-1):1} \approxeq \tilde{\bm{W}}_{(j-1):1}$.
	It then follows from a similar argument used in Lemma~\ref{lemma:width-zero-padding}
	that 
	if the initial weight matrices satisfy \eqref{cond-identity-init},
	then the weight matrices updated by any gradient based optimization
	also 
	satisfy \eqref{cond-identity-init}.
	This completes the proof 
	for the identity initialization.
\end{proof}

\section{Proof of Theorem~\ref{thm:convg-l2}}
\label{app:thm:convg-l2}
\begin{proof}
For notational convenience, for $j > i$, let
$$
\bm{W}_j \bm{W}_{j-1} \cdots \bm{W}_{i} = \bm{W}_{j:i}.
$$
By definition, 
it follows from the update rule that
\begin{align*}
    \bm{W}_\ell^{(k+1)}
    = 
    \bm{W}_\ell^{(k)}
    - \eta_\ell^{\textbf{k}_{(k,\ell-1)}}
    (\bm{W}_{(\ell-1):1}^{(k+1)}
    \bm{X} \bm{J}^{\textbf{k}_{(k,\ell-1)}}\bm{W}_{L:(\ell+1)}^{(k)})^T.
\end{align*}
By multiplying $\bm{W}_{(\ell-1):1}^{(k+1)}\bm{X}$ from right, 
and $\bm{W}_{L:(\ell+1)}^{(k)}$ from left
and subtracting $\bm{W}^*\bm{X}$
in the both sides, 
we obtain
\begin{multline*}
    (\bm{W}_{L:(\ell+1)}^{(k)}\bm{W}_{\ell:1}^{(k+1)} - \bm{W}^*)\bm{X} \\
    = 
    (\bm{W}_{L:\ell}^{(k)}\bm{W}_{(\ell-1):1}^{(k+1)} - \bm{W}^*)\bm{X}
    - \eta_\ell^{\textbf{k}_{(k,\ell-1)}}
    \bm{A}_\ell^{(k)}
    (\bm{J}^{\textbf{k}_{(k,\ell-1)}})^T
    \bm{B}_\ell^{(k+1)},
\end{multline*}
where
\begin{align*}
    \bm{A}_\ell^{(k)} &= \bm{W}_{L:(\ell+1)}^{(k)}
    (\bm{W}_{L:(\ell+1)}^{(k)})^T \in \mathbb{R}^{d_{\text{out}} \times d_{\text{out}}},
    \\
    \bm{B}_\ell^{(k)} &= \bm{X}^T(\bm{W}_{(\ell-1):1}^{(k)})^T
    \bm{W}_{(\ell-1):1}^{(k)}\bm{X}
    \in \mathbb{R}^{m \times m}.
\end{align*}
Since $\ell(a;b) = (a-b)^2/2$, 
we have
\begin{align*}
\bm{X}\bm{J}^{\textbf{k}_{(k,\ell-1)}} &=
\bm{X}(\bm{W}^{(k)}_{L:\ell}\bm{W}^{(k+1)}_{(\ell-1):1}\bm{X} - \bm{Y})^T 
\\
&=\bm{X}(\bm{W}^{(k)}_{L:\ell}\bm{W}^{(k+1)}_{(\ell-1):1}\bm{X} - \bm{Y}\bm{X}^\dagger\bm{X} + \bm{Y}\bm{X}^\dagger\bm{X} - \bm{Y})^T 
\\
&=(\bm{W}^{(k)}_{L:\ell}\bm{W}^{(k+1)}_{(\ell-1):1}\bm{X}\bm{X}^T - \bm{Y}\bm{X}^\dagger\bm{X}\bm{X}^T + \bm{Y}(\bm{X}^\dagger\bm{X}\bm{X}^T - \bm{X}^T))^T 
\\
&=
((\bm{W}^{(k)}_{L:\ell}\bm{W}^{(k+1)}_{(\ell-1):1} - \bm{W}^*)\bm{X}\bm{X}^T)^T
=
\bm{X}(\Delta^{\textbf{k}_{k,\ell-1}})^T,
\end{align*}
where $\bm{X}^\dagger\bm{X}\bm{X}^T = (\bm{X}^\dagger \bm{X})^T\bm{X}^T = (\bm{X}\bm{X}^\dagger \bm{X})^T = \bm{X}^T$
is used in the 4th equality.

Let
$$
\Delta^{\textbf{k}_{(k,\ell)}} := \bm{W}_{L:(\ell+1)}^{(k)}\bm{W}_{\ell:1}^{(k+1)}\bm{X}-\bm{W}^*\bm{X} \in \mathbb{R}^{d_{\text{out}} \times  m}.
$$
Then we have
\begin{align*}
    \Delta^{\textbf{k}_{(k,\ell)}} = 
    \Delta^{\textbf{k}_{(k,\ell-1)}}
    - \eta^{\textbf{k}_{(k,\ell-1)}}_\ell 
    \bm{A}_\ell^{(k)}
    \Delta^{\textbf{k}_{(k,\ell-1)}}
    \bm{B}_\ell^{(k+1)}.
\end{align*}
Since $\bm{A}_\ell^{(k)}$ and $\bm{B}_\ell^{(k)}$ are symmetric, 
we have diagonal transformations,
\begin{align*}
    (\bm{U}_{\ell}^{(k)})^{T}\bm{A}_\ell^{(k)}\bm{U}_{\ell}^{(k)} &=
    \bm{D}^{(k)}_{A,\ell} = \text{diag}(\lambda_{\ell, i}^{(k)}), 
    \quad 1 \le i \le  d_{\text{out}}, \\
    (\bm{V}_{\ell}^{(k)})^{T}\bm{B}_\ell^{(k)}\bm{V}_{\ell}^{(k)} &= \bm{D}^{(k)}_{B,\ell}=\text{diag}(\mu^{(k)}_{\ell, j}), \quad
    1 \le j \le m,
\end{align*}
where $\bm{V}_{\ell}^{(k)}$ and $\bm{U}_{\ell}^{(k)}$ are orthogonal matrices, $\lambda_{\ell, 1}^{(k)} \ge \cdots \ge \lambda_{\ell, d_\text{out}}^{(k)}$,
and $\mu^{(k)}_{\ell, 1} \ge \cdots \ge \mu^{(k)}_{\ell, m}$.
We remark that 
$\mu^{(k)}_{\ell, d_\text{in}+1} = \cdots = \mu^{(k)}_{\ell, m} = 0$
if $d_\text{in} = n_0 < m$.
%
Thus, we have
\begin{equation} \label{eqn-thm:convg-l2}
    \Delta^{\textbf{k}_{(k,\ell)}} = 
    \Delta^{\textbf{k}_{(k,\ell-1)}}
    - \eta^{\textbf{k}_{(k,\ell-1)}}_\ell 
    \bm{U}_{\ell}^{(k)} \bm{D}^{(k)}_{A,\ell} (\bm{U}_{\ell}^{(k)})^{T}
    \Delta^{\textbf{k}_{(k,\ell-1)}}
    {\bm{V}}_{\ell}^{(k+1)} \bm{D}^{(k+1)}_{B,\ell} ({\bm{V}}_{\ell}^{(k+1)})^{T}.
\end{equation}
Let 
$
\tilde{\Delta}^{\textbf{k}_{(k,t,\ell)}} =
(\bm{U}_{\ell}^{(k)})^{T}\Delta^{\textbf{k}_{(k,t)}}{\bm{V}}_{\ell}^{(k+1)}.
$
Then, \eqref{eqn-thm:convg-l2} becomes
\begin{align*}
    \tilde{\Delta}^{\textbf{k}_{(k,\ell,\ell)}} = 
    \tilde{\Delta}^{\textbf{k}_{(k,\ell-1,\ell)}}
    - \eta^{\textbf{k}_{(k,\ell-1)}}_\ell 
    \bm{D}^{(k)}_{A,\ell}
    \tilde{\Delta}^{\textbf{k}_{(k,\ell-1,\ell)}}
    \bm{D}^{(k+1)}_{B,\ell}.
\end{align*}
Then, the $(i,j)$-entry of $\tilde{\Delta}_{\textbf{k}_{k,\ell}}$
is 
\begin{align*}
(\tilde{\Delta}^{\textbf{k}_{(k,\ell)}})_{ij}
= 
\left(1
-\eta^{\textbf{k}_{(k,\ell-1)}}_\ell   \lambda_{\ell, i}^{(k)}\mu_{\ell, j}^{(k+1)}\right)(\tilde{\Delta}^{\textbf{k}_{(k,\ell-1)}})_{ij}, 
\quad 1 \le i \le d_\text{out}, 1 \le j \le m,
\end{align*}
and we have
\begin{align*}
\|\tilde{\Delta}^{\textbf{k}_{(k,\ell,\ell)}}\|_F^2
&= \sum_{i,j} \left(1
-\eta^{\textbf{k}_{(k,\ell-1)}}_\ell   \lambda_{\ell, i}^{(k)}\mu_{\ell, j}^{(k+1)}\right)^2
(\tilde{\Delta}^{\textbf{k}_{(k,\ell-1)}})_{ij}^2 = \mathcal{F}(\eta^{\textbf{k}_{(k,\ell-1)}}_\ell).
\end{align*}
We then choose the learning rate which minimizes $\mathcal{F}(\eta^{\textbf{k}_{(k,\ell-1)}}_\ell)$
and it is 
\begin{equation} \label{app:L2-Optimal-LR}
\eta^{\textbf{k}_{(k,\ell-1)}}_\text{opt}
= \frac{
	\left\|\frac{\partial \mathcal{L}}{\partial \bm{W}_\ell}\big|_{\bm{\theta}=\bm{\theta}^{\textbf{k}_{(k,\ell-1)}}}\right\|_F^2}{\left\|\bm{W}_{L:(\mathfrak{i}(\ell)+1)}^{\textbf{k}_{(k,\ell-1)}}\frac{\partial \mathcal{L}}{\partial \bm{W}_\ell}\big|_{\bm{\theta}=\bm{\theta}^{\textbf{k}_{(k,\ell-1)}}}\bm{W}_{(\mathfrak{i}(\ell)-1):1}^{\textbf{k}_{(k,\ell-1)}}\bm{X}\right\|_F^2}.
\end{equation}
Thus, with the optimal learning rate of \eqref{app:L2-Optimal-LR}, we obtain
\begin{align*}
\|{\Delta}^{\textbf{k}_{(k,\ell)}}\|_F^2
&=\|{\Delta}^{\textbf{k}_{(k,\ell-1)}}\|_F^2
-\eta^{\textbf{k}_{(k,\ell-1)}}_\text{opt}
\left\|\frac{\partial \mathcal{L}}{\partial \bm{W}_\ell}\big|_{\bm{\theta}=\bm{\theta}^{\textbf{k}_{(k,\ell-1)}}}\right\|_F^2
\\
&=\|{\Delta}^{\textbf{k}_{(k,\ell-1)}}\|_F^2
-\frac{
	\left\|\frac{\partial \mathcal{L}}{\partial \bm{W}_\ell}\big|_{\bm{\theta}=\bm{\theta}^{\textbf{k}_{(k,\ell-1)}}}\right\|_F^4}{\left\|\bm{W}_{L:(\mathfrak{i}(\ell)+1)}^{\textbf{k}_{(k,\ell-1)}}\frac{\partial \mathcal{L}}{\partial \bm{W}_\ell}\big|_{\bm{\theta}=\bm{\theta}^{\textbf{k}_{(k,\ell-1)}}}\bm{W}_{(\mathfrak{i}(\ell)-1):1}^{\textbf{k}_{(k,\ell-1)}}\bm{X}\right\|_F^2}.
\end{align*}
For a matrix $\bm{M}$, the $j$-th column 
and the $i$-th row of $\bm{M}$ are denoted by $(\bm{M})^j$ and $(\bm{M})_i$, respectively.
We note that all rows of $\Delta^{\textbf{k}_{(k,\ell-1,\ell)}}$ 
are in $\text{range}(\bm{X}^T)$
and $\text{span}\{ (\bm{V}_{\ell}^{(k+1)})^j; 1\le j \le r_x \} = \text{range}(\bm{X}^T)$, where $r_x = \text{rank}(\bm{X})$.
We remark that if $\mu_{\ell,k}^{(k+1)} = 0$  for some $k \le r_x$,
we choose the corresponding $(\bm{V}_{\ell}^{(k+1)})^k$
so that $\text{range}(\bm{X}) = \text{span}\{ (\bm{V}_{\ell}^{(k+1)})^j; 1\le j \le r_x \}$ holds.
Thus,
$(\Delta^{\textbf{k}_{(k,\ell)}}{\bm{V}}_{\ell}^{(k+1)})^j = 0$ for $j > r_x$.
This gives that the $(i,j)$-entry of $\tilde{\Delta}_{\textbf{k}_{k,\ell}}$
is equal to
\begin{align*}
    (\tilde{\Delta}^{\textbf{k}_{(k,\ell)}})_{ij}
    = 
    \left(1
    -\eta^{\textbf{k}_{(k,\ell-1)}}_\ell   \lambda_{\ell, i}^{(k)}\mu_{\ell, j}^{(k+1)}\right)(\tilde{\Delta}^{\textbf{k}_{(k,\ell-1)}})_{ij}, 
    \quad 1 \le i \le d_\text{out}, 1 \le j \le r_x, 
\end{align*}
and zero otherwise.

Suppose that $(\bm{W}_L^{(0)})^j \in K $ for all $1\le j \le n_{L-1}$
where 
$\text{range}(\bm{Y}\bm{X}^\dagger) \subset K \subset \mathbb{R}^{n_L}$.
It then can be checked that 
$(\bm{W}_L^{(k)})^j \in K $ for all $k$ and $j$
and thus 
$(\Delta^{\textbf{k}_{(k,\ell-1)}})^j \in K$.
Also, from the similar argument used in the above, we have
$$
\text{span}\{(\bm{U}^{(k)}_\ell)^j | j=1,\cdots, r\} = K,\qquad r = \dim K.
$$
Thus,
$((\bm{U}^{(k)}_\ell)^T\Delta^{\textbf{k}_{(k,\ell)}}))_i = 0$ for $i > r$
and we have
\begin{equation} \label{eqn2-thm:convg-l2}
(\tilde{\Delta}^{\textbf{k}_{(k,\ell)}})_{ij}
= 
\left(1
-\eta^{\textbf{k}_{(k,\ell-1)}}_\ell   \lambda_{\ell, i}^{(k)}\mu_{\ell, j}^{(k+1)}\right)(\tilde{\Delta}^{\textbf{k}_{(k,\ell-1)}})_{ij}, 
\quad 1 \le i \le r, 1 \le j \le r_x, 
\end{equation}
and zero otherwise.

If the learning rate $\eta^{\textbf{k}_{(k,\ell-1)}}_\ell$
is chosen to satisfy
\begin{equation} \label{eqn3-thm:convg-l2}
    0 < \eta^{\textbf{k}_{(k,\ell-1)}}_\ell <
    \frac{2}{\max_{i,j} \left(\lambda_{\ell, i}^{(k)}\mu_{\ell, j}^{(k+1)}\right)}
    =
    \frac{2}{
    \sigma_{\max}^2(\bm{W}_{L:(\mathfrak{i}(\ell)+1)}^{\textbf{k}_{(k,\ell-1)}})
    \sigma_{\max}^2(\bm{W}_{(\mathfrak{i}(\ell)-1):1}^{\textbf{k}_{(k,\ell-1)}}\bm{X})},
\end{equation}
we have
\begin{align*}
    ((\tilde{\Delta}^{\textbf{k}_{(k,\ell)}})_{ij})^2
    \le 
    ((\tilde{\Delta}^{\textbf{k}_{(k,\ell-1)}})_{ij})^2
    (\gamma^{\textbf{k}_{(k,\ell-1)}})^2,
\end{align*}
where $\gamma^{\textbf{k}_{(k,\ell-1)}} =\max\{\gamma^{\textbf{k}_{(k,\ell-1)}}_1, \gamma^{\textbf{k}_{(k,\ell-1)}}_2 \}$,
\begin{align*}
    \gamma^{\textbf{k}_{(k,\ell-1)}}_1
     &=
     1-\eta^{\textbf{k}_{(s,\ell-1)}}_\ell
     \sigma_{r}^2(\bm{W}_{L:(\mathfrak{i}(\ell)+1)}^{\textbf{k}_{(s,\ell-1)}})
     \sigma_{r}^2(\bm{W}_{(\mathfrak{i}(\ell)-1):1}^{\textbf{k}_{(k,\ell-1)}}\bm{X})
     , \\
     \gamma^{\textbf{k}_{(k,\ell-1)}}_2 &=
     \eta^{\textbf{k}_{(s,\ell-1)}}_\ell
     \|\bm{W}_{L:(\mathfrak{i}(\ell)+1)}^{\textbf{k}_{(s,\ell-1)}})\|^2
     \|\bm{W}_{(\mathfrak{i}(\ell)-1):1}^{\textbf{k}_{(k,\ell-1)}}\bm{X}\|^2-1
     .
\end{align*}
Note that from the relation of $\|M\|_F^2 = \text{Tr}(MM^T)$, we have
\begin{align*}
    \|\tilde{\Delta}^{\textbf{k}_{(k,\ell)}}\|_F^2
&= \text{Tr}((\bm{U}_{\ell}^{(k)})^{T}\Delta^{\textbf{k}_{(k,\ell)}}\tilde{\bm{V}}_{k,\ell}
(\tilde{\bm{V}}_{k,\ell})^T
(\Delta^{\textbf{k}_{(k,\ell)}})^T\bm{U}_{\ell}^{(k)})
\\
&= \text{Tr}((\bm{U}_{\ell}^{(k)})^{T}\Delta^{\textbf{k}_{(k,\ell)}}
(\Delta^{\textbf{k}_{(k,\ell)}})^T\bm{U}_{\ell}^{(k)}) \\
&=\text{Tr}(\Delta^{\textbf{k}_{(k,\ell)}}
(\Delta^{\textbf{k}_{(k,\ell)}})^T\bm{U}_{\ell}^{(k)}(\bm{U}_{\ell}^{(k)})^{T})
\\
&= \text{Tr}(\Delta^{\textbf{k}_{(k,\ell)}}
(\Delta^{\textbf{k}_{(k,\ell)}})^T)
= \|\Delta^{\textbf{k}_{(k,\ell)}}\|_F^2.
\end{align*}
Therefore,
\begin{align*}
    \|\tilde{\Delta}^{\textbf{k}_{(k,\ell)}}\|_F^2
    \le \|\tilde{\Delta}^{\textbf{k}_{(k,\ell-1)}}\|_F^2
    (\gamma^{\textbf{k}_{(k,\ell-1)}})^2 \iff
    \|\Delta^{\textbf{k}_{(k,\ell)}}\|_F^2
    \le \|\Delta^{\textbf{k}_{(k,\ell-1)}}\|_F^2
    (\gamma^{\textbf{k}_{(k,\ell-1)}})^2.
\end{align*}
By recursively applying the above, we obtain
\begin{align*}
    \|\Delta^{\textbf{k}_{k}}\|_F^2
    \le
    \|\Delta^{\textbf{k}_{0}}\|_F^2
    \prod_{s=0}^{k-1} \left(\prod_{\ell=1}^L
    (\gamma^{\textbf{k}_{(s,\ell-1)}})^2\right),
\end{align*}
which completes the proof.
\end{proof}

\section{Proof of Lemma~\ref{lemma-min-sing-value}}
\label{app:lemma-min-sing-value}
\begin{proof}
	Suppose $\|\bm{W}^{\textbf{k}_0} - \bm{W}^*\|_F \le \tilde{\sigma}_{\min} - c/\|\bm{X}\|$
	where $\tilde{\sigma}_{\min} = \sigma_{\min}(\bm{W}^*\bm{X})/\|\bm{X}\|$,
	where $c$ will be chosen later.
	It then follows from the assumption that
	$$
	\|\bm{W}^{\textbf{k}_0}\bm{X} - \bm{W}^*\bm{X}\|_F \le
	\|\bm{W}^{\textbf{k}_0} - \bm{W}^*\|_F \|\bm{X}\|
	\le \sigma_{\min}(\bm{W}^*\bm{X}) - c.
	$$
	Then for any $\bm{W}$ satisfying $\|\bm{W}\bm{X} - \bm{W}^*\bm{X}\|_F \le \|\bm{W}^{\textbf{k}_0}\bm{X} - \bm{W}^*\bm{X}\|_F$,
	we have 
	\begin{align*}
	\sigma_{\min}(\bm{W}\bm{X}) &\ge \sigma_{\min}(\bm{W}^*\bm{X}) - \sigma_{\max}(\bm{W}\bm{X} - \bm{W}^*\bm{X}) \\
	&\ge 
	\sigma_{\min}(\bm{W}^*\bm{X}) - \|\bm{W}\bm{X} - \bm{W}^*\bm{X}\|_F 
	\ge c > 0.
	\end{align*}
	From Theorem~\ref{thm:convg-l2}, since $\|\bm{W}^{\textbf{k}_j}\bm{X} - \bm{W}^*\bm{X}\|_F \le\|\bm{W}^{\textbf{k}_0}\bm{X} - \bm{W}^*\bm{X}\|_F$ for any $j$, 
	we obtain $\sigma_{\min}(\bm{W}^{\textbf{k}_j}\bm{X}) \ge c > 0$.
	
	For notational convenience, let 
	$A = \bm{W}^{\textbf{k}_{(k,\ell-1)}}_{L:(\mathfrak{i}(\ell)+1)}$,
	$B = \bm{W}^{\textbf{k}_{(k,\ell-1)}}_{\mathfrak{i}(\ell)}$,
	and
	$C = \bm{W}^{\textbf{k}_{(k,\ell-1)}}_{(\mathfrak{i}(\ell)-1):1}\bm{X}$.
	Then, $\bm{W}^{\textbf{k}_{(k,\ell-1)}}\bm{X} = ABC$.
	Note that $\sigma_{s}(ABC) = \sigma_{s}(C^TB^TA^T)$.
	It then follows from
	\begin{equation} \label{app:sv-lowbd}
	\begin{split}
	0 < c \le  \sigma_{\min}(ABC) \le \sigma_{s}(ABC) &= \max_{S:\dim (S) = s} \min_{x \in S, \|x\|=1} \|ABCx\|
	\\
	&\le \|AB\|\max_{S:\dim (S) = s} \min_{x \in S, \|x\|=1} \|Cx\| \\
	&= \|AB\|\sigma_{s}(C), \quad \min\{n_0, n_L\} \le s \le 1,
	\end{split}
	\end{equation}
	that $\sigma_{s}(C) > \frac{c}{\|AB\|}$.
	Similarly, $\sigma_{s}(A) > \frac{c}{\|BC\|}$.
	
	Note that it follow from Theorem~\ref{thm:role of width} that 
	for any $s$ and $\ell$, 
	\begin{equation} \label{matrix-update-balanced}
	\begin{split}
	\bm{W}_j^{(s)} &\approxeq \tilde{\bm{W}}_j^{(s)}
	\in
	\begin{cases}
	\mathbb{R}^{\max\{n_0, n_L\} \times n_0}, & \text{if $j = 1$}, \\
	\mathbb{R}^{n_L \times \max\{n_0, n_L\}}, & \text{if $j = L$},
	\end{cases}, \\
	\bm{W}_j^{(s)} &\approxeq_1 \tilde{\bm{W}}_j^{(s)}
	\in
	\mathbb{R}^{\max\{n_0, n_L\} \times \max\{n_0, n_L\}},
	\text{if $1 < j < L$}.
	\end{split}
	\end{equation}
	Then, for any $\textbf{k}=(k_1,\cdots,k_L)$ and $\ell \in \{1,\cdots,L-1\}$, we have
	\begin{align*}
	\bm{W}_{L:\ell+1}^{\textbf{k}} 
	= \tilde{\bm{W}}_L^{(k_L)}\cdots
	\tilde{\bm{W}}_{\ell+1}^{(k_{\ell+1})},
	\quad 
	\bm{W}_{(\ell-1):1}^{\textbf{k}}\bm{X}
	= \tilde{\bm{W}}_{\ell-1}^{(k_{\ell-1})}\cdots
	\tilde{\bm{W}}_{1}^{(k_{1})}\bm{X}.
	\end{align*}
	Since $n_\ell \ge \max\{n_0, n_L\}$, we have
	$$
	\sigma_{\min}(\bm{W}_{L:(\mathfrak{i}(\ell)+1)}^{\textbf{k}_{(s,\ell-1)}})
	\ge \prod_{j = \mathfrak{i}(\ell)+1}^L \sigma_{\min}(\tilde{\bm{W}}_{j}^{\textbf{k}_{(s,\ell-1)}}).
	$$
	Similarly, 
	$\sigma_{\min}(\bm{W}_{(\mathfrak{i}(\ell)-1):1}^{\textbf{k}_{(s,\ell-1)}}\bm{X})
	\ge 
	\sigma_{\min}(\bm{X})
	\prod_{j=1}^{\mathfrak{i}(\ell)-1} \sigma_{\min}({\bm{W}}_{j}^{\textbf{k}_{(s,\ell-1)}}).
	$
	From \eqref{app:sv-lowbd}, we have
	\begin{equation} \label{norm-lower-bound}
	\begin{split}
	\|\bm{W}_{L:(\mathfrak{i}(\ell)+1)}^{\textbf{k}_{(s,\ell-1)}}\|
	&\ge \sigma_{s}(\bm{W}_{L:(\mathfrak{i}(\ell)+1)}^{\textbf{k}_{(s,\ell-1)}})
	\ge \frac{c}{\|\bm{W}_{\mathfrak{i}(\ell):1}^{\textbf{k}_{(s,\ell-1)}}\bm{X}\|}, \\ 
	\|\bm{W}_{(\mathfrak{i}(\ell)-1):1}^{\textbf{k}_{(s,\ell-1)}}\bm{X}\|
	&\ge \sigma_{s}(\bm{W}_{(\mathfrak{i}(\ell)-1):1}^{\textbf{k}_{(s,\ell-1)}}\bm{X})
	\ge
	\frac{c}{\|\bm{W}_{L:(\mathfrak{i}(\ell)+1)}^{\textbf{k}_{(s,\ell-1)}}\|}.
	\end{split}
	\end{equation}
	
	Let 
	\begin{equation}
	\mathcal{R}(\bm{\theta}^{\textbf{k}_{(s,\ell-1)}})
	= \max_{1 \le j \le \ell}
	\|\bm{W}_{\mathfrak{i}(j)}^{\textbf{k}_{(s,\ell-1)}} - \bm{W}_{\mathfrak{i}(j)}^{(0)}\|,
	\end{equation}
	and 
	\begin{equation}
	\mathcal{R}(sL + \ell)
	= \max\left\{ \max_{0 \le i < s} \mathcal{R}(\bm{\theta}^{\textbf{k}_{(i,L-1)}}), \mathcal{R}(\bm{\theta}^{\textbf{k}_{(s,\ell-1)}}) \right\}.
	\end{equation}
	By applying the induction on the number of iterations of the BCGD,
	we claim that 
	there exists $0 < R < 1$ such that 
	\begin{align*}
	\mathcal{R}(k) \le R, \forall k.
	\end{align*}
	Since $\mathcal{R}(0) = 0$, the base case holds trivially.
	Suppose $\mathcal{R}(sL+\ell-1) \le R$.
	We want to show that $\mathcal{R}(sL+\ell) \le R$.
	Note that since 
	$\bm{W}_{\mathfrak{i}(j)}^{\textbf{k}_{(s,\ell)}} = \bm{W}_{\mathfrak{i}(j)}^{\textbf{k}_{(s,\ell-1)}}$
	for $j \ne \ell$, 
	it suffices to consider $\bm{W}_{\mathfrak{i}(\ell)}^{\textbf{k}_{(s,\ell)}}$.
	Suppose the learning rates satisfy \eqref{LR-l2-loss-exact}.
	It follows from the BCGD updates
	\begin{align*}
	\bm{W}_{\mathfrak{i}(\ell)}^{\textbf{k}_{(s,\ell)}}
	&=
	\bm{W}_{\mathfrak{i}(\ell)}^{\textbf{k}_{(s,\ell-1)}}
	-\eta_\ell^{\textbf{k}_{(s,\ell-1)}}
	(\bm{W}_{L:(\mathfrak{i}(\ell)+1)}^{\textbf{k}_{(s,\ell-1)}})^T
	\Delta^{\textbf{k}_{(s,\ell-1)}}
	(\bm{W}_{(\mathfrak{i}(\ell)-1):1}^{\textbf{k}_{(s,\ell-1)}}\bm{X})^T,
	\end{align*}
	that
	\begin{align*}
	&\|\bm{W}_{\mathfrak{i}(\ell)}^{\textbf{k}_{(s,\ell)}} - \bm{W}^{(0)}_{\mathfrak{i}(\ell)}\|
	\\
	&\le
	\|\bm{W}_{\mathfrak{i}(\ell)}^{\textbf{k}_{(s,\ell-1)}}- \bm{W}^{(0)}_{\mathfrak{i}(\ell)}\|
	+\eta_\ell^{\textbf{k}_{(s,\ell-1)}}
	\|\bm{W}_{L:(\mathfrak{i}(\ell)+1)}^{\textbf{k}_{(s,\ell-1)}}\|
	\|\bm{W}_{(\mathfrak{i}(\ell)-1):1}^{\textbf{k}_{(s,\ell-1)}}\bm{X}\|
	\|\Delta^{\textbf{k}_{(s,\ell-1)}}\|, \\
	&\le 
	\|\bm{W}_{\mathfrak{i}(\ell)}^{\textbf{k}_{(s,\ell-1)}}- \bm{W}^{(0)}_{\mathfrak{i}(\ell)}\|
	+
	\frac{\eta\|\Delta^{\textbf{k}_{(s,\ell-1)}}\|_F}{\|\bm{W}_{L:(\mathfrak{i}(\ell)+1)}^{\textbf{k}_{(s,\ell-1)}}\|
		\|\bm{W}_{(\mathfrak{i}(\ell)-1):1}^{\textbf{k}_{(s,\ell-1)}}\bm{X}\|}.
	\end{align*}
	Using \eqref{norm-lower-bound}, we obtain
	\begin{equation} \label{main-eqn-01}
	\|\bm{W}_{\mathfrak{i}(\ell)}^{\textbf{k}_{(s,\ell)}}-\bm{W}^{(0)}_{\mathfrak{i}(\ell)}\|
	\le
	\|\bm{W}_{\mathfrak{i}(\ell)}^{\textbf{k}_{(s,\ell-1)}}-\bm{W}^{(0)}_{\mathfrak{i}(\ell)}\|
	+
	\eta\frac{
		\|\bm{W}_{L:\mathfrak{i}(\ell)}^{\textbf{k}_{(s,\ell-1)}}\|
		\|\bm{W}_{\mathfrak{i}(\ell):1}^{\textbf{k}_{(s,\ell-1)}}\bm{X}\|
		\|\Delta^{\textbf{k}_{(s,\ell-1)}}\|_F}{c^2}.
	\end{equation}
	
%
	
	Also, note that by the induction hypothesis and \eqref{matrix-update-balanced}, we have 
	$\sigma_{\max}(\bm{W}_{j}^{\textbf{k}_{(s,\ell-1)}}) < 1 + R$ and
	\begin{equation} \label{ineq-min-sing}
	\begin{split}
	R &> \|\bm{W}_{j}^{\textbf{k}_{(s,\ell-1)}} - \bm{W}_{j}^{(0)}\|  
	\ge \|(\bm{W}_{j}^{\textbf{k}_{(s,\ell-1)}} - \bm{W}_{j}^{(0)})z\| \\
	&\ge \|\bm{W}_{j}^{(0)}z\| - \|\bm{W}_{j}^{\textbf{k}_{(s,\ell-1)}}z\| 
	= 
	1-\sigma_{\min}(\tilde{\bm{W}}_{j}^{\textbf{k}_{(s,\ell-1)}}) \\
	\implies& \sigma_{\min}(\tilde{\bm{W}}_{j}^{\textbf{k}_{(s,\ell-1)}}) > 1 - R.
	\end{split}
	\end{equation}
	where $\|z\|=1$.
	Here, we set $z$ to be the right singular vector of $\bm{W}_j^{\textbf{k}_{(s,\ell-1)}}$
	which corresponds to $\sigma_{\min}(\tilde{\bm{W}}_j^{\textbf{k}_{(s,\ell-1)}})$.
	Then, $z$ has zero-values from ($\max\{n_0,n_L\} +1$)-th to $n_{j-1}$-th entries.
	Recall that $\bm{W}_j^{(0)}$ is equivalent to an orthogonal matrix upto zero-valued padding. 
	This allows us to conclude $\|\bm{W}_{j}^{(0)}z\| = 1$, which makes the fourth equality of \eqref{ineq-min-sing} hold.
	Thus, we have 
	\begin{align*}
	\sigma_{\min}(\bm{W}_{L:(\mathfrak{i}(\ell)+1)}^{\textbf{k}_{(s,\ell-1)}})\sigma_{\min}(\bm{W}_{(\mathfrak{i}(\ell)-1):1}^{\textbf{k}_{(s,\ell-1)}}\bm{X})
	&\ge \sigma_{\min}(\bm{X})(1-R)^{L-1}, 
	\\
	\sigma_{\max}(\bm{W}_{L:(\mathfrak{i}(\ell)+1)}^{\textbf{k}_{(s,\ell-1)}})\sigma_{\max}(\bm{W}_{(\mathfrak{i}(\ell)-1):1}^{\textbf{k}_{(s,\ell-1)}}\bm{X})
	&\le \sigma_{\max}(\bm{X})(1+R)^{L-1}. 
	\end{align*}
	It then follows from \eqref{Rate-l2-loss-exact} that
	\begin{equation} \label{rate-bound}
	\gamma^{\textbf{k}_{(k,j-1)}} =1 -\frac{\eta}{\kappa^2(\bm{W}_{L:(\mathfrak{i}(\ell)+1)}^{\textbf{k}_{(k,\ell-1)}})\kappa^2(\bm{W}_{(\mathfrak{i}(\ell)-1):1}^{\textbf{k}_{(s,\ell-1)}}\bm{X})}< \gamma := 1 - \frac{\eta}{\kappa^2(\bm{X})}
	\left(\frac{1-R}{1+R}\right)^{2(L-1)},
	\end{equation}
	for $0 \le k < s$ with $1 \le j \le L$
	and $k=s$ with $1 \le j < \ell$.
	From \eqref{main-eqn-01}, \eqref{rate-bound} and Theorem~\ref{thm:convg-l2}, we obtain
	\begin{align*}
	\|\bm{W}_{\mathfrak{i}(\ell)}^{(s+1)}-\bm{W}^{(0)}_{\mathfrak{i}(\ell)}\|
	\le
	\|\bm{W}_{\mathfrak{i}(\ell)}^{(s)}-\bm{W}^{(0)}_{\mathfrak{i}(\ell)}\|
	+
	\frac{\eta(1+R)^{L+1}}{c^2}\|\bm{X}\|\|\Delta^{\textbf{k}_{0}}\|_F
	\gamma^{sL+\ell-1}.
	\end{align*}
	The recursive relation with respect to $s$ gives
	\begin{align*}
	\|\bm{W}_{\mathfrak{i}(\ell)}^{(s+1)}-\bm{W}^{(0)}_{\mathfrak{i}(\ell)}\|
	&\le \sum_{t=0}^{s} \frac{(1+R)^{L+1}}{c^2}\eta \|\bm{X}\|\|\Delta^{\textbf{k}_{0}}\|_F
	\gamma^{tL+\ell-1}
	\\
	&\le
	\frac{(1+R)^{L+1}}{c^2}\eta \|\bm{X}\|\|\Delta^{\textbf{k}_{0}}\|_F\frac{1}{1-\gamma^{L}}
	\\
	&\le
	\frac{(1+R)^{L+1}}{c^2}\eta \|\bm{X}\|\|\Delta^{\textbf{k}_{0}}\|_F\frac{1}{L\frac{\eta}{\kappa^2_{r_x}(\bm{X})}
		\left(\frac{1-R}{1+R}\right)^{2(L-1)}} \\
	&\le \frac{\|\bm{X}\|^2\|\bm{W}^{\textbf{k}_{0}}-\bm{W}^*\|_F\kappa^2(\bm{X})}{c^2}
	\frac{(1+R)^{L+1}}{L\left(\frac{1-R}{1+R}\right)^{2(L-1)}}.
	\end{align*}
	Let $\tilde{c} = c/\|\bm{X}\|$.
	If $R = R_L:=\frac{(5L-3)-\sqrt{(5L-3)^2-4L}}{2L}$ 
	and 
	\begin{equation} \label{assumption-c}
	\tilde{c} \ge \kappa^2(\bm{X})\left(\frac{-1+\sqrt{1+4h(L)\tilde{\sigma}_{\min}/\kappa^2(\bm{X})}}{2h(L)}\right),
	\end{equation}
	where $h(L) = \frac{LR_L(1-R_L)^{2L-2}}{(1+R_L)^{3L-1}}$,
	we have
	$$
	\|\bm{W}_{\mathfrak{i}(\ell)}^{(s+1)}-\bm{W}^{(0)}_{\mathfrak{i}(\ell)}\|\le 
	\frac{\|\bm{W}^{\textbf{k}_{0}}-\bm{W}^*\|_F\kappa^2(\bm{X})}{\tilde{c}^2}
	\frac{(1+R)^{L+1}}{L\left(\frac{1-R}{1+R}\right)^{2(L-1)}} \le R.
	$$
	This can be checked as follow.
	First, we note that the maximum of $x\frac{\left(\frac{1-x}{1+x}\right)^{2(L-1)}}{(1+x)^{L+1}}$
	where $0 < x <1$ is obtained at $x=R_L$.
	It also follows from the assumption of $\|\bm{W}^{\textbf{k}_{0}}-\bm{W}^*\|_F \le \tilde{\sigma}_{\min} -\tilde{c}$ that 
	\begin{align*}
	&\tilde{c} \ge \kappa^2(\bm{X})\left(\frac{-1+\sqrt{1+4h(L)\tilde{\sigma}_{\min}/\kappa^2(\bm{X})}}{2h(L)}\right) \\
	&\iff
	\frac{2h(L)}{\kappa^2(\bm{X})}\tilde{c} + 1 \ge \sqrt{1+4h(L)\tilde{\sigma}_{\min}/\kappa^2(\bm{X})}
	\\
	&\implies \frac{(\tilde{\sigma}_{\min} -\tilde{c})\kappa^2(\bm{X})}{\tilde{c}^2}
	\le h(L)=LR\frac{(1-R)^{2(L-1)}}{(1+R)^{3L-1}} 
	\\
	&\implies \frac{\|\bm{W}^{\textbf{k}_{0}}-\bm{W}^*\|_F\kappa^2(\bm{X})}{\tilde{c}^2}
	\le LR\frac{\left(\frac{1-R}{1+R}\right)^{2(L-1)}}{(1+R)^{L+1}} 
	\\
	&\iff 
	\frac{\|\bm{W}^{\textbf{k}_{0}}-\bm{W}^*\|_F\kappa^2(\bm{X})}{\tilde{c}^2}
	\frac{(1+R)^{L+1}}{L\left(\frac{1-R}{1+R}\right)^{2(L-1)}} \le R.
	\end{align*}
	Hence, $\|\bm{W}_{\mathfrak{i}(\ell)}^{(s+1)} - \bm{W}^{(0)}_{\mathfrak{i}(\ell)}\| < R$.
	Thus, by induction, we conclude that
	$\mathcal{R}(k) < R$ for all $k$.
	
	By letting $\tilde{c} = \kappa^2(\bm{X})\left(\frac{-1+\sqrt{1+4h(L)\tilde{\sigma}_{\min}/\kappa^2(\bm{X})}}{2h(L)}\right)$, 
	the assumption on $\|\bm{W}^{\textbf{k}_0}-\bm{W}^*\|_F$ becomes 
	\begin{align*}
	\|\bm{W}^{\textbf{k}_0}-\bm{W}^*\|_F &\le 
	\tilde{\sigma}_{\min}
	- \kappa^2(\bm{X})\left(\frac{-1+\sqrt{1+4h(L)\tilde{\sigma}_{\min}/\kappa^2(\bm{X})}}{2h(L)}\right) \\
	&=  \frac{1}{2h(L)}\left(2h(L)\tilde{\sigma}_{\min} +\kappa^2(\bm{X})-\kappa(\bm{X})\sqrt{\kappa^2(\bm{X})+4h(L)\tilde{\sigma}_{\min}}\right)
	\\
	&=\frac{1}{2h(L)}\cdot \frac{(2h(L)\tilde{\sigma}_{\min} +\kappa^2(\bm{X}))^2-\kappa^2(\bm{X})(\kappa^2(\bm{X})+4h(L)\tilde{\sigma}_{\min})}{2h(L)\tilde{\sigma}_{\min}  +\kappa^2(\bm{X})+\kappa(\bm{X})\sqrt{\kappa^2(\bm{X})+4h(L)\tilde{\sigma}_{\min}}}
	\\
	&=
	\frac{2h(L)\tilde{\sigma}_{\min}}{2h(L)\tilde{\sigma}_{\min} +\kappa^2(\bm{X})\left(1+\sqrt{1+4h(L)\tilde{\sigma}_{\min}/\kappa^2(\bm{X})}\right)}
	\\
	&= \frac{\tilde{\sigma}_{\min}}{1 +  \kappa^2(\bm{X})\left(\frac{1+\sqrt{1+4h(L)\tilde{\sigma}_{\min}/\kappa^2(\bm{X})}}{2h(L)\tilde{\sigma}_{\min}}\right)}.
	\end{align*} 
	Therefore, under the above assumption on $\|\bm{W}^{\textbf{k}_0}-\bm{W}^*\|_F$,
	we have 
	\begin{align*}
	\gamma^{\textbf{k}_{(k,\ell-1)}} < \gamma_L := 1 - \frac{\eta}{\kappa^2(\bm{X})}
	\left(\frac{1-R_L}{1+R_L}\right)^{2(L-1)}.
	\end{align*}
	Furthermore, it follows from
	\begin{multline*}
	LR_L = \frac{5L-3}{2}\left(1 - \sqrt{1-\frac{4L}{(5L-3)^2}}\right)
	\\
	= \frac{\frac{2L}{5L-3}}{1 + \sqrt{1-\frac{4L}{(5L-3)^2}}}
	= \frac{2}{5-3/L}\cdot \frac{1}{1+\sqrt{1-\frac{4L}{(5L-3)^2}}},
	\end{multline*}
	that $\lim_{L\to \infty} LR_L = \frac{1}{5}$ and $\lim_{L\to \infty} R_L = 0$.
	Also, since $LR_L$ and $R_L$ are decreasing functions of $L$, we have
	\begin{align*}
	\left(\frac{1-R_L}{1+R_L}\right)^{2(L-1)} \ge \left(1 - \frac{2R_L}{1+R_L}\right)^{2L} \ge 1 - \frac{4LR_L}{1+R_L} \ge \frac{1}{5}.
	\end{align*}
	Hence, we can conclude that 
	\begin{align*}
	\gamma_L = 1 - \frac{\eta}{\kappa^2(\bm{X})}\frac{4LR_L}{1+R_L}
	\le \gamma = 1 - \frac{\eta}{5\kappa^2(\bm{X})},
	\end{align*}
	which completes the proof.
\end{proof}

\section{Proof of Theorem~\ref{thm:l2-dout1}}
\label{app:thm:l2-dout1}
\begin{proof}
	Since $n_\ell \ge \max\{n_0, n_L\}$ 
	and the initial weight matrices are from the orth-identity initialization (Section~\ref{subsec:initialization}), 
	it follows from Theorem~\ref{thm:role of width}
	that $\bm{W}_{(\ell-1):1}^{(0)} \approxeq_1 \tilde{\bm{W}}^{(0)}_{(\ell-1):1} \in \mathbb{R}^{\max\{n_0,n_L\}\times n_0}$
	and 
	$(\bm{W}_{(\ell-1):1}^{(0)})^T\bm{W}_{(\ell-1):1}^{(0)} = \bm{I}_{n_0}$.
	Thus, 
    $$
    \sigma_{\max}(\bm{W}_{(\ell-1):1}^{(0)}) = 1 = \sigma_{\min}(\bm{W}_{(\ell-1):1}^{(0)}).
    $$ 
    Note that since $\bm{X}$ is a full row-rank matrix, $\bm{X}\bm{X}^T$ is invertible. 
    In what follows, we will show that $\|\bm{W}_{L:(L-\ell+1)}^{(1)}\| = 0$
    if 
    \begin{equation} \label{condition-non-zero}
        \bm{W}^* = \bm{Y}\bm{X}^\dagger = 
    \bm{W}^{\textbf{k}_{(0,\ell-1)}}\left(\bm{I}_{n_0} - \|\bm{X}\|^2(\bm{XX}^T)^{-1}/\eta \right).
    \end{equation}
%
    Suppose $\bm{W}^*$ does not satisfy the condition of \eqref{condition-non-zero} for all $\ell$.
    For $\ell = 1$, we have $\eta_1^{\textbf{k}_{(0,0)}} = \eta/\|\bm{X}\|^2$ since $(\bm{W}_{(L-1):1}^{(0)})^T\bm{W}_{(L-1):1}^{(0)} = \bm{I}_{n_0}$.
    Suppose $\bm{W}_L^{(1)} = \bm{0}$
    and let $\Delta^{\textbf{k}_{(0,\ell-1)}}_{W} = \bm{W}^{\textbf{k}_{(0,\ell-1)}} - \bm{W}^*$.
    Then, 
    \begin{align*}
        \bm{0} = \bm{W}_L^{(1)} &= \bm{W}_L^{(0)} - \eta_1^{\textbf{k}_{(0,0)}}
    (\bm{W}_{(L-1):1}^{(0)}\bm{XX}^T(\Delta_W^{\textbf{k}_{(0,0)}})^T)^T, \\
    \bm{W}_L^{(0)}&= \eta_1^{\textbf{k}_{(0,0)}}
    \Delta_W^{\textbf{k}_{(0,0)}}
    \bm{XX}^T
    (\bm{W}_{(L-1):1}^{(0)})^T,
    \\
    \bm{W}^{\textbf{k}_{(0,0)}}&=
    \eta_1^{\textbf{k}_{(0,0)}}
     (\bm{W}^{\textbf{k}_{(0,0)}}-\bm{W}^*)
     \bm{XX}^T
     (\bm{W}_{(L-1):1}^{(0)})^T\bm{W}_{(L-1):1}^{(0)}, \\
     \bm{W}^* &= \bm{W}^{\textbf{k}_{(0,0)}}\left(\bm{I}_{n_0} - (\eta_1^{\textbf{k}_{(0,0)}}\bm{XX}^T)^{-1} \right),
    \end{align*}
    which contradicts to the assumption of $\bm{W}^*$ being not satisfying \eqref{condition-non-zero}.
    Hence, $\bm{W}_L^{(1)} \ne \bm{0}$.
    Now, suppose $\|\bm{W}_{L:(L-\ell+2)}^{(1)}\| \ne 0$
    and we want to show $\|\bm{W}_{L:(L-\ell+1)}^{(1)}\| \ne 0$.
    Suppose not, i.e, $\bm{W}_{L:(L-\ell+1)}^{(1)} = \bm{0}$.
    Then, we have 
    \begin{align*}
        \bm{W}_{L-\ell+1}^{(1)} &= \bm{W}_{L-\ell+1}^{(0)} - \eta_\ell^{\textbf{k}_{(0,\ell-1)}}
    (\bm{W}_{(L-\ell):1}^{(0)}\bm{XX}^T(\Delta_W^{\textbf{k}_{(0,\ell-1)}})^T\bm{W}_{L:(L-\ell+2)}^{(1)}
    )^T, \\
    \bm{0} = \bm{W}_{L:(L-\ell+1)}^{(1)} &= \bm{W}_{L:(L-\ell+2)}^{(1)}\bm{W}_{L-\ell+1}^{(0)} \\
    &\qquad\quad- \eta_\ell^{\textbf{k}_{(0,\ell-1)}}
    \|\bm{W}_{L:(L-\ell+2)}^{(1)}\|^2\Delta_W^{\textbf{k}_{(0,\ell-1)}}
    \bm{XX}^T(\bm{W}_{(L-\ell):1}^{(0)})^T, \\
    \bm{W}_{L:(L-\ell+2)}^{(1)}\bm{W}_{L-\ell+1}^{(0)}&= \eta_1^{\textbf{k}_{(0,0)}}
    \Delta_W^{\textbf{k}_{(0,\ell-1)}}
    \bm{XX}^T
    (\bm{W}_{(L-\ell):1}^{(0)})^T,
    \\
    \bm{W}^{\textbf{k}_{(0,\ell-1)}}&= \eta_1^{\textbf{k}_{(0,0)}}
    \Delta_W^{\textbf{k}_{(0,\ell-1)}}
    \bm{XX}^T
    (\bm{W}_{(L-1):1}^{(0)})^T\bm{W}_{(L-\ell):1}^{(0)},
    \\
    \bm{W}^* &= \bm{W}^{\textbf{k}_{(0,\ell-1)}}\left(\bm{I}_{n_0} - (\eta_1^{\textbf{k}_{(0,0)}}\bm{XX}^T)^{-1} \right),
    \end{align*}
    which contradicts to the assumption of $\bm{W}^*$.
    Hence, $\bm{W}_{L:(L-\ell+1)}^{(1)} \ne \bm{0}$.
    By induction, we conclude that 
    $\bm{W}_{L:(L-\ell+1)}^{(1)} \ne \bm{0}$ for all $\ell$.
    Thus, it follows from Theorem~\ref{thm:convg-l2}
    that 
    \begin{equation}
        \begin{split}
        \|\bm{W}^{\textbf{k}_{1}}\bm{X} - \bm{W}^*\bm{X}\|_F
        &<
        \|\bm{W}^{\textbf{k}_{0}}\bm{X} - \bm{W}^*\bm{X}\|_F
        \left(1 -\frac{\eta}{\kappa^2(\bm{X})} \right)^{L}.
        \end{split}
    \end{equation} 
    Since $L$ is chosen to satisfy 
    $$
    \|\bm{W}^{\textbf{k}_{1}}\bm{X} - \bm{W}^*\bm{X}\|_F \le 
    \|\bm{W}^{\textbf{k}_{0}}\bm{X} - \bm{W}^*\bm{X}\|_F
        \left(1 -\frac{\eta}{\kappa^2(\bm{X})} \right)^{L}
        \le \frac{\sigma_{\min}(\bm{W}^*\bm{X})}{c},
    $$
    where $c$ is defined in \eqref{def:c-min},
    it follows from Lemma~\ref{lemma-min-sing-value}
    and Theorem~\ref{thm-l2-identity}
    that 
    $\|\bm{W}_{L:j}^{(s)}\| \ne 0$ for all $j$ and $s$,
    and 
    $$
    \|\bm{W}^{\textbf{k}_{s}}\bm{X} - \bm{W}^*\bm{X}\|_F \le 
    \|\bm{W}^{\textbf{k}_{1}}\bm{X} - \bm{W}^*\bm{X}\|_F(\gamma^{L-1})^{s-1}\left(1 -\frac{\eta}{\kappa^2(\bm{X})} \right)^{s-1}.
    $$
    Note that $\left(1 -\frac{1}{\kappa^2(\bm{X})} \right)^{s-1}$ is from the fact that 
    $\|\bm{W}_{L:2}^{(s)}\| \ne 0$ for all $1 \le s$.
    Hence, we have 
    $$
    \|\bm{W}^{\textbf{k}_{s}}\bm{X} - \bm{W}^*\bm{X}\|_F \le 
    \|\bm{W}^{\textbf{k}_{0}}\bm{X} - \bm{W}^*\bm{X}\|_F(\gamma^{L-1})^{s-1}\left(1 -\frac{\eta}{\kappa^2(\bm{X})} \right)^{L+s-1},
    $$
    and the proof is completed.
\end{proof}

\section{Proof of Theorem~\ref{thm:convg-convex}}
\label{app:thm:convg-convex}

\begin{proof}
	For notational convenience, let $\ell(z) = \ell(z;b)$.
	Since $\ell(\cdot)$ is convex, differentiable
	and $|\ell'(z)-\ell'(x)|\le C_\text{Lip}|z-x|$, we have
	\begin{align*}
	\ell(z) \le \ell(x) + \ell'(x)(z-x) + \frac{1}{2}\ell''(x)(z-x)^2
	\le  \ell(x) + \ell'(x)(z-x) + \frac{C_\text{Lip}}{2}(z-x)^2.
	\end{align*}
	Let 
	$\bm{W}^{\textbf{k}_{(k,\ell)}} = \bm{W}_{L:(L-\ell+1)}^{(k+1)}\bm{W}_{(L-\ell):1}^{(k)}$, 
	$\hat{\textbf{y}}_{(k,\ell)}^i =\bm{W}^{\textbf{k}_{(k,\ell)}}\x^i$
	and 
	$$\hat{\textbf{Y}}^{(k,\ell)}=[\hat{\textbf{y}}_{(k,\ell)}^1,\cdots,\hat{\textbf{y}}_{(k,\ell)}^m]= \bm{W}^{\textbf{k}_{(k,\ell)}}\bm{X}.$$
	Then, we have
	\begin{equation} \label{thm:loss-ineq}
	\begin{split}
	&\ell((\hat{\textbf{y}}_{(k,\ell)}^i)_j; \textbf{y}_j^i) 
	\\
	&\le 
	\ell((\hat{\textbf{y}}_{(k,\ell-1)}^i)_j; \textbf{y}_j^i)
	+\ell'((\hat{\textbf{y}}_{(k,\ell-1)}^i)_j; \textbf{y}_j^i)
	((\hat{\textbf{y}}_{(k,\ell)}^i)_j - (\hat{\textbf{y}}_{(k,\ell-1)}^i)_j)
	\\
	&\qquad+ \frac{C_\text{Lip}}{2}((\hat{\textbf{y}}_{(k,\ell)}^i)_j - (\hat{\textbf{y}}_{(k,\ell-1)}^i)_j)^2.
	\end{split}
	\end{equation}
	For notational convenience, for $j > i$, let
	$$
	\bm{W}_j \bm{W}_{j-1} \cdots \bm{W}_{i} = \bm{W}_{j:i}.
	$$
	It follows from the BCGD update rule that
	\begin{align*}
	\bm{W}_{L-\ell+1}^{(k+1)}
	= 
	\bm{W}_{L-\ell+1}^{(k)}
	- \eta_\ell^{\textbf{k}_{(k,\ell-1)}}
	(\bm{W}_{(L-\ell):1}^{(k)}
	\bm{X} \bm{J}^{\textbf{k}_{(k,\ell-1)}}\bm{W}_{L:(L-\ell+2)}^{(k+1)})^T.
	\end{align*}
	By multiplying $\bm{W}_{(L-\ell):1}^{(k)}\bm{X}$ from right, 
	and $\bm{W}_{L:(L-\ell+2)}^{(k+1)}$ from left in the both sides, 
	we obtain
	\begin{multline*}
	\bm{W}_{L:(L-\ell+1)}^{(k+1)}\bm{W}_{(L-\ell):1}^{(k)}\bm{X} \\
	= 
	\bm{W}_{L:(L-\ell+2)}^{(k+1)}\bm{W}_{(L-\ell+1):1}^{(k)}\bm{X}
	- \eta_\ell^{\textbf{k}_{(k,\ell-1)}}
	\bm{A}_\ell^{(k+1)}
	(\bm{J}^{\textbf{k}_{(k,\ell-1)}})^T
	\bm{C}_\ell^{(k)},
	\end{multline*}
	where
	\begin{align*}
	\bm{A}_\ell^{(k)} &= \bm{W}_{L:(L-\ell+2)}^{(k)}
	(\bm{W}_{L:(L-\ell+2)}^{(k)})^T \in \mathbb{R}^{d_{\text{out}} \times d_{\text{out}}},
	\\
	\bm{B}_\ell^{(k)} &= (\bm{W}_{(L-\ell):1}^{(k)}\bm{X})^T
	\bm{W}_{(L-\ell):1}^{(k)}\bm{X}
	\in \mathbb{R}^{m \times m}.
	\end{align*}
	Thus, we have
	\begin{align*}
	\hat{\textbf{Y}}_{(k,\ell)} - \hat{\textbf{Y}}_{(k,\ell-1)}
	&= (\bm{W}^{\textbf{k}_{(k,\ell)}} - \bm{W}^{\textbf{k}_{(k,\ell-1)}})\bm{X}
	= - \eta_\ell^{\textbf{k}_{(k,\ell-1)}}
	\bm{A}_\ell^{(k)}
	(\bm{J}^{\textbf{k}_{(k,\ell-1)}})^T
	\bm{B}_\ell^{(k+1)},
	\end{align*}
	where $\hat{\textbf{Y}}_{(k,\ell)} = \bm{W}^{\textbf{k}_{(k,\ell)}}\bm{X}$.
	Let
	$$
	\mu_{\max}^{(k,\ell-1)} = 
	\sigma_{\max}^2(\bm{W}_{L:(L-\ell+2)}^{(k+1)})
	\sigma_{\max}^2(\bm{W}_{(L-\ell):1}^{(k)}\bm{X}).
	$$
	Also, let $\Delta \mathcal{L}^{\textbf{k}_{(k,\ell)}} = \mathcal{L}(\bm{\theta}^{\textbf{k}_{(k,\ell)}})
	- 
	\mathcal{L}(\bm{\theta}^{*})$,
	$\Delta^{\textbf{k}_{(k,\ell)}} = (\bm{W}^{\textbf{k}_{(k,\ell-1)}}-\bm{W}^*)\bm{X}$,
	and 
	\begin{align*}
	\mathcal{J}^{\textbf{k}_{(k,\ell-1)}} &=
	(\bm{W}_{(L-\ell):1}^{(k)}\bm{X}\bm{J}^{\textbf{k}_{(k,\ell-1)}}\bm{W}_{L:(L-\ell+2)}^{(k+1)})^T, \\
	\tilde{\mathcal{J}}^{\textbf{k}_{(k,\ell-1)}}&=\bm{W}_{L:(L-\ell+1)}^{(k+1)}
	\mathcal{J}^{\textbf{k}_{(k,\ell-1)}}
	\bm{W}_{(L-\ell):1}^{(k)}\bm{X}
	= \bm{A}_\ell^{(k+1)}
	(\bm{J}^{\textbf{k}_{(k,\ell-1)}})^T
	\bm{B}_\ell^{(k)},
	\end{align*}
	Let $\mathcal{L}(\hat{\textbf{Y}}_{(k,\ell)}) = \mathcal{L}(\bm{\theta}^{\textbf{k}_{(k,\ell)}})$.
	By combining it with \eqref{thm:loss-ineq}, 
	\begin{align*}
	\mathcal{L}(\hat{\textbf{Y}}_{(k,\ell)})
	&\le 
	\mathcal{L}(\hat{\textbf{Y}}_{(k,\ell-1)})
	-\eta_\ell^{\textbf{k}_{(k,\ell-1)}}
	\langle (\bm{J}^{\textbf{k}_{(k,\ell-1)}})^T, \bm{A}_\ell^{(k+1)}
	(\bm{J}^{\textbf{k}_{(k,\ell-1)}})^T
	\bm{B}_\ell^{(k)}\rangle_F
	\\
	&\qquad+\frac{C_\text{Lip}}{2}(\eta_\ell^{\textbf{k}_{(k,\ell-1)}})^2
	\|\bm{A}_\ell^{(k+1)}
	(\bm{J}^{\textbf{k}_{(k,\ell-1)}})^T
	\bm{B}_\ell^{(k)}\|_F^2 \\
	&=\mathcal{L}(\hat{\textbf{Y}}_{(k,\ell-1)})
	-\eta_\ell^{\textbf{k}_{(k,\ell-1)}}
	\| \mathcal{J}^{\textbf{k}_{(k,\ell-1)}}\|_F^2
	\\
	&\qquad+\frac{C_\text{Lip}}{2}(\eta_\ell^{\textbf{k}_{(k,\ell-1)}})^2\|\bm{A}_\ell^{(k+1)}
	(\bm{J}^{\textbf{k}_{(k,\ell-1)}})^T
	\bm{B}_\ell^{(k)}\|_F^2.
	\end{align*}
	It then can be checked that the learning rate which minimizes the above upper bound is 
	\begin{equation}
	\eta_\text{opt}^{\textbf{k}_{(k,\ell-1)}} = 
	\frac{\| \mathcal{J}^{\textbf{k}_{(k,\ell-1)}}\|_F^2}
	{C_\text{Lip}
		\|\bm{W}_{L:(\ell-1)}^{(k+1)}
		\mathcal{J}^{\textbf{k}_{(k,\ell-1)}}
		\bm{W}_{(\ell-1):1}^{(k)}\bm{X}\|_F^2}.
	\end{equation}
	Also, it follows from 
	\begin{equation} \label{gen-loss-grad-ineq}
	\begin{split}
	\|\tilde{\mathcal{J}}^{\textbf{k}_{(k,\ell-1)}}\|_F^2
	&\le \sigma_{\max}^2(\bm{W}_{L:(L-\ell+1)}^{(k+1)})\sigma_{\max}^2(\bm{W}_{(L-\ell):1}^{(k)}\bm{X})
	\|{\mathcal{J}}^{\textbf{k}_{(k,\ell-1)}}\|_F^2
	\\
	&= \mu_{\max}^{(k,\ell-1)}\|{\mathcal{J}}^{\textbf{k}_{(k,\ell-1)}}\|_F^2
	\end{split}
	\end{equation}
	that 
	\begin{align*}
	&\mathcal{L}(\hat{\textbf{Y}}_{(k,\ell)})
	\\
	&\le \mathcal{L}(\hat{\textbf{Y}}_{(k,\ell-1)})
	-\eta_\ell^{\textbf{k}_{(k,\ell-1)}}
	\| \mathcal{J}^{\textbf{k}_{(k,\ell-1)}}\|_F^2
	+\frac{C_\text{Lip}}{2}(\eta_\ell^{\textbf{k}_{(k,\ell-1)}})^2\|\bm{A}_\ell^{(k+1)}
	(\bm{J}^{\textbf{k}_{(k,\ell-1)}})^T
	\bm{B}_\ell^{(k)}\|_F^2
	\\
	&\le\mathcal{L}(\hat{\textbf{Y}}_{(k,\ell-1)})
	-\eta_\ell^{\textbf{k}_{(k,\ell-1)}}
	\| \mathcal{J}^{\textbf{k}_{(k,\ell-1)}}\|_F^2+\frac{C_\text{Lip}}{2}(\eta_\ell^{\textbf{k}_{(k,\ell-1)}})^2
	\mu_{\max}^{(k,\ell-1)}
	\|\mathcal{J}^{\textbf{k}_{(k,\ell-1)}}\|_F^2
	\\
	&=
	\mathcal{L}(\hat{\textbf{Y}}_{(k,\ell-1)})
	-(1-\frac{C_\text{Lip}}{2}\eta_\ell^{\textbf{k}_{(k,\ell-1)}}
	\mu_{\max}^{(k,\ell-1)})
	\eta_\ell^{\textbf{k}_{(k,\ell-1)}}
	\| \mathcal{J}^{\textbf{k}_{(k,\ell-1)}}\|_F^2.
	\end{align*}
	If $ 
	0< \eta_\ell^{\textbf{k}_{(k,\ell-1)}} < \frac{2}{C_\text{Lip}\mu_{\max}^{(k,\ell-1)}},
	$
	unless $\|\mathcal{J}^{\textbf{k}_{(k,\ell-1)}}\|_F=0$,
	the loss function is strictly decreasing.
	
	Suppose $ 
	0< \eta_\ell^{\textbf{k}_{(k,\ell-1)}} \le \frac{1}{C_\text{Lip}\mu_{\max}^{(k,\ell-1)}}.
	$
	Then, since 
	$
	-(1- \frac{C_\text{Lip}}{2}\eta_\ell^{\textbf{k}_{(k,\ell-1)}}\mu_{\max}^{(k,\ell-1)} )
	\le -\frac{1}{2},
	$
	we have
	\begin{align*}
	\mathcal{L}(\hat{\textbf{Y}}_{(k,\ell)})
	&\le 
	\mathcal{L}(\hat{\textbf{Y}}_{(k,\ell-1)})
	-\frac{\eta_\ell^{\textbf{k}_{(k,\ell-1)}}}{2}
	\| \mathcal{J}^{\textbf{k}_{(k,\ell-1)}}\|_F^2.
	\end{align*}
	By summing up the above, we have
	\begin{align*}
	\sum_{s=0}^{k-1} \sum_{\ell = 1}^L \frac{\eta_\ell^{\textbf{k}_{(s,\ell-1)}}}{2}
	\| \mathcal{J}^{\textbf{k}_{(s,\ell-1)}}\|_F^2
	&\le 
	\sum_{s=0}^{k-1} \sum_{\ell = 1}^L
	\left(
	\mathcal{L}(\hat{\textbf{Y}}_{(s,\ell-1)}) - \mathcal{L}(\hat{\textbf{Y}}_{(s,\ell)}) \right)
	\le \mathcal{L}(\hat{\textbf{Y}}_{(0,0)}) < \infty.
	\end{align*}
	Therefore, 
	$\lim_{k \to \infty} \eta_\ell^{\textbf{k}_{(k,\ell)}}
	\| \mathcal{J}^{\textbf{k}_{(k,\ell)}}\|_F^2 = 0$ for any $0 \le \ell < L$.
	Also, it follows from the above that 
	$$
	\frac{1}{k}\sum_{s=0}^{k-1} 
	\eta_\ell^{\textbf{k}_{(s,\ell)}}
	\| \mathcal{J}^{\textbf{k}_{(s,\ell)}}\|_F^2
	\le
	\frac{1}{k}\sum_{s=0}^{k-1} \sum_{\ell = 1}^L
	\eta_\ell^{\textbf{k}_{(s,\ell)}}
	\| \mathcal{J}^{\textbf{k}_{(s,\ell)}}\|_F^2
	\le \frac{2}{k}\mathcal{L}(\hat{\textbf{Y}}_{(0,0)}) = \mathcal{O}\left(\frac{1}{k}\right).
	$$
	Furthermore, for all $0 \le \ell < L$, 
	if $0<\inf_k \eta_\ell^{\textbf{k}_{(k,\ell)}} \le  \sup_k \eta_\ell^{\textbf{k}_{(k,\ell)}} \le 1$, 
	we conclude that $\lim_{k \to \infty}
	\| \mathcal{J}^{\textbf{k}_{(k,\ell)}}\|_F^2 = 0$
	and 
	$\lim_{k \to \infty}
	\|\eta_\ell^{\textbf{k}_{(k,\ell)}} \mathcal{J}^{\textbf{k}_{(k,\ell)}}\|_F^2 = 0$.
	For each $\ell$, 
	$\lim_{k\to \infty} \bm{W}_\ell^{(k)} = \bm{W}_\ell^*$.
	That is, the BCGD finds a critical point.
	Since all local minima are global (see, \cite{Laurent2018deep}),
	$\{\bm{W}_\ell^*\}_{\ell=1}^L$ is a global minimizer.
\end{proof}

\section{Proof of Theorem~\ref{thm:convg-l2-loss-BCSGD}}
\label{app:thm:convg-l2-loss-BCSGD}
\begin{proof}
For notational convenience, for $j > i$, let
$$
\bm{W}_{j:i}:=\bm{W}_j \bm{W}_{j-1} \cdots \bm{W}_{i}.
$$
By definition, 
it follows from the update rule that
\begin{align*}
    \bm{W}_{\ell_k}^{(\textbf{k}_k(\ell_k)+1)}
    = 
    \bm{W}_{\ell_k}^{(\textbf{k}_k(\ell_k))}
    - \eta^{\textbf{k}_{k}}_{\ell_k}
    (\bm{W}^{\textbf{k}_k}_{(\ell_k-1):1}
    \bm{X}_{:,i_k}\bm{J}_{i_k,:}^{\textbf{k}_k}
    \bm{W}^{\textbf{k}_k}_{L:(\ell_k+1)}
    )^T,
\end{align*}
where $i_k$ is randomly chosen indices from $[m]$ and 
and $\ell_k$ is an index from $[L]$.
By multiplying $\bm{W}_{(\ell_k-1):1}^{\textbf{k}_{k}}\bm{X}$ from right,  $\bm{W}_{L:(\ell_k+1)}^{\textbf{k}_{k}}$ from left
and subtracting $\bm{W}^*\bm{X}$ in the both sides, 
we obtain
\begin{align*}
    (\bm{W}^{\textbf{k}_{k+1}}- \bm{W}^*)\bm{X} = 
    (\bm{W}^{\textbf{k}_{k}} - \bm{W}^*)\bm{X}
    - \eta^{\textbf{k}_{k}}_{\ell_k}
    \bm{A}_{\ell_k}^{\textbf{k}_{k}}
    (\bm{X}_{:i_k}\bm{J}_{i_k:}^{\textbf{k}_k})^T
    \tilde{\bm{B}}_{\ell_k}^{\textbf{k}_{k}}\bm{X}
\end{align*}
where
\begin{align*}
    \bm{A}_{\ell_k}^{\textbf{k}_{k}} &= \bm{W}_{L:(\ell_k+1)}^{\textbf{k}_{k}}
    (\bm{W}_{L:(\ell_k+1)}^{\textbf{k}_{k}})^T \in \mathbb{R}^{d_{\text{out}} \times d_{\text{out}}},
    \\
    \tilde{\bm{B}}_{\ell_k}^{\textbf{k}_{k}} &= (\bm{W}_{(\ell_k-1):1}^{\textbf{k}_{k}})^T
    \bm{W}_{(\ell_k-1):1}^{\textbf{k}_{k}}
    \in \mathbb{R}^{d_{\text{in}} \times d_{\text{in}}}.
\end{align*}

Since $\bm{A}_{\ell_k}^{\textbf{k}_{k}}$ is symmetric, 
they are diagonalizable. Thus,
\begin{align*}
    (\bm{U}_{\ell_k}^{\textbf{k}_{k}})^{T}\bm{A}_{\ell_k}^{\textbf{k}_{k}}\bm{U}_{\ell_k}^{\textbf{k}_{k}} &=
    \bm{D}^{\textbf{k}_{k}}_{A,\ell_k} = \text{diag}(\lambda_{\ell_k, i}^{\textbf{k}_{k}}), 
    \quad 1 \le i \le  d_{\text{out}}, 
\end{align*}
where $\bm{U}_{\ell_k}^{\textbf{k}_{k}}$ is orthogonal.
Let
$\Delta_W^{\textbf{k}_{k}} := \bm{W}^{\textbf{k}_{k}}-\bm{W}^*$
and 
$
\Delta^{\textbf{k}_{k}} := \Delta_W^{\textbf{k}_{k}}\bm{X}.
$
Since $\ell(a;b)=(a-b)^2/2$, 
we have
\begin{align*}
    \Delta^{\textbf{k}_{k+1}} &= 
    \Delta^{\textbf{k}_{k}}
    - \eta^{\textbf{k}_{k}}_{\ell_k}
    \bm{A}_{\ell_k}^{\textbf{k}_{k}}
    \left(\Delta_W^{\textbf{k}_{k}}
    (\bm{X}_{:i_k}\bm{X}_{:i_k}^T)
    -(\y^{i_k} - \bm{W}^*\x^{i_k})\bm{X}_{:i_k}^T \right)
    \tilde{\bm{B}}_{\ell_k}^{\textbf{k}_{k}}\bm{X}.
\end{align*}
Let 
$\mathcal{E} = \bm{Y} - \bm{W}^*\bm{X}$ and
$\mathcal{E}_{:,i_k}:= \y^{i_k} - \bm{W}^*\x^{i_k}$.
Then
\begin{multline*}
    (\bm{U}_{\ell_k}^{\textbf{k}_{k}})^T\Delta^{\textbf{k}_{k+1}} 
    \\
    = 
    (\bm{U}_{\ell_k}^{\textbf{k}_{k}})^T\Delta^{\textbf{k}_{k}}
    - \eta^{\textbf{k}_{k}}_{\ell_k}
    \bm{D}_{A,\ell_k}^{\textbf{k}_k} (\bm{U}_{\ell_k}^{\textbf{k}_{k}})^{T}
    \left(\Delta_W^{\textbf{k}_{k}}
    (\bm{X}_{:i_k}\bm{X}_{:i_k}^T)
    -\mathcal{E}_{:,i_k}\bm{X}_{:i_k}^T \right)
    \tilde{\bm{B}}^{\textbf{k}_k}_{\ell_k}\bm{X}.
\end{multline*}
Let $\tilde{\Delta}^{\textbf{k}_{s,t,\ell}}= (\bm{U}_{\ell}^{\textbf{k}_{s}})^T \Delta^{\textbf{k}_{t}}$.
Then
\begin{equation} \label{eqn-BCSGD-err}
    \tilde{\Delta}^{\textbf{k}_{k,k+1,\ell_k}} = 
    \tilde{\Delta}^{\textbf{k}_{k,k,\ell_k}}
    - \eta^{\textbf{k}_{k}}_{\ell_k}
    \bm{D}_{A,\ell_k}^{\textbf{k}_k} (\bm{U}_{\ell_k}^{\textbf{k}_{k}})^T
    \left(\Delta_W^{\textbf{k}_{k}}
    (\bm{X}_{:i_k}\bm{X}_{:i_k}^T)
    -\mathcal{E}_{:,i_k}\bm{X}_{:i_k}^T \right)
    \tilde{\bm{B}}^{\textbf{k}_k}_{\ell_k}\bm{X}.
\end{equation}
Let $\bm{u}_{\ell_k, j}^{\textbf{k}_k}$ be the $j$-th column of
$\bm{U}_{\ell_k}^{\textbf{k}_{k}}$.
The $j$-th row of \eqref{eqn-BCSGD-err} satisfies 
\begin{align*}
    &\|(\tilde{\Delta}^{\textbf{k}_{k+1}})_{j:}\|^2 \\
    &= 
    \| (\bm{u}_{\ell_k, j}^{\textbf{k}_k})^T\Delta^{\textbf{k}_{k}}
    - \eta^{\textbf{k}_{k}}_{\ell_k}
    \lambda_{\ell_k, j}^{\textbf{k}_k}
    (\bm{u}_{\ell_k, j}^{\textbf{k}_k})^T
    \left(\Delta_W^{\textbf{k}_{k}}
    (\bm{X}_{:i_k}\bm{X}_{:i_k}^T)
    -\mathcal{E}_{:,i_k}\bm{X}_{:i_k}^T \right)
    \tilde{\bm{B}}^{\textbf{k}_k}_{\ell_k}\bm{X}\|^2 \\
    &=
    \|(\tilde{\Delta}^{\textbf{k}_{k}})_{j:}\|^2
    + \left(\eta^{\textbf{k}_{k}}_{\ell_k}
    \lambda_{\ell_k, j}^{\textbf{k}_k} \right)^2
    \|
    (\bm{u}_{\ell_k, j}^{\textbf{k}_k})^T
    (\Delta_W^{\textbf{k}_{k}}
    \bm{X}_{:i_k}
    -\mathcal{E}_{:,i_k})
    \bm{X}_{:i_k}^T\tilde{\bm{B}}^{\textbf{k}_k}_{\ell_k}\bm{X}\|^2
    \\
    &\qquad-2\eta^{\textbf{k}_{k}}_{\ell_k}
    \lambda_{\ell_k, j}^{\textbf{k}_k} 
    (\bm{u}_{\ell_k, j}^{\textbf{k}_k})^T
    (\Delta_W^{\textbf{k}_{k}}
    \bm{X}_{:i_k}
    -\mathcal{E}_{:,i_k})
    \bm{X}_{:i_k}^T\tilde{\bm{B}}^{\textbf{k}_k}_{\ell_k}\bm{X}
    (\Delta^{\textbf{k}_{k}})^T\bm{u}_{\ell_k, j}^{\textbf{k}_k}.
\end{align*}
Note that 
\begin{align*}
    \|(\bm{u}_{\ell_k, j}^{\textbf{k}_k})^T
    (\Delta_W^{\textbf{k}_{k}}\bm{X}_{:i_k}-\mathcal{E}_{:,i_k})
    \bm{X}_{:i_k}^T\tilde{\bm{B}}^{\textbf{k}_k}_{\ell_k}\bm{X}\|^2 
    = \|\bm{X}_{:i_k}^T\tilde{\bm{B}}^{\text{k}_k}_{\ell_k}\bm{X}\|^2\bm{Q},
\end{align*}
where
$$
\bm{Q} = \|(\bm{u}_{\ell_k, j}^{\textbf{k}_k})^T\Delta_W^{\textbf{k}_{k}}
\bm{X}_{:i_k}\|^2 + \|(\bm{u}_{\ell_k,j}^{\textbf{k}_k})^T\mathcal{E}_{:,i_k}\|^2
-2
(\bm{u}_{\ell_k, j}^{\textbf{k}_k})^T\Delta_W^{\textbf{k}_{k}}
\bm{X}_{:i_k}(\mathcal{E}_{:,i_k})^T
\bm{u}_{\ell_k, j}^{\textbf{k}_k}.
$$
Thus,
\begin{align*}
    &\|(\tilde{\Delta}^{\textbf{k}_{k+1}})_{j:}\|^2 \\
    &=
    \|(\tilde{\Delta}^{\textbf{k}_{k}})_{j:}\|^2
    -2\eta^{\textbf{k}_{k}}_{\ell_k}
    \lambda_{\ell_k, j}^{\textbf{k}_k} 
    (\bm{u}_{\ell_k, j}^{\textbf{k}_k})^T
    \Delta_W^{\textbf{k}_{k}}
    (\bm{X}_{:i_k}\bm{X}_{:i_k}^T) \tilde{\bm{B}}^{\text{k}_k}_{\ell_k}\bm{X}
    (\Delta^{\textbf{k}_{k}})^T
    \bm{u}_{\ell_k, j}^{\textbf{k}_k}
    \\
    &\qquad
    + (\eta^{\textbf{k}_{k}}_{\ell_k}
    \lambda_{\ell_k, j}^{\textbf{k}_k} )^2\|\bm{X}_{:i_k}^T\tilde{\bm{B}}^{\text{k}_k}_{\ell_k}\bm{X}\|^2\bm{Q}
	\\
    &\qquad +2\eta^{\textbf{k}_{k}}_{\ell_k}
    \lambda_{\ell_k, j}^{\textbf{k}_k} 
    (\bm{u}_{\ell_k, j}^{\textbf{k}_k})^T
    \Delta_W^{\textbf{k}_{k}}
    \tilde{\bm{B}}^{\text{k}_k}_{\ell_k}\bm{X}
    \bm{X}_{:i_k}(\mathcal{E}_{:,i_k})^T
    \bm{u}_{\ell_k, j}^{\textbf{k}_k}.
\end{align*}
Let 
$$
\bm{B}_{\ell_k}^{\textbf{k}_{k}} =\bm{X}^T (\bm{W}_{(\ell_k-1):1}^{\textbf{k}_{k}})^T
\bm{W}_{(\ell_k-1):1}^{\textbf{k}_{k}}\bm{X}.
$$
Let us reparameterize the learning rate as 
$\eta^{\textbf{k}_{k}}_{\ell_k} = \tilde{\eta}_{\ell_k}^{\textbf{k}_{k}}/\|\bm{X}_{:,i_k}^T\tilde{\bm{B}}^{\text{k}_k}_{\ell_k}\bm{X}\|^2$
and define a discrete probability distribution $\bm{\pi}$ on $[m]$ to be
$\bm{\pi}(i) = \|\bm{X}_{:i}^T\tilde{\bm{B}}^{\text{k}_k}_{\ell_k}\bm{X}\|^2/\|\bm{X}^T\tilde{\bm{B}}^{\text{k}_k}_{\ell_k}\bm{X}\|_F^2$
for $1 \le i \le m$.
Since 
$$
\bm{X}\mathcal{E}^T = \bm{X}(\bm{Y}(\bm{I}-\bm{X}^\dagger\bm{X})^T)^T
= \bm{X}(\bm{I}-\bm{X}^\dagger\bm{X})\bm{Y}^T
= (\bm{X}- \bm{X}\bm{X}^\dagger\bm{X})\bm{Y}^T= 0,
$$
we have $\mathbf{E}_{i_k}\left[\frac{\bm{X}_{:i_k}(\mathcal{E}_{:,i_k})^T}{\|\bm{X}_{:i_k}^T\tilde{\bm{B}}^{\text{k}_k}_{\ell_k}\bm{X}\|^2}\right] = \frac{\bm{X}\mathcal{E}^T}{\|\bm{X}^T\tilde{\bm{B}}^{\text{k}_k}_{\ell_k}\bm{X}\|_F^2} = 0$.
By taking the expectation $\mathbf{E}_{i_k}$ with respect to $i_k \sim \bm{\pi}$, 
we obtain
\begin{align*}
    &\mathbf{E}_{i_k}[\|(\tilde{\Delta}^{\textbf{k}_{k+1}})_{j:}\|^2]
    \\
    &=
    \|(\tilde{\Delta}^{\textbf{k}_{k}})_{j:}\|^2
    -2\tilde{\eta}_{\ell_k}^{\textbf{k}_{k}} 
    \lambda_{\ell_k, j}^{\textbf{k}_k}
    (\bm{u}_{\ell_k, j}^{\textbf{k}_k})^T
    \Delta_W^{\textbf{k}_{k}}
    \mathbf{E}_{i_k}\left[\frac{\bm{X}_{:i_k}\bm{X}_{:i_k}^T}{\|\bm{X}_{:i_k}^T\tilde{\bm{B}}^{\text{k}_k}_{\ell_k}\bm{X}\|^2}\right] \tilde{\bm{B}}^{\textbf{k}_k}_{\ell_k}\bm{X}(\Delta^{\textbf{k}_{k}})^T
    \bm{u}_{\ell_k, j}^{\textbf{k}_k}
    \\
    &\qquad
    +(\tilde{\eta}_{\ell_k}^{\textbf{k}_{k}}
    \lambda_{\ell_k, j}^{\textbf{k}_k})^2
    \left((\bm{u}_{\ell_k, j}^{\textbf{k}_k})^T
    \Delta_W^{\textbf{k}_{k}}
    \mathbf{E}_{i_k}\left[\frac{\bm{X}_{:i_k}\bm{X}_{:i_k}^T}{\|\bm{X}_{:i_k}^T\tilde{\bm{B}}^{\text{k}_k}_{\ell_k}\bm{X}\|^2}\right] (\Delta_W^{\textbf{k}_{k}})^T
    \bm{u}_{\ell_k, j}^{\textbf{k}_k}
    \right)
    \\
    &\qquad\qquad
    +(\tilde{\eta}_{\ell_k}^{\textbf{k}_{k}}
    \lambda_{\ell_k, j}^{\textbf{k}_k})^2
    \left((\bm{u}_{\ell_k, j}^{\textbf{k}_k})^T
    \mathbf{E}_{i_k}\left[\frac{\mathcal{E}_{:,i_k}(\mathcal{E}_{:,i_k})^T}{\|\bm{X}_{:i_k}^T\tilde{\bm{B}}^{\text{k}_k}_{\ell_k}\bm{X}\|^2}\right] 
    \bm{u}_{\ell_k, j}^{\textbf{k}_k}
    \right)
    \\
    &=
    \|(\tilde{\Delta}^{\textbf{k}_{k}})_{j:}\|^2
    -2\frac{\tilde{\eta}_{\ell_k}^{\textbf{k}_{k}} \lambda_{\ell_k,j}^{\textbf{k}_k}}{\|\bm{W}_{(\ell_k-1):1}^{\textbf{k}_{k}}\bm{X}\|_F^4}
    \|(\bm{W}_{(\ell_k-1):1}^{\textbf{k}_{k}}\bm{X})(\Delta^{\textbf{k}_{k}})^T\bm{u}_{\ell_k, j}^{\textbf{k}_k}\|^2
    \\
    &\qquad
    +\frac{(\tilde{\eta}^{\ell_k}_{\textbf{k}_{k}}
    \lambda_{\ell_k,i}^{\textbf{k}_k})^2}{\|\bm{W}_{(\ell_k-1):1}^{\textbf{k}_{k}}\bm{X}\|_F^4}
    \left( \|(\Delta^{\textbf{k}_{k}})^T\bm{u}_{\ell_k, j}^{\textbf{k}_k}\|^2
    +\|\mathcal{E}^T\bm{u}_{\ell_k, j}^{\textbf{k}_k}\|^2
    \right)
    \\
    &=
    \|(\tilde{\Delta}^{\textbf{k}_{k}})_{j:}\|^2
    +\frac{(\tilde{\eta}_{\ell_k}^{\textbf{k}_{k}}
    \lambda_{\ell_k,j}^{\textbf{k}_k})^2}{\|\bm{W}_{(\ell_k-1):1}^{\textbf{k}_{k}}\bm{X}\|_F^4}
    \|(\bm{u}_{\ell_k, j}^{\textbf{k}_k})^T\mathcal{E}\|^2
    \\
    &\qquad
    -\frac{\tilde{\eta}_{\ell_k}^{\textbf{k}_{k}}\lambda_{\ell_k,j}^{\textbf{k}_k}}{\|\bm{W}_{(\ell_k-1):1}^{\textbf{k}_{k}}\bm{X}\|_F^4}
    (\bm{u}_{\ell_k, j}^{\textbf{k}_k})^T
    \Delta^{\textbf{k}_{k}}
    \left(
    -\tilde{\eta}_{\ell_k}^{\textbf{k}_{k}}\lambda_{\ell_k,j}^{\textbf{k}_k}\bm{I}
    +2
    \bm{B}^{\textbf{k}_k}_{\ell_k}\right)(\Delta^{\textbf{k}_{k}})^T
    \bm{u}_{\ell_k, j}^{\textbf{k}_k}.
\end{align*}
Suppose
$$
0 < \tilde{\eta}_{\ell_k}^{\textbf{k}_{k}} < \frac{2 \lambda_{\min}(\bm{B}_{\ell_k}^{\textbf{k}_k})}{ \lambda_{\max}(\bm{A}_{\ell_k}^{\textbf{k}_k})}
$$
and let 
$\bm{M}^{\textbf{k}_k}_{\ell_k,j}:=-\tilde{\eta}_{\ell_k}^{\textbf{k}_{k}}\lambda_{\ell_k,j}^{\textbf{k}_k}\bm{I}
    +2\bm{B}^{\textbf{k}_k}_{\ell_k}.
$
Then, since 
$\lambda_{\min}(\bm{M}^{\textbf{k}_k}_{\ell_k, j}) = 2\lambda_{\min}(\bm{B}_{\ell_k}^{\textbf{k}_k}) - \tilde{\eta}_{\ell_k}^{\textbf{k}_{k}}\lambda_{\ell_k,j}^{\textbf{k}_k}
 > 0$,
$\bm{M}^{\textbf{k}_k}_{\ell_k,j}$ is a positive definite symmetric matrix for all $j$.
Thus,
\begin{align*}
    &\mathbf{E}_{i_k}[\|(\tilde{\Delta}^{\textbf{k}_{k+1}})_{j:}\|^2]
    \\
    &\le
    \|(\tilde{\Delta}^{\textbf{k}_{k}})_{j:}\|^2
    -\frac{\tilde{\eta}_{\ell_k}^{\textbf{k}_{k}} \lambda_{\ell_k,j}^{\textbf{k}_k}}{\|\bm{W}_{(\ell_k-1):1}^{\textbf{k}_{k}}\bm{X}\|_F^4}
    \lambda_{\min}(\bm{M}^{\textbf{k}_k}_{\ell_k, j})
    \|(\bm{u}_{\ell_k, j}^{\textbf{k}_k})^T
    \Delta^{\textbf{k}_{k}}\|^2
    \\
    &\qquad+\frac{(\tilde{\eta}_{\ell_k}^{\textbf{k}_{k}}\lambda_{\ell_k,j}^{\textbf{k}_k})^2}{\|\bm{W}_{(\ell_k-1):1}^{\textbf{k}_{k}}\bm{X}\|_F^4}
    \|(\bm{u}_{\ell_k, j}^{\textbf{k}_k})^T\mathcal{E}\|^2
    \\
    &\le \left(1 - \frac{\tilde{\eta}_{\ell_k}^{\textbf{k}_{k}} \lambda_{\ell_k,j}^{\textbf{k}_k}
    \lambda_{\min}(\bm{M}^{\textbf{k}_k}_{\ell_k, j})
    }{\|\bm{W}_{(\ell_k-1):1}^{\textbf{k}_{k}}\bm{X}\|_F^4}\right)
    \|(\bm{u}_{\ell_k, j}^{\textbf{k}_k})^T\Delta^{\textbf{k}_{k}}\|^2 
    +\frac{(\tilde{\eta}_{\ell_k}^{\textbf{k}_{k}}\lambda_{\ell_k,j}^{\textbf{k}_k})^2}{\|\bm{W}_{(\ell_k-1):1}^{\textbf{k}_{k}}\bm{X}\|_F^4} \|(\bm{u}_{\ell_k, j}^{\textbf{k}_k})^T
    \mathcal{E}\|^2,
\end{align*}
and similarly, we have
\begin{align*}
&\mathbf{E}_{i_k}[\|(\tilde{\Delta}^{\textbf{k}_{k+1}})_{j:}\|^2]
\\
&\ge \left(1 - \frac{\tilde{\eta}_{\ell_k}^{\textbf{k}_{k}} \lambda_{\ell_k,j}^{\textbf{k}_k}
	\lambda_{\max}(\bm{M}^{\textbf{k}_k}_{\ell_k, j})
}{\|\bm{W}_{(\ell_k-1):1}^{\textbf{k}_{k}}\bm{X}\|_F^4}\right)
\|(\bm{u}_{\ell_k, j}^{\textbf{k}_k})^T\Delta^{\textbf{k}_{k}}\|^2 
+\frac{(\tilde{\eta}_{\ell_k}^{\textbf{k}_{k}}\lambda_{\ell_k,j}^{\textbf{k}_k})^2}{\|\bm{W}_{(\ell_k-1):1}^{\textbf{k}_{k}}\bm{X}\|_F^4} \|(\bm{u}_{\ell_k,j}^{\textbf{k}_k})^T
\mathcal{E}\|^2.
\end{align*}
Since 
\begin{align*}
    -\tilde{\eta}_{\ell_k}^{\textbf{k}_{k}} \lambda_{\ell_k,j}^{\textbf{k}_k}
    \lambda_{\min}(\bm{M}^{\textbf{k}_k}_{\ell_k, j})
    &= 
	-\tilde{\eta}_{\ell_k}^{\textbf{k}_{k}} \lambda_{\ell_k,i}^{\textbf{k}_k}
    (2\lambda_{\min}(\bm{B}_{\ell_k}^{\textbf{k}_k}) - \tilde{\eta}_{\ell_k}^{\textbf{k}_{k}}\lambda_{\ell_k,i}^{\textbf{k}_k}
    )
    \\
    &=
    \left((\tilde{\eta}_{\ell_k}^{\textbf{k}_{k}}\lambda_{\ell_k,i}^{\textbf{k}_k})
    -\lambda_{\min}(\bm{B}_{\ell_k}^{\textbf{k}_k})
    \right)^2
    -\lambda^2_{\min}(\bm{B}_{\ell_k}^{\textbf{k}_k}),
\end{align*}
if we set
$
\tilde{\eta}_{\ell_k}^{\textbf{k}_{k}} = \eta \frac{\lambda_{\min}(\bm{B}_{\ell_k}^{\textbf{k}_k})}{\lambda_{\max}(\bm{A}_{\ell_k}^{\textbf{k}_k})},
$
where $0 < \eta < 2$,
we have
$$
-\tilde{\eta}_{\ell_k}^{\textbf{k}_{k}} \lambda_{\ell_k,i}^{\textbf{k}_k}
    \lambda_{\min}(\bm{M}^{\textbf{k}_k}_{\ell_k, i}) 
    \le -\lambda_{\min}^2(\bm{B}^{\textbf{k}_k}_{\ell_k})
\left(1 - (1-\eta/\kappa(\bm{A}^{\textbf{k}_k}_{\ell_k}))^2 \right) := -\gamma^{\textbf{k}_k}_{\ell_k}.
$$
Thus, we obtain
\begin{multline*}
    \mathbf{E}_{i_k}[\|(\tilde{\Delta}^{\textbf{k}_{k+1}})_{j:}\|^2]
    \\
    \le
    \left(1 - \frac{\gamma^{\textbf{k}_k}_{\ell_k}}{\|\bm{W}_{(\ell_k-1):1}^{\textbf{k}_{k}}\bm{X}\|_F^4}\right)\|(\tilde{\Delta}^{\textbf{k}_{k}})_{j:}\|^2
    +
    \frac{\eta^2\lambda^2_{\min}(\bm{B}_{\ell_k}^{\textbf{k}_k})\|(\bm{u}_{\ell_k, j}^{\textbf{k}_k})^T\mathcal{E}\|^2}{\|\bm{W}_{(\ell_k-1):1}^{\textbf{k}_{k}}\bm{X}\|_F^4}.
\end{multline*}
By summing up with respect to $j$, we have
\begin{align*}
    \mathbf{E}_{i_k}[\|{\Delta}^{\textbf{k}_{k+1}}\|_F^2]
    &\le
    \left(1 - \frac{\left(1 - (1-\eta/\kappa^2(\bm{W}_{L:(\ell_k+1)}^{\textbf{k}_{k}}))^2 \right)}{\tilde{\kappa}^4(\bm{W}_{(\ell_k-1):1}^{\textbf{k}_{k}}\bm{X})}\right)\|{\Delta}^{\textbf{k}_{k}}\|_F^2
    +
    \frac{\eta^2\|\mathcal{E}\|_F^2}{\tilde{\kappa}^4(\bm{W}_{(\ell_k-1):1}^{\textbf{k}_{k}}\bm{X})},
\end{align*}
where $\tilde{\kappa}(\cdot)$ is the scaled condition number defined to be
$
\tilde{\kappa}(\bm{X}) = \frac{\|\bm{X}\|_F}{|\sigma_{\min}(\bm{X})|}.
$
Similarly, since 
\begin{align*}
\tilde{\eta}_{\ell_k}^{\textbf{k}_{k}}\lambda_{\ell_k,i}^{\textbf{k}_k}\lambda_{\max}(\bm{M}^{\textbf{k}_k}_{\ell_k, j})
&= 
2\lambda_{\max}(\bm{B}_{\ell_k}^{\textbf{k}_k})(\tilde{\eta}_{\ell_k}^{\textbf{k}_{k}}\lambda_{\ell_k,i}^{\textbf{k}_k}) - (\tilde{\eta}_{\ell_k}^{\textbf{k}_{k}}\lambda_{\ell_k,i}^{\textbf{k}_k})^2
\\
&=
\lambda^2_{\max}(\bm{B}_{\ell_k}^{\textbf{k}_k}) -
\left((\tilde{\eta}_{\ell_k}^{\textbf{k}_{k}}\lambda_{\ell_k,i}^{\textbf{k}_k}) - \lambda_{\max}(\bm{B}_{\ell_k}^{\textbf{k}_k})\right)^2
\\
&= 
\lambda^2_{\max}(\bm{B}_{\ell_k}^{\textbf{k}_k}) -
\left(\eta \lambda_{\ell_k,i}^{\textbf{k}_k}\frac{\lambda_{\min}(\bm{B}_{\ell_k}^{\textbf{k}_k})}{\lambda_{\max}(\bm{A}_{\ell_k}^{\textbf{k}_k})} - \lambda_{\max}(\bm{B}_{\ell_k}^{\textbf{k}_k})\right)^2
\\
&\le
\lambda^2_{\max}(\bm{B}_{\ell_k}^{\textbf{k}_k})
\left(1 - \left(1 - \frac{\eta}{\kappa(\bm{B}_{\ell_k}^{\textbf{k}_k})}\right)^2 \right),
\end{align*}
we have
\begin{align*}
\mathbf{E}_{i_k}[\|{\Delta}^{\textbf{k}_{k+1}}\|_F^2]
\ge
r\|\tilde{\Delta}^{\textbf{k}_{k}}\|^2_F
+
\frac{\eta^2\|\mathcal{E}\|_F^2}{\kappa^4(\bm{W}_{L:(\ell_k+1)}^{\textbf{k}_{k}})\tilde{\kappa}^4(\bm{W}_{(\ell_k-1):1}^{\textbf{k}_{k}}\bm{X})},
\end{align*}
where
$$
r= 1 - \frac{\lambda^2_{\max}(\bm{B}_{\ell_k}^{\textbf{k}_k})
	\left(1 - \left(1 - \frac{\eta}{\kappa(\bm{B}_{\ell_k}^{\textbf{k}_k})}\right)^2 \right)}{\|\bm{W}_{(\ell_k-1):1}^{\textbf{k}_{k}}\bm{X}\|_F^4}.
$$
\end{proof}

\bibliographystyle{siamplain}
\bibliography{references}
\end{document}